\documentclass{article}
\PassOptionsToPackage{numbers,sort&compress}{natbib}

\usepackage[eandd, preprint]{neurips_2026}

\usepackage[utf8]{inputenc}
\usepackage[T1]{fontenc}
\usepackage{hyperref}
\usepackage{url}
\usepackage{booktabs}
\usepackage{amsfonts}
\usepackage{nicefrac}
\usepackage{microtype}
\usepackage[table,xcdraw,dvipsnames]{xcolor}

\usepackage[margin=10pt]{caption}
\usepackage{graphicx}
\usepackage{amsmath}
\usepackage{amssymb}
\usepackage[capitalize]{cleveref}
\crefname{section}{Sec.}{Secs.}
\Crefname{section}{Section}{Sections}
\Crefname{table}{Table}{Tables}
\crefname{table}{Tab.}{Tabs.}
\crefname{equation}{}{}
\Crefname{equation}{}{}

\usepackage{epsfig}
\usepackage{enumitem}
\setlist{nosep}

\usepackage{pifont}

\usepackage{diagbox}
\usepackage{derivative}
\usepackage{mwe}
\usepackage{bbm}
\usepackage{multirow}
\usepackage{comment}
\usepackage{wrapfig}
\usepackage{dsfont}
\usepackage{inconsolata}
\usepackage{float}
\usepackage[ruled,vlined,linesnumbered]{algorithm2e}
\usepackage{stfloats}
\usepackage{placeins}
\usepackage{tabularx}
\usepackage{array}

\definecolor{navyblue}{HTML}{0071BC}
\definecolor{lightyellow}{HTML}{FEFAC8}
\definecolor{oai-green-200}{HTML}{D9FFD8}
\definecolor{oai-green-400}{HTML}{A6FFA3}
\definecolor{oai-green-600}{HTML}{51DA4C}

\crefname{section}{Sec.}{Secs.}
\Crefname{section}{Section}{Sections}
\Crefname{table}{Table}{Tables}
\crefname{table}{Tab.}{Tabs.}

\newcommand{\App}[1] {Appendix \ref{app:#1}}

\newcommand{\gcmark}{\textcolor{green!70!black}{\ding{51}}}
\newcommand{\rxmark}{\textcolor{red}{\ding{55}}}

\newcommand{\ourmodel}[1]{\textcolor{red}{SpatialMosaicVLM}}
\newcommand{\ourdataset}[1]{\textcolor{black}{\textit{SpatialMosaic}}}
\newcommand{\ourbenchmark}[1]{\textcolor{black}{\textit{SpatialMosaic-Bench}}}
\newcommand{\numvqa}[1]{\textcolor{black}{2M}}
\newcommand{\numvqabench}[1]{\textcolor{black}{1M}}
\newcommand{\numtask}[1]{\textcolor{black}{11}}

\newcommand{\revA}[1]{\textbf{\textcolor{OliveGreen}{9AwX}}}
\newcommand{\revB}[1]{\textbf{\textcolor{RoyalBlue}{jKHT}}}
\newcommand{\revC}[1]{\textbf{\textcolor{Maroon}{Xbmj}}}

\title{SpatialMosaic: A Multi-View VLM Dataset for\\ Partial Visibility}

\author{%
  Kanghee Lee$^{1}$\quad \ Jungi Hong$^{1}$\quad \ Sion Lee$^{2}$\quad \ Injae Lee$^{1}$ \\ 
  \textbf{Minseok Kwak$^{3}$\quad Kwonyoung Ryu$^{4}$\quad Jaesik Park$^{1}$}  
  \vspace{.8em} \\
  $^1$ Seoul National University \quad $^2$ Kyung Hee University \\
  $^3$University College London \quad $^{4}$POSTECH \\
  \texttt{\{kanghee.lee, jaesik.park\}@snu.ac.kr}
  \vspace{.8em} \\
  \url{https://kanghee-lee.github.io/spatialmosaic/}
}

\begin{document}
\maketitle
\vspace{-3.0em}
\begin{figure}[h]
    \centering
    \includegraphics[width=\linewidth]{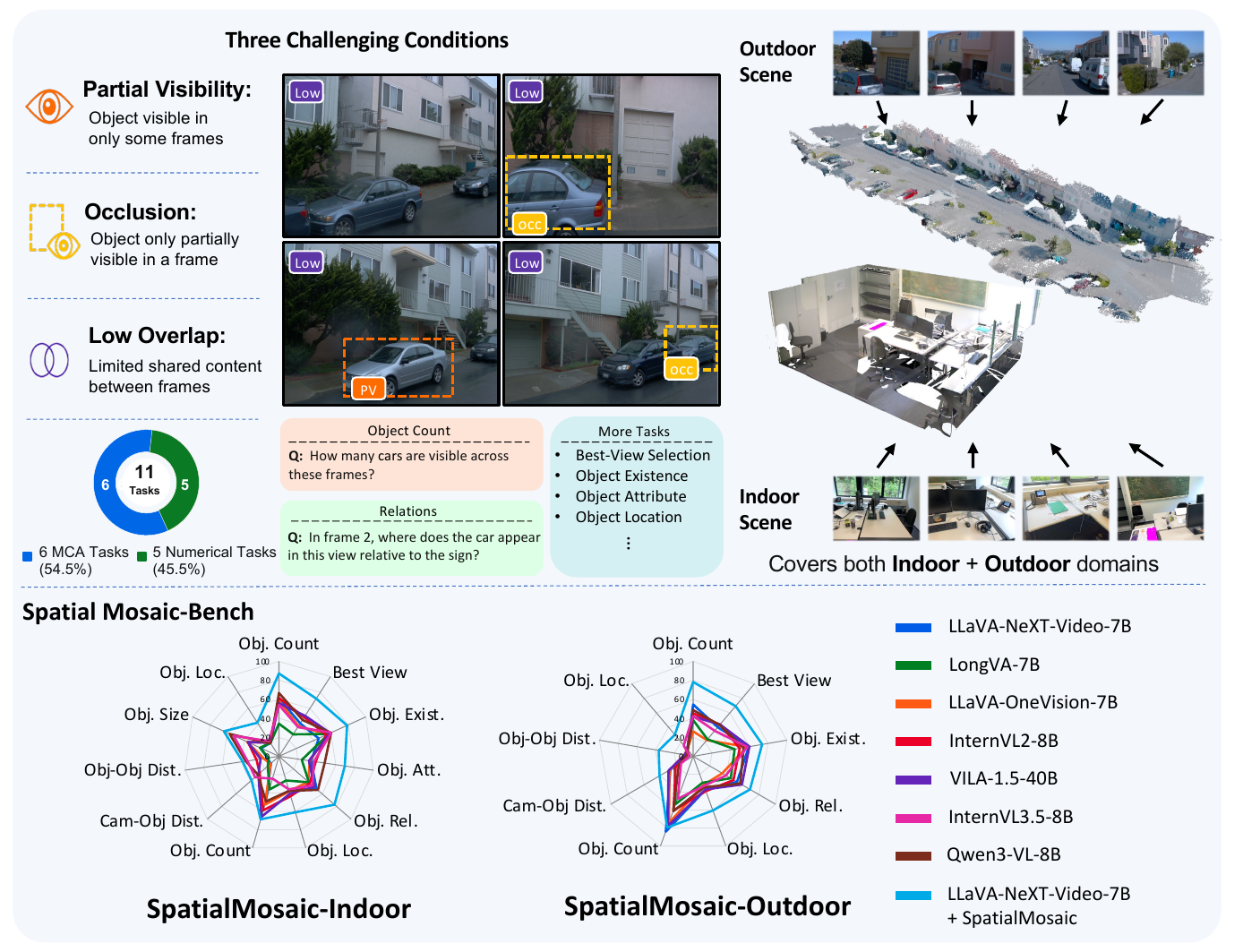}
    \caption{{\textbf{SpatialMosaic} is a multi-view VLM dataset spanning both indoor and outdoor scenes under three realistic and challenging conditions: partial visibility, occlusion, and low-overlap. It provides 11 geometry-grounded tasks for training and evaluation, covering categorical reasoning and numerical spatial estimation. The radar plots summarize performance on SpatialMosaic-Indoor and SpatialMosaic-Outdoor, highlighting the difficulty of reasoning from sparse and fragmented multi-view observations.}}
    \label{fig:teaser}
\end{figure}
\begin{abstract}
Recent progress in Multimodal Large Language Models (MLLMs) has enabled 3D scene understanding and spatial reasoning directly from multi-view images, without requiring explicit 3D reconstructions. 
Nevertheless, key challenges that frequently arise in real-world environments, such as partial visibility, occlusion, and low-overlap conditions that require reasoning from fragmented visual cues, remain under-explored. To address these limitations, we propose a scalable multi-view data generation and annotation pipeline that constructs realistic spatial reasoning QAs, resulting in \ourdataset{}, a comprehensive instruction-tuning dataset with \numvqa{} QA pairs. 
We further introduce \ourbenchmark{}, a challenging benchmark for evaluating multi-view spatial reasoning under complex and diverse scenarios, consisting of \numvqabench{} QA pairs across \numtask{} tasks with both multiple-choice and numerical-answer formats. 
Our dataset spans both indoor and outdoor scenes, enabling comprehensive evaluation across diverse real-world scenarios. In addition, we provide a practical baseline for multi-view settings by integrating geometry encoders into VLMs for improved cross-view consistency and spatial grounding. Extensive experiments demonstrate that our dataset effectively enhances spatial reasoning under challenging multi-view conditions, validating the effectiveness of our data generation pipeline in constructing realistic and challenging QAs.

\end{abstract}

\section{Introduction}
\label{sec:intro}

\definecolor{beige}{RGB}{249,238,220}  
\definecolor{lemon}{RGB}{251,248,216} 

\begin{table}[t!]
    \centering
    \tiny
    \setlength{\tabcolsep}{1.1pt}
    \renewcommand{\arraystretch}{0.82}
    \caption{
    Multi-view spatial dataset comparison.
    ``In'' and ``Out'' indicate explicit indoor and outdoor scene coverage.
    ``Diag.'' indicates per-example condition labels for diagnostic analyses such as partial visibility, occlusion, and low-overlap views. ``Open'' indicates that the dataset is publicly available.
    }
    \label{tbl:dataset_comparison}
    \resizebox{0.9\textwidth}{!}{
      \begin{tabular}{ccccccccccccc}
        \toprule
        \multirow{2}{*}{Dataset} 
        & \multicolumn{2}{c}{Split} 
        & \multirow{2}{*}{\#Views} 
        & \multirow{2}{*}{\#Tasks} 
        & \multirow{2}{*}{\#Images} 
        & \multicolumn{2}{c}{\#QAs} 
        & \multirow{2}{*}{In}
        & \multirow{2}{*}{Out}
        & \multirow{2}{*}{Diag.}
        & \multirow{2}{*}{Open} \\
        & Train & Eval & & & & Train & Eval & & & & \\
        \midrule
        VSI-Bench~\cite{yang2025thinking} 
        & \rxmark & \gcmark & 8/16/32 & 8 & 288 videos & -- & 5.1K 
        & \gcmark & \rxmark & \rxmark & \gcmark \\
        BLINK~\cite{fu2024blink} 
        & \rxmark & \gcmark & 1--4 & 14 & 7.3K & -- & 3.8K 
        & \gcmark & \gcmark & \rxmark & \gcmark \\

        MMIU~\cite{mmiu}
        & \rxmark & \gcmark & 2--12 & 52 & 77K & -- & 11K 
        & \gcmark & \gcmark & \rxmark & \gcmark \\

        All-Angles Bench~\cite{allangles}
        & \rxmark & \gcmark & 4 & 6 & -- & -- & 2.1K 
        & \gcmark & \rxmark & \rxmark & \gcmark \\

        UniQA-3D~\cite{zuo2025towards} 
        & \rxmark & \gcmark & 1--2 & 4 & 2.5K & -- & 2.5K 
        & \gcmark & \gcmark & \rxmark & \rxmark \\

        MultiSPA~\cite{xu2025multi} 
        & \gcmark & \gcmark & 1--5 & 26 & 1.1M & 27M & 7.8K 
        & \gcmark & \gcmark & \rxmark & \rxmark \\

        SPAR~\cite{zhang2025flatland} 
        & \gcmark & \gcmark & 1--5 & 12 & 16.3M & 7M & 7.2K 
        & \gcmark & \rxmark & \rxmark & \gcmark \\
        
        \midrule
        \ourdataset{} 
        & \gcmark & \gcmark & 2--5 & 11 & 8M & 2M & 1M 
        & \gcmark & \gcmark & \gcmark & \gcmark\\
        \bottomrule
      \end{tabular}
     }
     \vspace{-1.0em}
\end{table}

Spatial reasoning in 3D environments is a cornerstone of embodied intelligence, enabling agents to interpret complex scenes and interact effectively with the physical world. 
Recent progress in MLLMs~\cite{alayrac2022flamingo,li2023blip,zhu2023minigpt,liu2023visual} has raised the possibility of endowing them with human-level 3D spatial understanding, extending their success in 2D perception to complex tasks such as depth estimation~\cite{zhang2025flatland,zuo2025towards,xu2025multi}, metric distance prediction~\cite{xu2025multi,daxberger2025mm,zhang2025flatland}, and holistic spatial reasoning capabilities~\cite{xu2025multi,fu2024blink,zhang2025flatland}.
However, existing approaches often rely on pre-built 3D representations or off-the-shelf reconstruction modules~\cite{yen2021inerf,martin2021nerf}. Such methods require explicit 3D inputs at inference time, which limits their applicability in real-world environments where pre-built 3D maps are unavailable.

To address these limitations, recent studies~\cite{xu2025multi,zhang2025flatland} have investigated deriving 3D spatial reasoning directly from multi-view images, thereby mitigating reliance on pre-constructed geometric priors or conventional 3D reconstruction pipelines. This paradigm not only alleviated such dependencies but also demonstrated superior performance in challenging 3D spatial reasoning tasks. However, existing works still fall short of capturing real-world conditions where available views are sparse, share only limited visual overlap, and are not provided as temporally adjacent video frames. In contrast, humans can integrate such partial observations across views to implicitly reconstruct coherent 3D scenes and reason about occluded objects that are not fully visible. Whether MLLMs can achieve comparable robustness under such imperfect conditions remains an open question. Building on these observations, we present fundamentally under-explored scenarios and constraints for 3D spatial reasoning in multi-view systems~\cite{allangles,zhao2023mmicl,naeem2023i2mvformer}.

In this paper, we define three types of scenarios that each represent a unique spatial reasoning constraint. \textit{Partial visibility} denotes a condition in multi-view settings where an object is visible only in a subset of views, instead of being observed from all camera viewpoints. \textit{Occlusion} refers to a condition within a single view where the object is partially obscured by other instances or truncated by the camera's field of view. Lastly, \textit{low-overlap} condition represents scenarios where the available views exhibit minimal cross-view overlap, providing limited information for spatial inference.

To study these scenarios systematically, we propose a novel multi-view spatial data generation and annotation pipeline tailored to partial-visibility, occlusion, and low-overlap scenarios. With this pipeline, we construct \ourdataset{}, a comprehensive multi-view instruction-tuning dataset containing \numvqa{} QA pairs that capture challenging, frequently occurring real-world scenarios. Unlike prior multi-view spatial datasets which focus exclusively on either indoor or outdoor layouts, our dataset spans both domains, enabling more comprehensive evaluation and training across diverse real-world scenes. 
To support evaluation in the multi-view settings, we additionally provide a practical baseline that integrates geometry encoders into a VLM framework.
Finally, we release \ourbenchmark{}, which provides a more comprehensive and challenging evaluation of spatial capabilities compared to existing multi-view benchmarks~\cite{fu2024blink,zhang2025flatland,li2024mvbench,yi2019clevrer}. It consists of \numvqabench{} QA pairs across \numtask{} tasks, covering both categorical reasoning and numerical spatial estimation under realistic and challenging scenarios spanning indoor and outdoor environments.
The key contributions are summarized as follows:
\begin{itemize}
    \item We define three under-explored spatial reasoning constraints in multi-view settings, namely \textit{partial visibility}, \textit{occlusion}, and \textit{low-overlap}. Based on these constraints, we propose a scalable data generation pipeline that constructs \ourdataset{}, a realistic and challenging instruction-tuning dataset.
    
    \item To enable comprehensive evaluation under challenging multi-view conditions, we construct \ourbenchmark{}, which spans diverse indoor and outdoor environments. By covering both domains, the benchmark facilitates more comprehensive assessment across heterogeneous scene layouts.
    
    \item We provide a practical baseline for multi-view settings by combining 3D reconstruction models with VLMs to enable cross-view alignment and robust reasoning in realistic, multi-view environments.
    
\end{itemize}

\begin{figure*}[t!]
\centering
\includegraphics[width=0.99\textwidth]{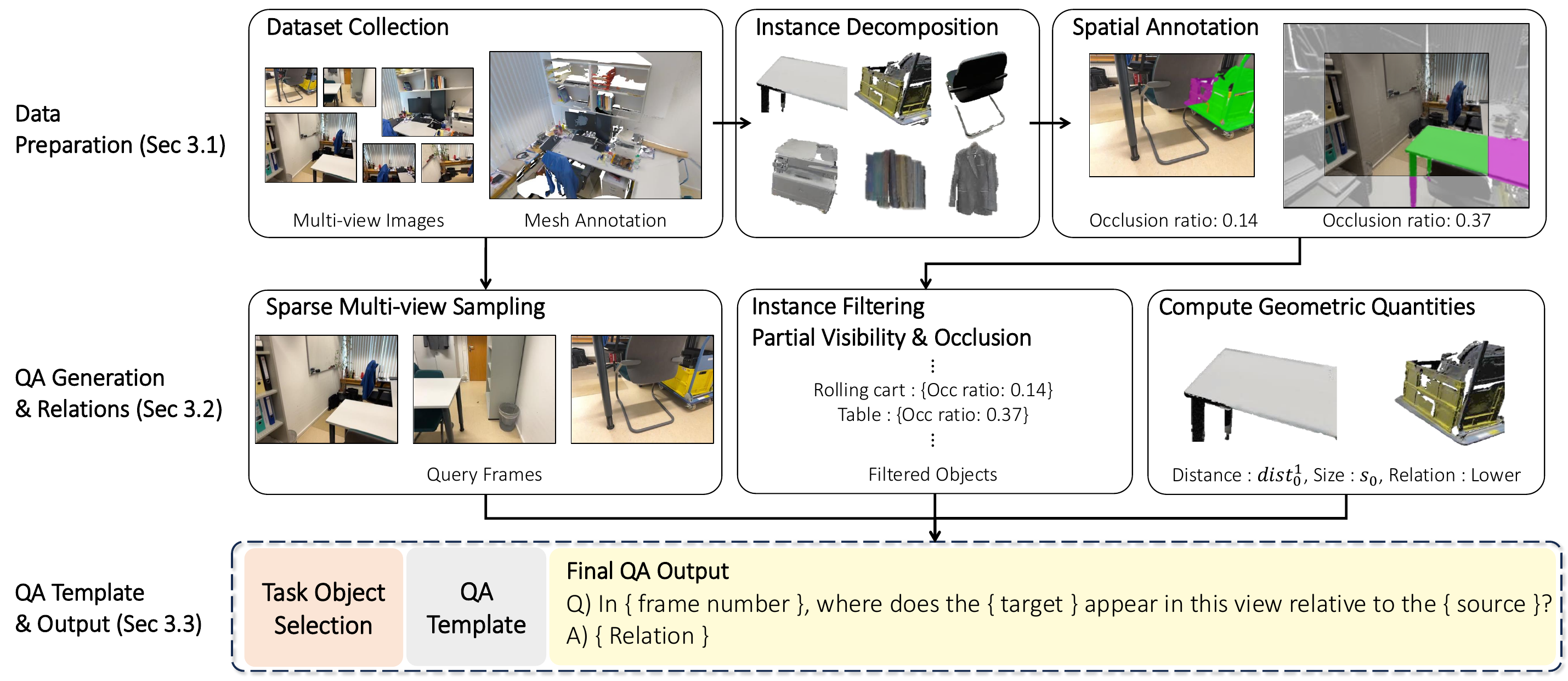}
    \caption{\textbf{\ourdataset{} data generation pipeline.} Given multi-view images and point clouds, we compute various multi-view spatial annotations, including object-level and image-level occlusion ratios for each instance (Sec.~\ref{subsec:data_preparation}). Images are then filtered by overlap to ensure diverse viewpoints, and instances are filtered based on visibility constraints (Sec.~\ref{subsec:QA_Generation_and_Relations}). Finally, spatial relations are computed and used to populate task-specific templates, generating geometrically grounded QA pairs (Sec.~\ref{subsec:QA Template and Output}).}

    \label{fig:data_generation}
\end{figure*}

\vspace{-0.5em}
\section{Related Work}
\label{sec:related}
\noindent
\textbf{Spatial reasoning with MLLMs.}
Multimodal large language models (MLLMs) have demonstrated strong capabilities in open-world visual understanding, excelling at classification, segmentation, and captioning. 
Early MLLMs extend pretrained large language models (LLMs) with visual encoders, enabling instruction-following and open-ended reasoning on single-image inputs~\cite{alayrac2022flamingo,li2023blip,zhu2023minigpt,liu2023visual}.
More recent works move beyond perception by explicitly modeling spatial relations through spatial-reasoning datasets~\cite{yi2018neural}, neural-symbolic modules~\cite{andreas2016neural}, and step-wise reasoning systems~\cite{suris2023vipergpt,hwang2023text2scene}.
Despite these advances, current MLLMs primarily focus on visible 2D cues, limiting their ability to reason when objects are partially visible or occluded in complex spatial configurations.

\noindent
\textbf{Towards 3D-aware Vision-Language Models.}
While 2D-based MLLMs have made progress in spatial reasoning, their reliance on single-view inputs limits the ability to capture full 3D scene structure. Recent approaches incorporate explicit 3D signals, extending vision-language models with depth maps, point clouds, or multi-view consistency.
Recent 3D-aware VLMs integrate 3D representations for visual question answering, grounding, and embodied navigation~\cite{deng20253d,hong20233d,chen2024grounded,fu2024scene,zhi2025lscenellm}.
Transformer-based 3D grounding methods align textual queries with object locations~\cite{huang2022multi}, while real-world 3D QA and grounding benchmarks measure spatial understanding~\cite{azuma2022scanqa,ma2022sqa3d,achlioptas2020referit3d}.
Nevertheless, current models remain challenged by cross-view inconsistencies, which degrade reasoning robustness in complex environments.

\noindent
\textbf{3D Reconstruction Models.}
Recently, large-scale 3D reconstruction models learn generalizable geometric priors from massive multi-view data. Thanks to the rapid advancement of transformers, models~\cite{wang2024dust3r,wang2025vggt} provide 3D point maps with dense correspondences. These models operate over image patches or point tokens, coupled with differentiable geometric modules for epipolar reasoning or multi-view aggregation enabling robust generalization across scenes and domains. As a result, these models provide geometry-aware features that transfer across datasets and tasks. Building on this foundation, we integrate model patch tokens with 3D priors to strengthen multi-modal understanding in real-world spatial tasks. This integration improves cross-view consistency, stabilizes object identity under varying appearances, and enables robust reasoning under occlusion. We employ VGGT~\cite{wang2025vggt} as a geometry encoder to ground the language model with spatial information by fusing CLIP encoder.
\vspace{-0.5em}
\section{SpatialMosaic}
\label{sec:dataset} 
While recent VLM benchmarks~\cite{zhang2025flatland,xu2025multi,yang2025thinking} provide spatial reasoning tasks over multiple images, they either rely on video inputs with sequential frames or do not explicitly address challenges arising from partial visibility, occlusion, and varying image overlap conditions that are pervasive in real-world environments. In practice, these are precisely the scenarios where current VLMs fail, struggling to integrate partial observations and low-overlap images into coherent spatial reasoning. Motivated by these limitations, we first leverage the existing indoor dataset, ScanNet++~\cite{scannetpp}, to construct challenging spatial reasoning benchmarks. To avoid restricting the pipeline to indoor layouts and to enhance robustness in outdoor layouts, we further extend our dataset to outdoor scenes, Waymo Open Dataset (WOD)~\cite{Sun_2020_CVPR}, thereby encouraging generalization beyond indoor-specific scene structures. We propose a scalable data generation framework that produces more than 3M QA pairs covering both categorical reasoning and numerical spatial estimation, explicitly tailored to partial-visibility, occlusion, and low-overlap scenarios. Unlike prior benchmarks, our dataset includes a larger number of images with substantial perspective changes rather than sequential frames, yielding QA pairs that correspond to more realistic and challenging spatial reasoning tasks commonly encountered in real-world environments. Our data generation pipeline consists of three main stages, as illustrated in Fig.~\ref{fig:data_generation}: (1) Data Preparation, (2) QA Generation and Relations, and (3) QA Template and Output.

\begin{figure}[t]
    \centering
    \includegraphics[width=\columnwidth] {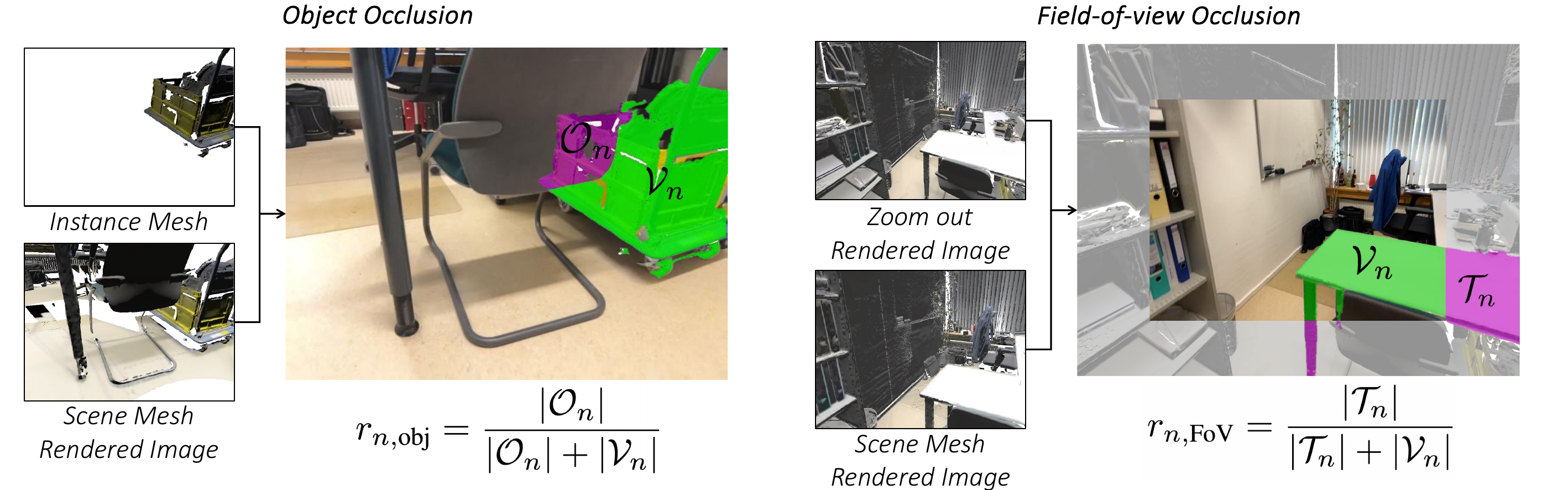}
    \caption{\textbf{Occlusion ratio calculation.} We render each instance independently to measure visible (green) and occluded (magenta) pixels. \textbf{Object Occlusion}: Object-level occlusion ($r_{\text{obj}}$) captures inter-object obstruction from the actual camera view. \textbf{Field-of-view Occlusion}: Field-of-view truncation ($r_{\text{FoV}}$) uses extended field-of-view rendering to quantify boundary occlusion from frame cropping.}
    \label{fig:occlusion_ratio}
\end{figure}
\newcommand{\Vobj}{\mathcal{V}_{\text{obj}}}
\newcommand{\Oobj}{\mathcal{O}_{\text{obj}}}
\newcommand{\Vfov}{\mathcal{V}_{\text{fov}}}
\newcommand{\Ofov}{\mathcal{O}_{\text{fov}}}

\subsection{Data Preparation}
\label{subsec:data_preparation}

\noindent{\textbf{Object occlusion ratio.}}
We introduce the object occlusion ratio to quantify the degree of occlusion for each instance. For each scene, ScanNet++~\cite{scannetpp} and WOD~\cite{Sun_2020_CVPR} provide annotated 3D point clouds $\mathcal{P} = \bigcup_{n=1}^{N} \mathcal{P}_n$, where $\mathcal{P}_n$ denotes the point cloud of instance $n$. We render the scene depth $\mathbf{D}$ and per-instance depth $\mathbf{D}_n$ by projecting $\mathcal{P}$ and $\mathcal{P}_n$ onto each camera with the provided parameters. A point $\mathbf{p}_n \in \mathcal{P}_n$ is occluded when another surface lies in front ($\mathbf{D}<\mathbf{D}_n$) and visible otherwise, and we define the occlusion ratio $r_{n,\text{obj}} \in [0,1]$ as the fraction of occluded points (Fig.~\ref{fig:occlusion_ratio}):
\begin{equation}
\mathcal{O}_n \!=\! \{\mathbf{p}_n \mid 0\!<\!\mathbf{D}\!<\!\mathbf{D}_n\},\;\;
\mathcal{V}_n \!=\! \{\mathbf{p}_n \mid \mathbf{D}_n\!\le\!\mathbf{D},\, \mathbf{D}_n\!<\!\infty\},\;\;
r_{n,\text{obj}} \!=\! \frac{|\mathcal{O}_n|}{|\mathcal{O}_n|+|\mathcal{V}_n|}.
\label{eq:occlusion_ratio}
\end{equation}

\noindent{\textbf{FoV occlusion ratio.}} In addition to object-level occlusion, instances may be partially occluded due to field of view (FoV) truncation, where parts of the object extend beyond the camera's field of view. To quantify this, we introduce the FoV occlusion ratio, which measures the proportion of the instance truncated by the image boundary.
As illustrated in Fig.~\ref{fig:occlusion_ratio}, we create an extended field of view reference image with doubled resolution $2H \times 2W$ centered around the original view by shifting the principal point to the center of the extended field. The extended intrinsic matrix $\tilde{\mathbf{K}}\in\mathbb{R}^{3\times 3}$ is defined as $\tilde{\mathbf{K}} = [f_x, 0, c_x + W/2; 0, f_y, c_y + H/2; 0, 0, 1]$
where $f_x, f_y$ are the focal lengths and $c_x, c_y$ are the principal point coordinates from the original intrinsic matrix $\mathbf{K}$. Using this extended intrinsic matrix, we project each point to obtain 
$(\tilde{u}, \tilde{v})$ in the extended image coordinate system. We then render both the instance depth map $\tilde{\mathbf{D}}_n$ and the complete scene depth map $\tilde{\mathbf{D}}$ in this extended view.

We define the visible point sets within the original FoV region $\mathcal{R}$ and the FoV-truncated region $\mathcal{T}_n$ as follows:
\begin{equation}
\begin{split}
\mathcal{V}_n &= \{\mathbf{p}_n \in \mathcal{P}_n \mid 0 < \tilde{\mathbf{D}}_n \leq \tilde{\mathbf{D}} < \infty, (\tilde{u}, \tilde{v}) \in \mathcal{R}\} \\
\mathcal{T}_n &= \{\mathbf{p}_n \in \mathcal{P}_n \mid 0 < \tilde{\mathbf{D}}_n \leq \tilde{\mathbf{D}} < \infty, (\tilde{u}, \tilde{v}) \in \tilde{\mathcal{R}} \setminus \mathcal{R}\}
\end{split}
\label{eq:4}
\end{equation}
\noindent where $\mathcal{R}=[W/2, 3W/2)\times[H/2, 3H/2)$ represents the original FoV region and $\tilde{\mathcal{R}} = [0, 2W) \times [0, 2H)$ represents the extended canvas region. The FoV occlusion ratio is then defined as:
\begin{equation}
r_{n,\text{FoV}} = \frac{|\mathcal{T}_n|}{|\mathcal{T}_n| + |\mathcal{V}_n|}
\label{eq:5}
\end{equation}
where $r_{n,\text{FoV}} \in [0, 1]$ quantifies the proportion of points truncated by the FoV boundaries, with $r_{n,\text{FoV}} = 0$ indicating the instance is fully within the original field of view.

\subsection{QA Generation and Relations}
\label{subsec:QA_Generation_and_Relations}

\noindent{\textbf{Sparse Multi-view Sampling.}} To construct multi-frame spatial reasoning tasks, we sample image pairs with limited overlap to encourage integration of diverse viewpoints. For each scene, we compute the overlap ratio between $i, j$th images by measuring the intersection-over-union of their visible 3D points. The overlap ratio is defined as follows,

\begin{equation}
\text{Overlap}(i, j) = \frac{|\mathcal{V}^i \cap \mathcal{V}^j|}{|\mathcal{V}^i \cup \mathcal{V}^j|}, \quad \mathcal{V} = \bigcup_{n=1}^{N} \mathcal{V}_n
\label{eq:6}
.\end{equation}
We retain only image pairs with overlap ratios below $\tau$, thereby filtering out redundant images with excessive shared content. This constraint encourages integrative spatial reasoning across sparse and diverse viewpoints rather than simple matching of overlapping regions.

\noindent{\textbf{Instance Filtering.}} Not all instances are suitable for generating meaningful QA pairs. We filter instances based on two criteria: (1) the target instance must not appear in all images within the selected combination to ensure partial visibility, and (2) the target instance must have an occlusion ratio below 0.9 in at least one image (objects with occlusion ratios above 0.9 are nearly impossible for humans to observe). Additionally, there are task-specific filtering criteria such as selecting objects as target instances only if it is above the minimum occlusion threshold (e.g., above 0.4), and ensuring that the source instance is visible while the target instance is not visible in the selected query image. Collectively, these constraints substantially elevate the spatial complexity of the generated QAs; they reduce reliance on redundant visibility cues, force models to reason under minimal information with asymmetric view conditions, and require accurate spatial inference even when object instances exhibit no correlation under a single image. As a result, our QA set captures challenging real-world visibility patterns that conventional datasets fail to represent.

\noindent{\textbf{Compute Geometric Quantities.}} To support both multiple-choice and numerical-answer tasks, we compute task-specific geometric quantities from the derived annotations, including 2D and 3D bounding boxes, visibility profiles, and camera parameters. For relation-based tasks, we transform each object's 3D oriented bounding box into the camera coordinate system and compare positional relationships along the X, Y, and Z axes. These relationships define the underlying reasoning context, allowing us to determine directional relations (e.g., "to the left side, above") between object pairs. 

\subsection{QA Template and Output}
\label{subsec:QA Template and Output}

\noindent{\textbf{Task and Object Selection.}} Based on the filtered instances and computed quantities, we select appropriate source and target objects for each task type. The selection process considers task requirements and ensures that selected objects satisfy visibility and occlusion constraints established in earlier stages.

\noindent{\textbf{QA Template Generation.}} We define templates for \numtask{} task categories using template-based generation, where placeholders such as [source\_obj], [target\_obj], or [quantities] are replaced with concrete object labels and computed spatial information. 
Fig.~\ref{fig:main_figure} illustrates representative task examples under challenging multi-view conditions. 
For multiple-choice tasks, answer options are automatically constructed with one correct answer and distractors generated from incorrect spatial configurations. For numerical-answer tasks, ground-truth answers are directly derived from computed geometric quantities, such as object counts, distances, sizes, and localization coordinates.
Additional QA templates and detailed generation pipelines are provided in \App{templates} and \App{task}.

\subsection{Human Verification and Bias Analysis}
Since \ourbenchmark{} is constructed through an automatic data generation pipeline, verifying the reliability of the generated examples is critical. We conduct a human verification study on a subset of 3,000 QAs, sampling 100 questions per task category across indoor and outdoor splits, for a total of 70 human hours. The sampled examples are balanced across task categories, visibility conditions, occlusion levels, and view-overlap levels.
Annotators inspect each example using a structured checklist that verifies question clarity, answer correctness, object grounding, visibility-condition validity, and whether the answer can be inferred from the provided multi-view images. After filtering problematic cases, 2,728 of the 3,000 questions are retained, yielding an overall acceptance rate of $90.9\%$. Further details of the verification protocol are provided in \App{human_verification}.

Moreover, we analyze potential biases that may arise during the dataset construction process. To mitigate such biases, we carefully design the benchmark and conduct empirical analyses. Further details of the bias analysis are provided in \App{bias_analysis}.

\subsection{SpatialMosaic Benchmark}

\ourbenchmark{} is a large-scale multi-view benchmark explicitly designed to evaluate spatial reasoning under occlusion and partial visibility. The benchmark comprises \numvqabench{} QA pairs across \numtask{} task categories derived from real-world indoor and outdoor scenes in ScanNet++~\cite{scannetpp} and WOD~\cite{Sun_2020_CVPR}.

\noindent{\textbf{Evaluation protocol.}}
Each QA instance consists of 2-5 frames and a question, paired with either multiple-choice options or a free-form numerical answer.
For multiple-choice tasks, models select one option, and performance is measured by accuracy against ground-truth answers. For numerical-answer tasks, model outputs are parsed into numeric values and evaluated using task-specific tolerance criteria.
To enable fine-grained analysis, we annotate every QA with two diagnostic scenarios and difficulty levels: (1) \textit{Visibility Scenario} indicates whether target objects are consistently visible or partially occluded across frames, and (2) \textit{GT Scenario} indicates whether all instances of the target category in the scene are captured or only a subset is visible. Together, these axes support structured performance breakdowns across a continuum of difficulty levels. 
This diagnostic framework enables us to analyze not only overall accuracy but also the specific multi-view conditions under which models succeed or fail. Additional evaluation details and analyses across difficulty levels are provided in \App{experimental_setting} and \App{analysis_conditions}.

\label{sec:benchmark}

\section{Analysis}
\label{sec:main_analysis}

We analyze whether the realistic and challenging conditions introduced in \ourbenchmark{} directly affect multi-view spatial reasoning.
Our benchmark is designed around three conditions that commonly arise in real multi-view observations, namely partial visibility, low-overlap views, and occlusion.
These conditions require models to integrate fragmented visual evidence across views rather than relying on complete evidence from a single viewpoint.
The full analysis is provided in Appendix~\ref{sec:analysis}, where we further study general multi-view challenges such as object scale and duplicate instances.
Here, we summarize the main findings for the proposed challenging conditions and provide representative analyses for low-overlap views and occlusion.

\noindent\textbf{Partial visibility.}
\ourbenchmark{} enforces partial visibility by constraining the query frame to contain only the source instance while the target instance remains invisible, preventing models from solving the task through single-view shortcuts.
As a result, the model must identify the target from reference views and infer its relation to the source by aligning and aggregating partial observations across views.
Our controlled comparison in Appendix~\ref{sec:analysis} shows that accuracy improves when the source and target are jointly visible.
This result indicates that partial visibility, which commonly occurs in real multi-view settings, substantially increases the difficulty of spatial reasoning.

\begin{figure*}[t]
    \centering
    \begin{minipage}{0.43\textwidth}
        \centering
        \includegraphics[width=\linewidth]{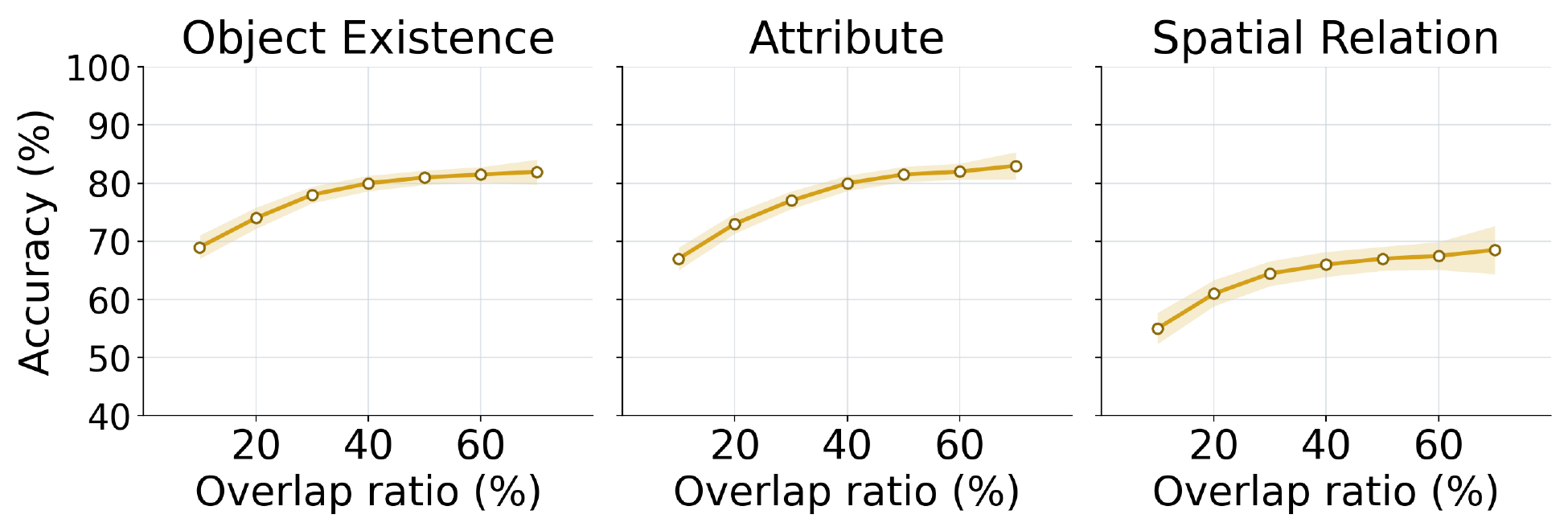}
        \caption*{(a) Accuracy by overlap ratio}
    \end{minipage}
    \hfill
    \begin{minipage}{0.56\textwidth}
        \centering
        \includegraphics[width=\linewidth]{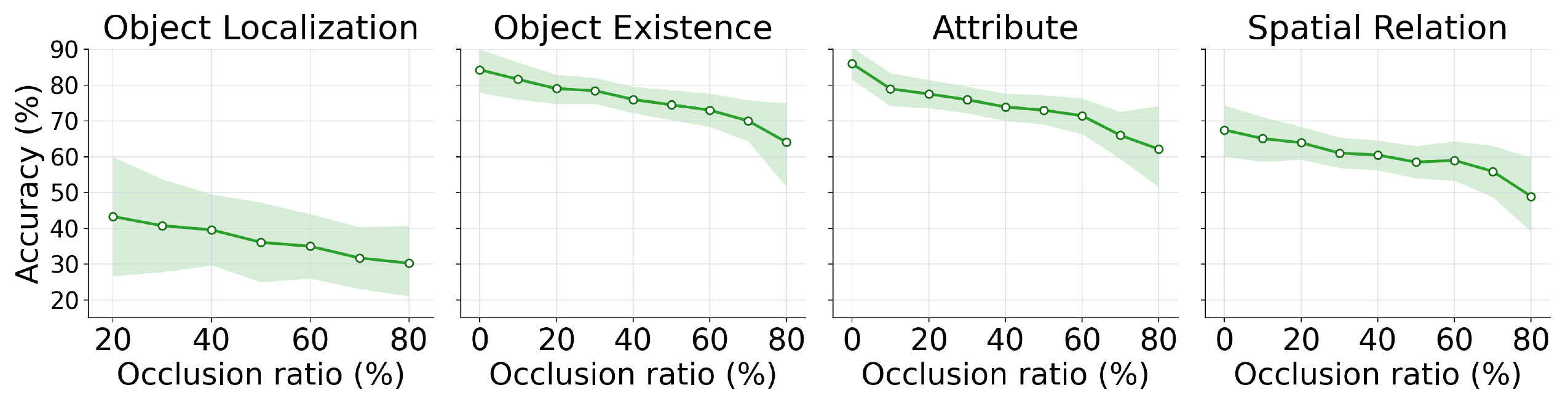}
        \caption*{(b) Accuracy by occlusion ratio}
    \end{minipage}
    \caption{{Quantitative analysis of model accuracy under increasing frame overlap and target occlusion.}}
    \label{fig:analys_occ_ovlap_quant_main}
\end{figure*}

\begin{figure*}[t]
    \centering
    \includegraphics[width=\textwidth]{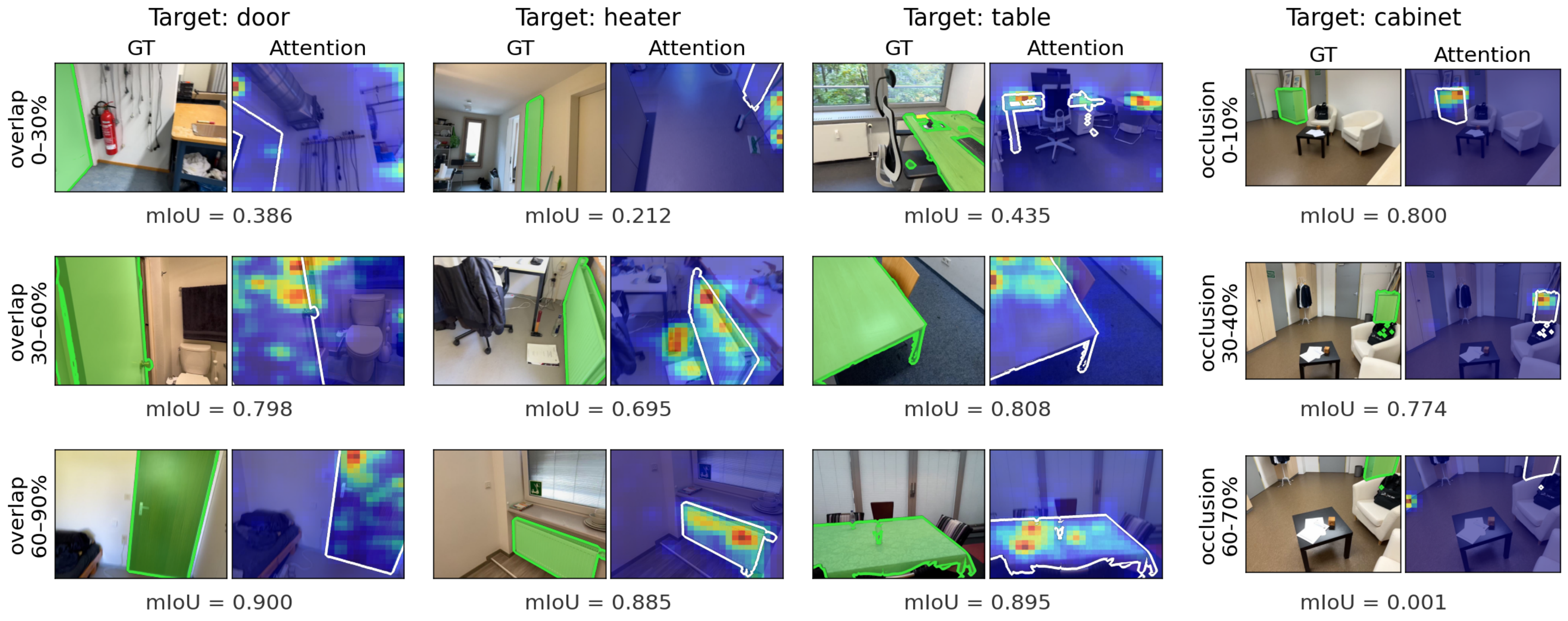}
    \caption{{Qualitative attention under increasing frame overlap (left three columns: door, heater, table) and target occlusion (right column: cabinet).}}
    \label{fig:analys_occ_ovlap_qual_main}
\end{figure*}

\noindent\textbf{Low-overlap.}
In multi-view settings, establishing correspondences across views is essential for aggregating information from different observations.
When the overlap between frames is low, the model has fewer shared visual regions for matching objects across views.
As shown in Fig.~\ref{fig:analys_occ_ovlap_qual_main}, attention is weakly aligned with the target object in low-overlap cases and often shifts toward nearby context or unrelated regions.
With higher overlap, attention becomes more consistent with the target mask, indicating that shared visual regions help the model establish cross-view correspondence.
The quantitative results in Fig.~\ref{fig:analys_occ_ovlap_quant_main} show the same trend, with VQA accuracy increasing as the overlap ratio becomes higher.

\noindent\textbf{Occlusion.}
A further challenge arises when the target object is only partially observable in individual frames.
In such cases, the model has fewer reliable visual cues for identifying the object and relating it to other instances.
Fig.~\ref{fig:analys_occ_ovlap_qual_main} illustrates this effect, with attention well aligned to the target under low occlusion but shifting toward surrounding context or visible distractors as occlusion becomes severe.
The accuracy trend in Fig.~\ref{fig:analys_occ_ovlap_quant_main} follows the same pattern, decreasing as the occlusion ratio increases.
This shows that occlusion weakens target grounding and propagates the resulting uncertainty to downstream spatial inference.

Overall, these analyses show that the proposed conditions are not merely dataset design choices, but measurable sources of difficulty for current VLMs.
Partial visibility forces models to aggregate incomplete observations, low-overlap views make cross-view correspondence less reliable, and occlusion weakens the visual cues needed for object grounding and relation inference.
Together, these results demonstrate that \ourbenchmark{} captures realistic failure modes of multi-view spatial reasoning and provides a fine-grained basis for evaluating models under challenging visual conditions.

\section{Practical Baseline for Multi-view Reasoning}
\label{sec:method}
\setlength{\intextsep}{0pt}
\begin{wrapfigure}{r}{0.5\textwidth}
\centering
\includegraphics[width=0.4\textwidth]{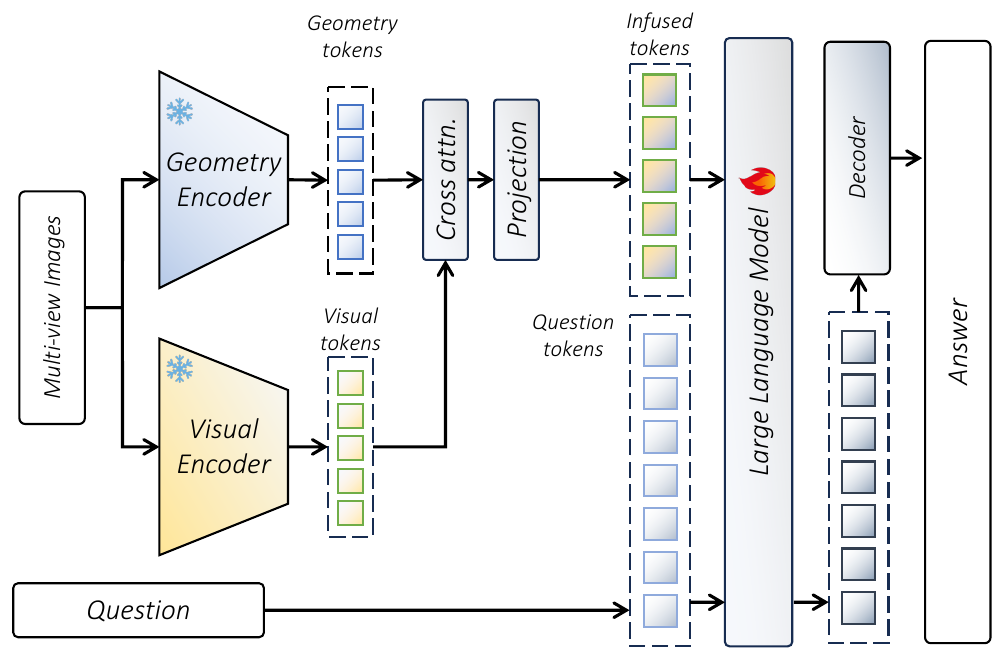}
    \caption{\textbf{Baseline for multi-view VLM.}}
    \label{fig:architecture}
\end{wrapfigure}
To support evaluation on \ourbenchmark{}, we provide a practical baseline that integrates a 3D reconstruction model into a VLM framework. As illustrated in Fig.~\ref{fig:architecture}, we use VGGT~\cite{wang2025vggt} to extract geometry-aware tokens from multi-view images and fuse them with visual tokens from the image encoder, enabling the model to better capture cross-view geometric cues. The fused tokens are then projected into the language model for answer generation. Further architecture details are provided in \App{architecture_details}.
\section{Experiments}
\label{sec:experiments}
\noindent
\textbf{Datasets.}
We evaluate baselines across two complementary spatial benchmarks, each examining a distinct aspect of multi-view spatial reasoning. First, we analyze model performance on \ourbenchmark{}, which consists of two evaluation splits: \textit{SpatialMosaic-Indoor} (Sec.~\ref{sec_eval_ours}) and \textit{SpatialMosaic-Outdoor} (Sec.~\ref{sec_eval_waymo}). 
\textit{SpatialMosaic-Indoor} is constructed from ScanNet++~\cite{scannetpp}, while \textit{SpatialMosaic-Outdoor} is built upon the Waymo dataset~\cite{Sun_2020_CVPR}.
These benchmarks assess robustness in realistic environments spanning both indoor and outdoor scenes, involving partial visibility, occlusion, and low-overlap conditions. \ourbenchmark{} offers a challenging evaluation protocol by constructing view sets with deliberately minimal geometric redundancy.
It is designed to probe robustness under incomplete spatial evidence and to reflect more realistic low-overlap conditions.
Next, we further evaluate on VSI-Bench~\cite{yang2025thinking}, an informative benchmark designed to assess conventional spatial understanding capabilities across spatial configuration and measurement-related tasks.

\noindent
\textbf{Baselines.}
Following prior efforts in spatial reasoning, we evaluate our method against a suite of open-sourced Video-Language Models, which serve as our baselines under identical multi-view input protocols. All baselines~\cite{chen2024far,li2024llava,llavanextvideo,zhang2024long,lin2024vila, bai2025qwen3, wang2025internvl3} receive the same set of multi-view frames at a resolution of 518 $\times$ 518 per view, ensuring consistent evaluation across architectures. Model outputs are derived using their default decoding settings without task-specific tuning.

\begin{table*}[t]
\centering
\setlength{\tabcolsep}{6pt}
\caption{\textbf{Quantitative results on SpatialMosaic-Indoor.} \textbf{Bold} and \underline{underline} indicate the best and second-best performance within open-sourced VLMs for each task, respectively. \colorbox{oai-green-600}{Highlighting} denotes the top-3 ranked models overall.}
\label{tab:eval_ours}
\resizebox{\textwidth}{!}{
\begin{tabular}{l|c|c|cccccc|ccccc}
\toprule
 & & & \rotatebox{75}{Obj. Count} & \rotatebox{75}{Best View.} & \rotatebox{75}{Obj. Exist.} & \rotatebox{75}{Obj. Att.} & \rotatebox{75}{Obj. Rel.} & \rotatebox{75}{Obj. Loc.} & \rotatebox{75}{Obj. Count} & \rotatebox{75}{Cam-Obj Dist.} & \rotatebox{75}{Obj-Obj Dist.} & \rotatebox{75}{Obj. Size} & \rotatebox{75}{Obj. Loc.} \\
\textbf{Methods} & {Rank} & {Avg.} &
\multicolumn{6}{c|}{\cellcolor{yellow!10}Multiple-Choice Answer} &
\multicolumn{5}{c}{\cellcolor{orange!10}Numerical Answer} \\
\midrule
\multicolumn{14}{l}
{\cellcolor{navyblue!5}\textit{SpatialMosaic-tiny Perf.}} \\
Human & - & 46.9 & 70.0 & 40.0 & 66.6 & 41.1 & 50.0 & 63.3 & 70.0 & 25.0 & 22.2 & 42.5 & 25.5 \\
LLaVA-NeXT-Video & - & 40.0 & 70.0 & 50.0 & 60.0 & 24.4 & 58.3 & 43.3 & 57.5 & 14.2 & 14.5 & 32.8 & 15.0 \\
\midrule
\multicolumn{14}{l}{\cellcolor{navyblue!5}\textit{Open-sourced VLMs}} \\
LLaVA-OneVision-0.5B & 13 & 34.0 & 29.5 & 44.4 & 55.3 & 20.7 & 36.3 & 38.3 & 44.4 & 30.9 & 25.9 & 11.5 & 17.0 \\
InternVL2-2B & 11 & 36.7 & 65.8 & 49.4 & 47.8 & 26.4 & 36.9 & \underline{53.1} & 53.6 & 13.7 & 14.1 & 10.8 & \underline{30.6} \\
\midrule
LLaVA-OneVision-7B & 8 & 38.6 & 58.5 & 37.5 & 56.1 & 32.6 & 40.3 & 37.8 & 52.7 & 12.1 & 16.5 & 37.1 & 17.8 \\
LongVA-7B & 14 & 32.9 & 34.5 & 27.5 & 57.4 & 24.2 & 41.3 & 26.2 & 36.1 & 15.6 & 11.0 & 22.2 & 16.9 \\
InternVL2-8B & 5 & 42.6 & 61.6 & 49.0 & 54.4 & 38.6 & 42.5 & 43.3 & 58.9 & 26.9 & 24.1 & 35.8 & 17.9 \\
VILA-1.5-8B & 12 & 34.8 & 40.2 & 24.7 & 52.5 & 32.5 & 39.0 & 32.4 & 61.1 & 2.4 & 2.0 & 40.5 & 16.8 \\
VILA-1.5-40B & 4 & 43.1 & 56.0 & \underline{50.7} & 58.9 & 31.6 & 48.7 & 40.6 & \underline{66.0} & 29.2 & 19.8 & 36.3 & 16.9 \\
Qwen3-VL-2B & 7 & 39.4 & 41.5 & 35.0 & 54.6 & 35.6 & 40.2 & 38.6 & 35.2 & 20.5 & 32.2 & 50.5 & 24.4 \\
Qwen3-VL-4B & \cellcolor{oai-green-200}3 & 46.5 & 62.2 & 46.0 & 56.2 & \underline{48.4} & 46.6 & 36.5 & 51.6 & 29.7 & \underline{37.3} & \underline{57.3} & 16.3 \\
Qwen3-VL-8B & \cellcolor{oai-green-400}2 & \underline{48.4} & \underline{66.8} & 45.4 & 60.0 & 48.2 & \underline{53.8} & 37.5 & 49.2 & 29.3 & 35.5 & 57.2 & 16.7 \\
InternVL3.5-2B & 10 & 37.4 & 52.6 & 39.8 & 54.7 & 30.9 & 40.2 & 38.7 & 19.6 & 17.3 & 24.0 & 43.1 & 22.9 \\
InternVL3.5-8B & 6 & 42.4 & 54.2 & 36.9 & \underline{60.3} & 36.9 & 46.4 & 35.9 & 24.3 & \underline{33.4} & 33.6 & 52.6 & 19.5 \\
\midrule
LLaVA-NeXT-Video-7B & 9 & 38.2 & 61.1 & 41.0 & 45.7 & 34.2 & 52.3 & 37.3 & 51.3 & 11.7 & 12.5 & 20.9 & 16.9 \\
LLaVA-NeXT-Video-7B + SpatialMosaic & \cellcolor{oai-green-600}1 & \textbf{67.6} & \textbf{87.3} & \textbf{72.2} & \textbf{78.5} & \textbf{69.9} & \textbf{77.1} & \textbf{61.0} & \textbf{68.8} & \textbf{38.3} & \textbf{38.8} & \textbf{63.9} & \textbf{42.4} \\
\bottomrule
\end{tabular}
}
\end{table*}

\subsection{Evaluation on SpatialMosaic-Indoor}
\label{sec_eval_ours}
\ourbenchmark{} provides a challenging and realistic evaluation characterized by partial visibility, occlusion, and minimal overlap across viewpoints. These conditions limit geometric redundancy and require models to infer spatial structure from fragmented observations rather than relying on stable cross-frame correspondences. Despite their strong spatial reasoning ability in conventional image or video settings, existing MLLM baselines struggle under these conditions. 
In Tab.~\ref{tab:eval_ours}, the results highlight the difficulty of integrating fragmented visual evidence across views when spatial cues are incomplete or partially observed.
Additionally, we evaluate the most challenging split using SpatialMosaic-tiny, a reduced version of the full benchmark consisting of 300 randomly selected questions, to benchmark against human performance.

\noindent

\begin{table*}[t!]
\centering
\small
\setlength{\tabcolsep}{10pt}
\renewcommand{\arraystretch}{0.88}
\caption{\textbf{Quantitative results on VSI-Bench.} Comparison with strong baselines across diverse spatial reasoning tasks.
The model combined with VGGT achieves the best overall performance, outperforming 72B-scale models while using significantly fewer parameters.}
\resizebox{\textwidth}{!}{
\begin{tabular}{l|c|c|cccccccc}
\toprule
 & & & \rotatebox{75}{Obj. Count} & \rotatebox{75}{Abs. Dist.} & \rotatebox{75}{Obj. Size} & \rotatebox{75}{Room Size} & \rotatebox{75}{Rel. Dist.} & \rotatebox{75}{Rel. Dir.} & \rotatebox{75}{Route Plan} & \rotatebox{75}{Appr. Order} \\
\textbf{Methods} & Rank & {Avg.}&
\multicolumn{4}{c}{\cellcolor{orange!10}Numerical Answer} &
\multicolumn{4}{c}{\cellcolor{yellow!10}Multiple-Choice Answer} \\
\midrule
\multicolumn{11}{l}{\cellcolor{navyblue!5}\textit{Proprietary Models (API)}} \\
GPT-4o & - & 34.0 & 46.2 & 5.3 & 43.8 & 38.2 & 37.0 & 41.3 & 31.5 & 28.5 \\
Gemini-1.5 Flash & - & 42.1 & 49.8 & 30.8 & 53.5 & 54.4 & 37.7 & 41.0 & 31.5 & 37.8 \\
Gemini-1.5 Pro & - & 45.4 & 56.2 & 30.9 & 64.1 & 43.6 & 51.3 & 46.3 & 36.0 & 34.6 \\
\midrule
\multicolumn{11}{l}{\cellcolor{navyblue!5}\textit{Open-sourced VLMs}} \\
LLaVA-OneVision-0.5B & 11 & 28.0 & 46.1 & 28.4 & 15.4 & 28.3 & 28.9 & 36.9 & 34.5 & 5.8 \\
InternVL2-2B & 12 & 27.4 & 21.8 & 24.9 & 22.0 & 35.0 & 33.8 & \underline{44.2} & 30.5 & 7.1 \\
\midrule
InternVL2-8B & 6 & 34.6 & 23.1 & \underline{28.7} & 48.2 & \underline{39.8} & 36.7 & 30.7 & 29.9 & 39.6 \\
LLaVA-OneVision-7B & 7 & 32.4 & 47.7 & 20.2 & 47.4 & 12.3 & 42.5 & 35.2 & 29.4 & 24.4 \\
LongVA-7B & 9 & 29.2 & 38.0 & 16.6 & 38.9 & 22.2 & 33.1 & 43.3 & 25.4 & 15.7 \\
VILA-1.5-8B & 10 & 28.9 & 17.4 & 21.8 & 50.3 & 18.8 & 32.1 & 34.8 & 31.0 & 24.8 \\
LongVILA-8B & 13 & 21.6 & 29.1 & 9.1 & 16.7 & 0.0 & 29.6 & 30.7 & 32.5 & 25.5 \\
\midrule
InternVL2-40B & 4 & 36.0 & 34.9 & 26.9 & 46.5 & 31.8 & 42.1 & 32.2 & 34.0 & 39.6 \\
VILA-1.5-40B & 8 & 31.2 & 22.4 & 24.8 & 48.7 & 22.7 & 40.5 & 25.7 & 31.5 & 32.9 \\
LLaVA-NeXT-Video-72B & \cellcolor{oai-green-400}2 & \underline{40.9} & \underline{48.9} & 22.8 & 57.4 & 35.3 & 42.4 & 36.7 & \underline{35.0} & \textbf{48.6} \\
LLaVA-OneVision-72B & \cellcolor{oai-green-200}3 & 40.2 & 43.5 & 23.9 & \underline{57.6} & 37.5 & 42.5 & 39.9 & 32.5 & \underline{44.6} \\
\midrule
LLaVA-NeXT-Video-7B & 5 & 35.6 & 48.5 & 14.0 & 47.8 & 24.2 & \underline{43.5} & 42.4 & 34.0 & 30.6 \\
LLaVA-NeXT-Video-7B + VGGT & \cellcolor{oai-green-600}1 
& \textbf{59.6} & \textbf{70.6} & \textbf{48.6} & \textbf{70.1} 
& \textbf{65.2} & \textbf{60.4} & \textbf{77.5} & \textbf{43.8} & 40.5\\
\bottomrule
\end{tabular}
}
\label{tab:vsi_bench}
\vspace{-3mm}
\end{table*}
\begin{table*}[t]
\centering
\small
\setlength{\tabcolsep}{6pt}
\renewcommand{\arraystretch}{0.95}
\caption{\textbf{Quantitative results on SpatialMosaic-Outdoor.} Zero-shot evaluation on outdoor scenes constructed from the Waymo dataset. \textbf{Bold} and \underline{underline} indicate the best and second-best performance within open-sourced VLMs for each task, respectively.}
\label{tab:eval_ours_waymo}
\resizebox{\textwidth}{!}{
\begin{tabular}{l|c|c|ccccc|cccc}
\toprule
 & & & \rotatebox{75}{Obj. Count} & \rotatebox{75}{Best View.} & \rotatebox{75}{Obj. Exist.} & \rotatebox{75}{Obj. Rel.} & \rotatebox{75}{Obj. Loc.} & \rotatebox{75}{Obj. Count} & \rotatebox{75}{Cam-Obj Dist.} & \rotatebox{75}{Obj-Obj Dist.} & \rotatebox{75}{Obj. Loc.} \\
\textbf{Methods} & {Rank} & {Avg.} &
\multicolumn{5}{c|}{\cellcolor{yellow!10}Multiple-Choice Answer} &
\multicolumn{4}{c}{\cellcolor{orange!10}Numerical Answer} \\
\midrule
\multicolumn{12}{l}{\cellcolor{navyblue!5}\textit{Open-sourced VLMs}} \\
LLaVA-NeXT-Video-7B & 4 & 46.2 & 54.7 & 40.2 & 59.4 & 53.5 & 37.5 & \underline{84.0} & 27.6 & 8.2 & 9.3 \\
LLaVA-OneVision-7B & 8 & 37.0 & 26.4 & 22.6 & 57.3 & 35.4 & 31.8 & 74.1 & \underline{28.8} & \underline{9.5} & 9.3 \\
LongVA-7B & 9 & 34.8 & 38.3 & 23.6 & 44.3 & 45.5 & 29.2 & 53.1 & 22.3 & 7.4 & 9.3 \\
InternVL2-8B & 6 & 40.0 & 45.3 & 43.6 & 49.7 & 49.0 & 39.0 & 60.3 & 21.3 & 4.5 & 9.7 \\
VILA-1.5-8B & 10 & 33.8 & 40.8 & 22.9 & 50.6 & 29.5 & 34.3 & 81.3 & 9.9 & 0.1 & 9.3 \\
VILA-1.5-40B & \cellcolor{oai-green-200}3 & 46.6 & 41.9 & 43.8 & 60.0 & 59.0 & 37.0 & 77.9 & \textbf{30.0} & 9.3 & 9.3 \\
Qwen3-VL-8B & 5 & 42.6 & 48.8 & 43.5 & 53.9 & 56.9 & 34.2 & 61.1 & 17.6 & 7.2 & 8.7 \\
InternVL3.5-8B & 7 & 37.3 & 42.6 & 35.3 & 57.8 & 39.7 & 33.0 & 47.3 & 14.8 & 4.0 & 15.1 \\
\midrule
LLaVA-NeXT-Video-7B + SpatialMosaic & \cellcolor{oai-green-400}2 & \underline{56.9} & \underline{77.9} & \textbf{67.2} & \underline{73.8} & \textbf{65.3} & \textbf{58.0} & \textbf{85.3} & 7.7 & 7.9 & \textbf{18.1} \\
LLaVA-NeXT-Video-7B + VGGT + SpatialMosaic & \cellcolor{oai-green-600}1 & \textbf{57.7} & \textbf{79.5} & \underline{67.1} & \textbf{76.8} & \textbf{65.3} & \underline{57.4} & 81.5 & 12.9 & \textbf{10.1} & \underline{16.0} \\
\bottomrule
\end{tabular}
}
\vspace{-4mm}
\end{table*}

\vspace{-1.0em}
\subsection{Evaluation on VSI-Bench}
\label{sec_eval_vsi}
To further examine conventional spatial reasoning capability beyond the task formulations defined in \ourbenchmark{}, we conduct evaluations on VSI-Bench. Unlike our benchmarks, which emphasize multi-view reasoning under challenging conditions, VSI-Bench primarily focuses on standard spatial reasoning tasks, such as relative distance estimation, directional reasoning, and object size comparison. 
As shown in Tab.~\ref{tab:vsi_bench}, the baseline model combined with VGGT achieves the best overall performance among all evaluated models, substantially outperforming strong proprietary and open-source VLM baselines. Notably, the practical baseline model outperforms 72B-scale model~\cite{llavanextvideo} by 18.7\%, despite using nearly 10$\times$ fewer parameters. The performance gains are consistent across diverse task categories, indicating that it maintains strong multi-view reasoning capability across various spatial reasoning tasks.
Although VSI-Bench does not explicitly focus on occlusion-heavy or low-overlap scenarios, the proposed baseline model maintains strong performance across its task categories, demonstrating strong spatial reasoning capability across both challenging multi-view settings and more conventional benchmarks. 

\subsection{Evaluation on SpatialMosaic-Outdoor}
\label{sec_eval_waymo}
Existing VLM benchmarks are often confined to a single environmental domain, typically focusing exclusively on either indoor layouts or outdoor scenes. Such domain-specific evaluation settings limit the ability to assess whether a model has acquired transferable multi-view spatial reasoning capabilities, as performance may be influenced by recurring structural patterns inherent to a particular layout distribution. 
To address this limitation, we extend our evaluation to encompass both indoor and outdoor environments. In particular, we evaluate \textit{SpatialMosaic-Outdoor} in a zero-shot setting, without fine-tuning on outdoor data, to examine out-of-domain generalization. This benchmark is constructed using the large-scale Waymo dataset~\cite{Sun_2020_CVPR}, introducing complex outdoor driving scenarios with substantially different scene geometry and spatial configurations compared to indoor environments. 
Despite this domain shift, the finetuned models maintain strong performance across task categories. These results indicate that the learned spatial representations transfer effectively beyond the indoor domain, demonstrating robust out-of-domain spatial reasoning capability. Overall, the consistent performance across both \textit{SpatialMosaic-Indoor} and \textit{SpatialMosaic-Outdoor} suggests that the model combined with VGGT mitigates layout-specific bias and supports reliable zero-shot generalization across heterogeneous environments.
\subsection{Ablation Studies}
\label{sec:ablation_studies}
We conduct ablation studies to evaluate the efficiency of the proposed baseline, which serves as a new baseline for the multi-view setting. For a fair comparison, all ablated models are fine-tuned using the same training data and follow identical optimization and decoding settings. Evaluations are conducted on \ourbenchmark{}, and VSI-Bench~\cite{yang2025thinking} under the same protocol. For each source dataset, we follow its original train/test split when constructing the training and evaluation sets.
As shown in Table~\ref{tab:ablation_merged}, removing geometric features consistently degrades performance across both \textit{SpatialMosaic-Bench} and VSI-Bench~\cite{yang2025thinking}.
The performance drop is particularly noticeable in tasks that require explicit spatial reasoning, such as object attribute understanding, spatial relations, and directional reasoning. These tasks depend heavily on geometric cues for resolving spatial configuration across views. Without the geometry encoder, the model struggles to accurately infer object properties and spatial relationships from multi-view observations.
Overall, these results demonstrate that incorporating explicit geometric representations significantly improves the model’s ability to reason about spatial structure, highlighting the importance of geometric cues for robust multi-view spatial understanding.

\begin{table*}[t!]
\centering
\scriptsize
\caption{\textbf{Ablation studies for geometry encoder on \ourbenchmark{} and VSI-Bench.}}
\label{tab:ablation_merged}
\vspace{-0.5em}

\setlength{\tabcolsep}{2.2pt}
\renewcommand{\arraystretch}{1.05}

\resizebox{\textwidth}{!}{
\begin{tabular}{
l|cccccccccccc
@{\hspace{1.5em}}
|ccccccccc
}
\toprule
&
\multicolumn{12}{c@{\hspace{1.5em}}|}{\cellcolor{navyblue!5}\textbf{\ourbenchmark{}}}
&
\multicolumn{9}{c}{\cellcolor{navyblue!5}\textbf{VSI-Bench}}
\\
\cmidrule(r{1.0em}){2-13}
\cmidrule(l{1.0em}){14-22}

&
&
\rotatebox{75}{Obj. Count}
& \rotatebox{75}{Best View}
& \rotatebox{75}{Obj. Exist.}
& \rotatebox{75}{Obj. Attr.}
& \rotatebox{75}{Obj. Rel.}
& \rotatebox{75}{Obj. Loc.}
& \rotatebox{75}{Obj. Count}
& \rotatebox{75}{Cam-Obj Dist.}
& \rotatebox{75}{Obj-Obj Dist.}
& \rotatebox{75}{Obj. Size}
& \rotatebox{75}{Obj. Loc.}
&
&
\rotatebox{75}{Obj. Count}
& \rotatebox{75}{Abs. Dist.}
& \rotatebox{75}{Obj. Size}
& \rotatebox{75}{Room Size}
& \rotatebox{75}{Rel. Dist.}
& \rotatebox{75}{Rel. Dir.}
& \rotatebox{75}{Route Plan}
& \rotatebox{75}{Appr. Order}
\\

\textbf{Methods}
&
\textbf{Avg.}
&
\multicolumn{6}{c}{\cellcolor{yellow!10}Multiple-Choice Answer}
&
\multicolumn{5}{c@{\hspace{1.5em}}|}{\cellcolor{orange!10}Numerical Answer}
&
\textbf{Avg.}
&
\multicolumn{4}{c}{\cellcolor{orange!10}Numerical Answer}
&
\multicolumn{4}{c}{\cellcolor{yellow!10}Multiple-Choice Answer}
\\
\midrule

LLaVA-NeXT-Video-7B
& 67.6
& 87.3
& \textbf{72.2}
& 78.5
& 69.9
& 77.1
& 61.0
& 68.8
& 38.3
& \textbf{38.8}
& \textbf{63.9}
& \textbf{42.4}
&
54.8
& 69.7
& 44.4
& 70.0
& 60.3
& 57.3
& 55.4
& 39.2
& \textbf{42.1}
\\

LLaVA-NeXT-Video-7B + VGGT
& \textbf{69.2}
& \textbf{89.0}
& \textbf{72.2}
& \textbf{80.1}
& \textbf{72.9}
& \textbf{79.4}
& \textbf{61.4}
& \textbf{73.8}
& \textbf{39.8}
& 38.5
& \textbf{63.9}
& 39.7
&
\textbf{59.6}
& \textbf{70.6}
& \textbf{48.6}
& \textbf{70.1}
& \textbf{65.2}
& \textbf{60.4}
& \textbf{77.5}
& \textbf{43.8}
& 40.5
\\

\bottomrule
\end{tabular}
}

\vspace{-0.8em}
\end{table*}
\vspace{-3mm}
\section{Conclusion}
In this work, we take an initial step toward tackling the challenges of partial visibility, occlusion, and low-overlap conditions in settings that require models to integrate fragmented visual cues to form coherent 3D understanding. To support this goal, we introduced an automatic multi-view data generation pipeline, enabling the construction of \ourdataset{} and \ourbenchmark{}, which capture challenging multi-view scenarios across indoor and outdoor scenes. We further introduce a practical baseline for multi-view settings, a hybrid framework that combines geometric cues from 3D reconstruction models to enable effective cross-view alignment and robust spatial reasoning. Experiments demonstrate that instruction-tuning on our dataset improves performance under incomplete visual evidence. These results highlight the importance of equipping models with the ability to aggregate partial observations and infer coherent 3D structure from limited cues. We believe that this work enhances the scalability and real-world applicability of MLLMs, contributing to narrowing the gap toward human-level multi-view reasoning.

\noindent\textbf{Limitation.}
Although our benchmark includes numerical-answer questions, it does not fully capture the various complexities of open-ended real-world environments. In addition, our VQA generation is based on a limited set of source datasets, and extending it to more diverse datasets could further improve the coverage and generality of the benchmark.


{\small
\bibliographystyle{plainnat}
\bibliography{main}
}
\appendix
%
%

\clearpage
\setcounter{page}{1}

\appendix
\section{Analysis}
\label{sec:analysis}
To better understand how different challenging factors affect multi-view spatial reasoning, we conduct a detailed analysis across a range of controlled conditions. 
We first examine general challenges that arise in multi-view settings, including small object scale and duplicate instances. We then analyze the realistic conditions emphasized in \ourbenchmark{}, including partial visibility, low-overlap views, and occlusion. 
By analyzing both perception-level grounding and final VQA accuracy, we identify when and why current MLLMs struggle to integrate visual evidence across views.
Our analysis shows that, under realistic and challenging multi-view conditions, current VLMs often fail to perform reliable cross-view alignment, instance correspondence, and multi-hop evidence aggregation.

\subsection{Challenges in Multi-View Reasoning}
\label{subsec:analysis_multiview_challenges}

\begin{figure*}[h!]
\centering
\includegraphics[width=0.9\textwidth]{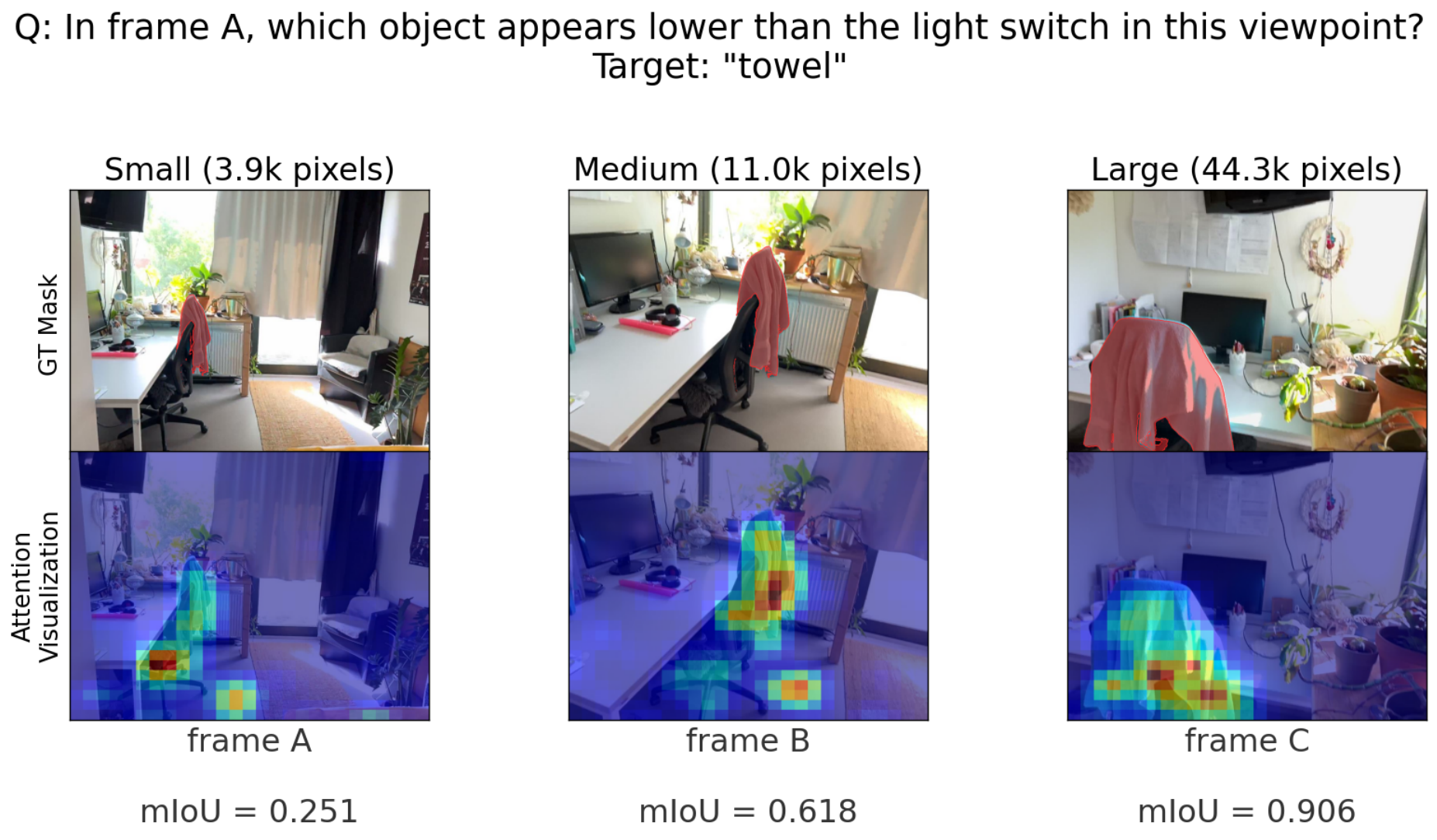}
    \caption{Qualitative analysis of object grounding under varying visible-pixel counts.}

    \label{fig:vispix_attn_qual}
\end{figure*}

\noindent\textbf{Effect of object scale.}
We first analyze how object scale affects multi-view reasoning by varying the number of visible pixels for the target instance.
Although multi-view reasoning is often discussed as a cross-frame aggregation problem, the model must first perceive and ground each relevant object within individual frames.
As shown in Fig.~\ref{fig:vispix_attn_qual}, when the target object occupies only a small image region, the model's attention is often diffuse or shifted toward nearby context.
As the number of visible pixels increases, attention becomes more concentrated on the target mask, leading to substantially higher grounding quality.

\begin{figure*}[h!]
\centering
\includegraphics[width=0.8\textwidth]{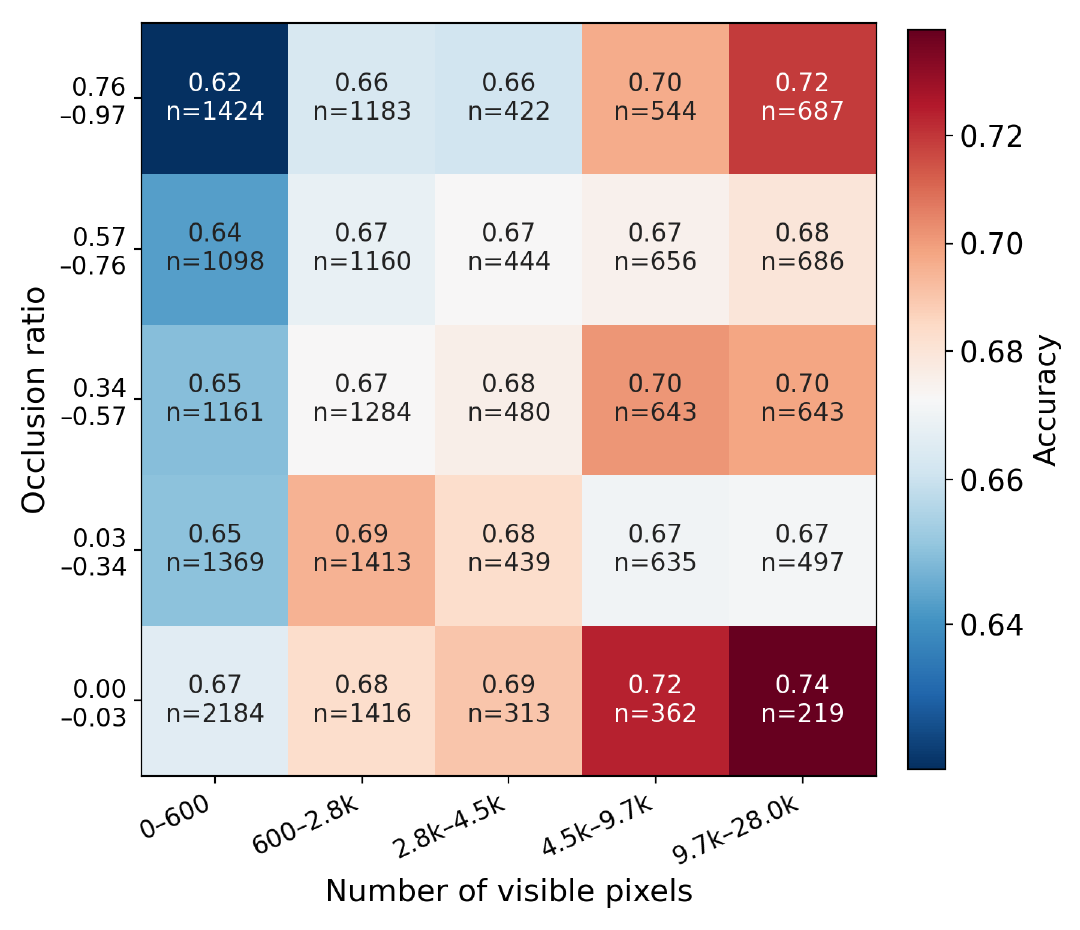}
    \caption{VQA accuracy by target-instance visible pixels.}

    \label{fig:vis_pix_occ_acc_quant}
\end{figure*}

We further quantify this trend using both perception-level grounding and final VQA accuracy.
Fig.~\ref{fig:vis_pix_miou_quant} demonstrates that grounding mIoU generally increases with the number of visible pixels across task categories, indicating that larger object scale leads to more accurate object grounding.
For final VQA accuracy, we use a joint analysis over visible pixels and occlusion ratio because the two factors are closely related.
Fig.~\ref{fig:vis_pix_occ_acc_quant} shows that performance generally improves with more visible pixels, while high occlusion ratio still makes examples difficult.
This indicates that even when the target object occupies a large image region, reasoning can fail if the object is only partially observable due to occlusion.

\begin{figure*}[h!]
\centering

\input{figure_tex/anlys_1_1_A_1_vis_pix_miou_bar_quant.tex}
    
    \caption{Grounding mIoU by target-instance visible pixels.}

    \label{fig:vis_pix_miou_quant}
\end{figure*}

\begin{figure*}[h!]
\centering
\includegraphics[width=0.99\textwidth]{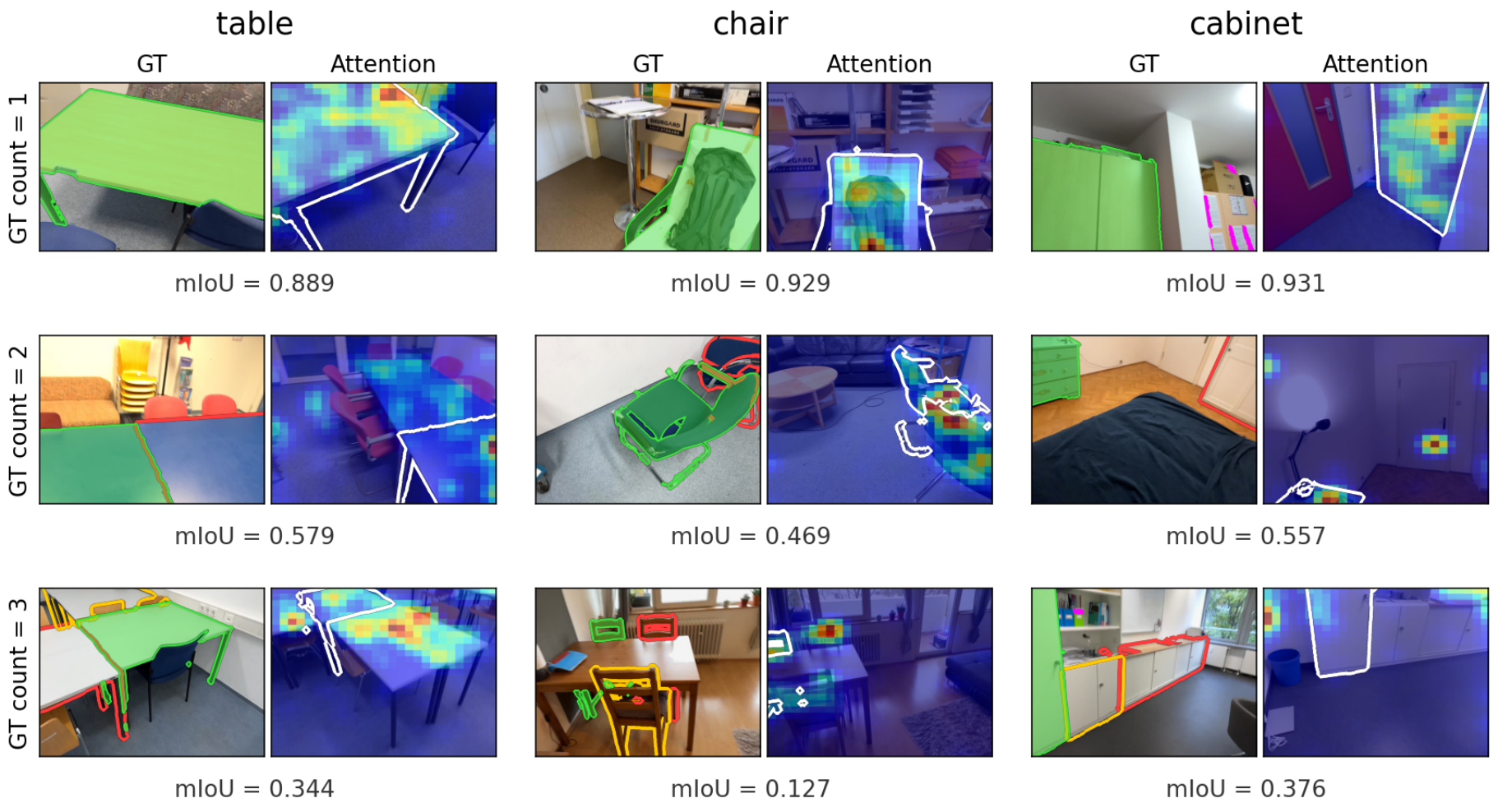}
    \caption{Cross-view correspondence under duplicate visible instances. As the number of same-category instances (GT count) grows, the focus instance's cross-view correspondence (A-IoU) weakens.}
    \label{fig:dup_grid_qual}
\end{figure*}

\noindent\textbf{Effect of duplicate instances.}
We next examine how duplicate instances affect cross-view alignment.                                                                                                                                                                                             
In multi-view settings, multiple objects from the same semantic category can appear across different frames, making instance-level correspondence ambiguous.
Thus, the challenge lies in matching the same physical object across views, rather than merely recognizing its semantic category.
Fig.~\ref{fig:dup_grid_qual} visualizes this correspondence ambiguity.
When a single instance of the target category is present, attention is well aligned with the target instance.
As the number of duplicate instances increases, however, attention increasingly shifts to other instances from the same category or becomes distributed across multiple candidates.
This suggests that the model can localize objects of the target category within individual frames, but struggles to maintain correspondence to the same physical object across views.

\begin{figure*}[t]
    \centering
    \begin{minipage}{0.49\textwidth}
        \centering
        \includegraphics[width=\linewidth]{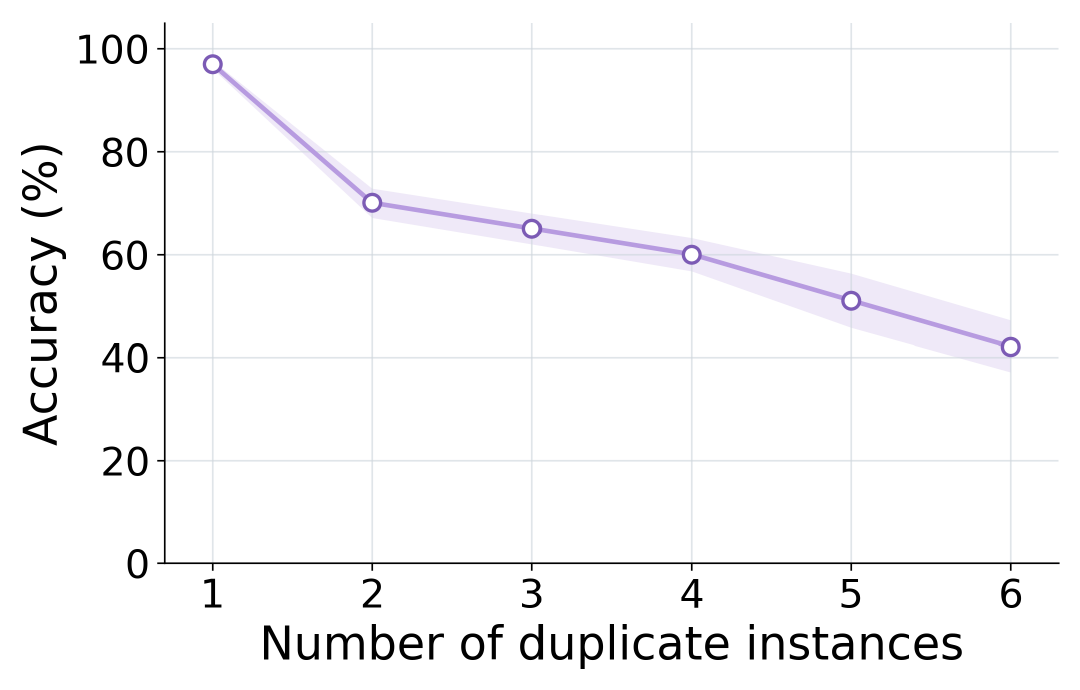}
        \caption*{(a) Accuracy by duplicate instance count}
    \end{minipage}
    \hfill
    \begin{minipage}{0.49\textwidth}
        \centering
        \includegraphics[width=\linewidth]{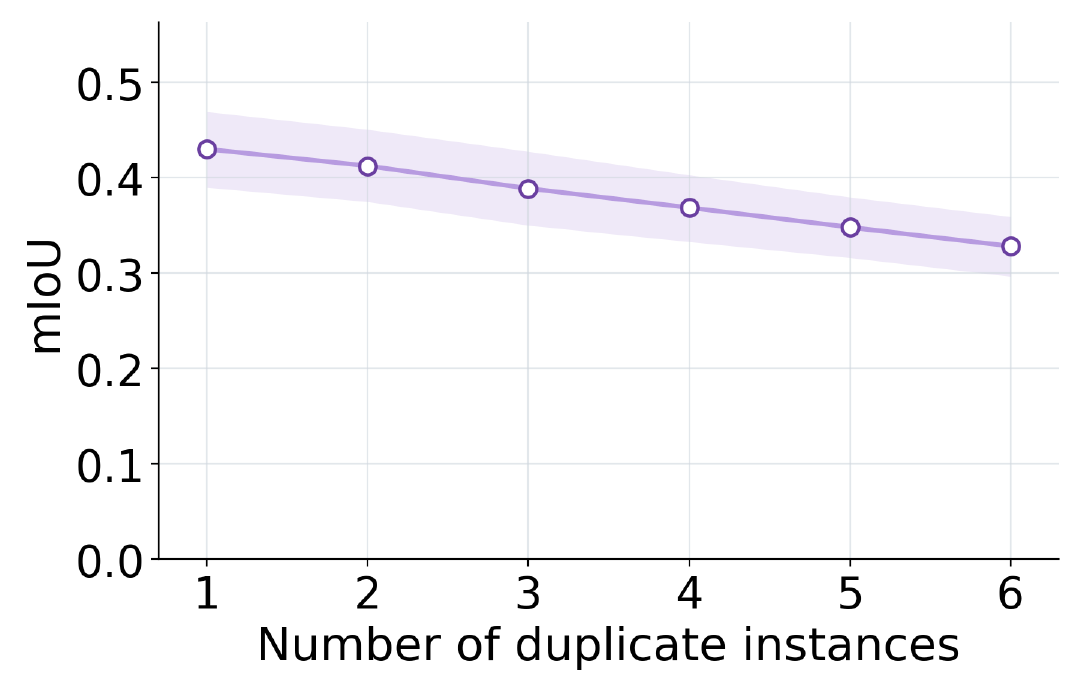}
        \caption*{(b) mIoU by duplicate instance count}
    \end{minipage}
    \caption{Quantitative analysis of duplicate instances in the object count task.}
    \label{fig:count_acc_miou_quant}
\end{figure*}

We further quantify this effect on the object counting task in Fig.~\ref{fig:count_acc_miou_quant}.
Both counting accuracy and grounding mIoU decrease as the scene contains more duplicate instances.
The quantitative results highlight a key limitation of current VLMs in spatial reasoning when same-category objects repeatedly appear across sparse views.

\subsection{Realistic Conditions in SpatialMosaic}
\label{subsec:analysis_realistic_conditions}

\noindent\textbf{Partial visibility.}
\ourbenchmark{} explicitly evaluates settings where the evidence required to answer a question is only partially available in any single view.
We construct this setting by enforcing a partial-visibility constraint in which the source instance is visible in the query frame while the target instance is not, requiring the model to infer their spatial relationship from other reference frames.
To assess whether this condition increases difficulty, we compare the original partially visible setting with an easier setting in which the source and target instances are both visible in the query frame.

\begin{figure*}[h!]
\centering
\includegraphics[width=0.9\textwidth]{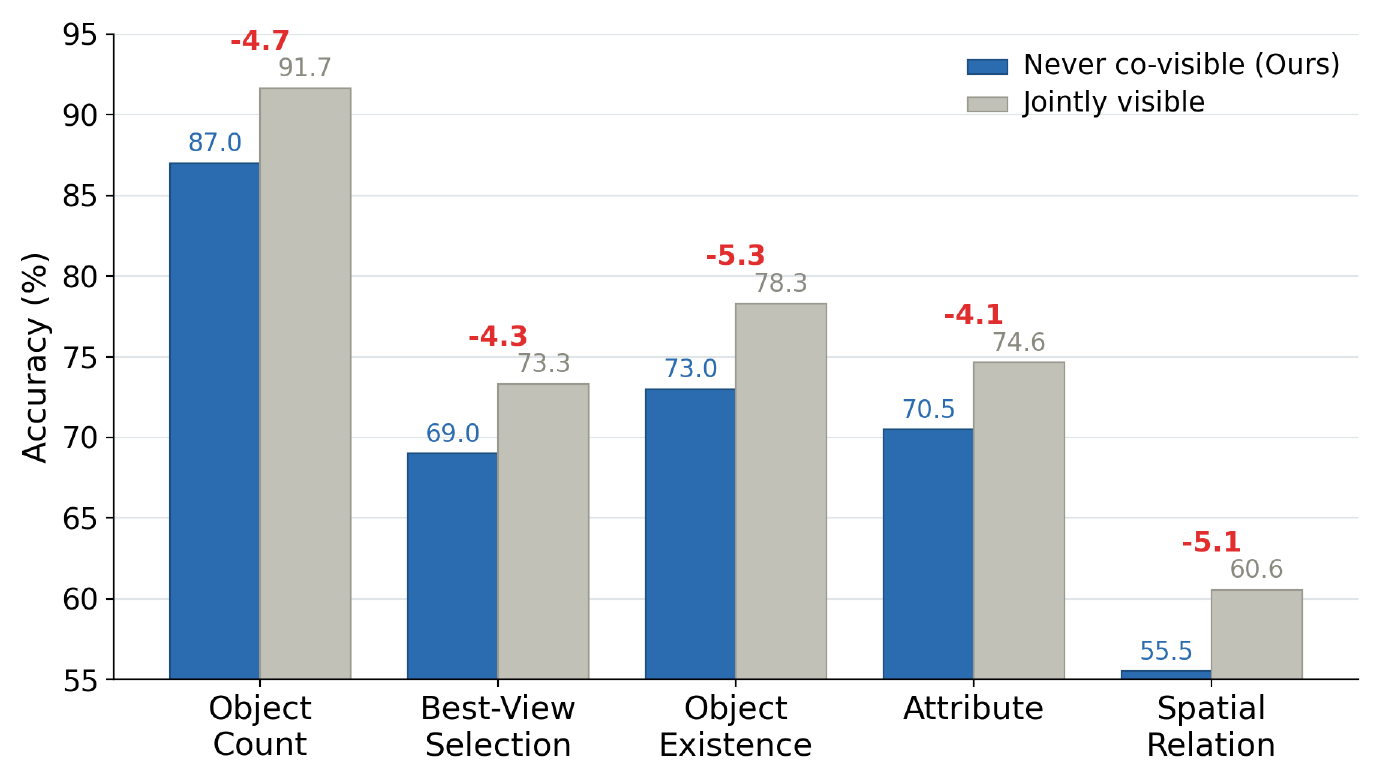}
    \caption{Effect of shared visibility on VQA accuracy.}

    \label{fig:shared_vis_quant}
\end{figure*}

As shown in Fig.~\ref{fig:shared_vis_quant}, making the source and target jointly visible consistently improves accuracy across all task categories.
This indicates that co-visibility provides a strong shortcut for spatial reasoning, as the model can compare the two instances within a single frame without resolving the target from other views.
In contrast, the never co-visible setting requires the model to identify the target from other frames and infer the relation by aligning and aggregating partial observations across views.
The consistent performance gap confirms that partial visibility is a challenging condition that substantially increases the burden of cross-view integration.

\begin{figure*}[h!]
\centering
\includegraphics[width=0.99\textwidth]{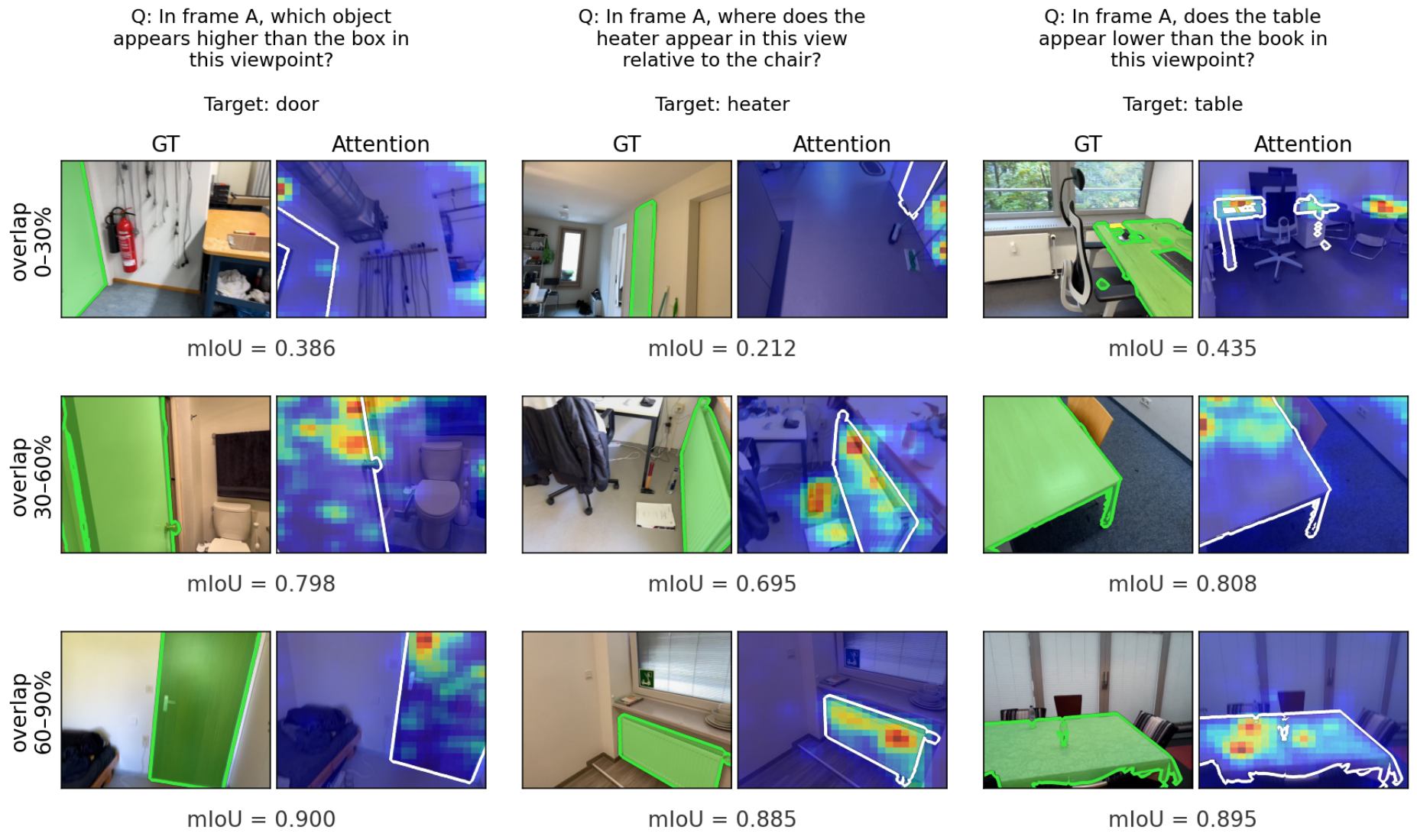}
    \caption{Cross-view correspondence versus frame overlap. As the overlap between the two frames increases, the target's cross-view correspondence (A-IoU) strengthens.}
    \label{fig:overlap_grid_qual}
\end{figure*}

\noindent\textbf{Low-overlap.}
We also analyze how the overlap ratio between frames affects cross-view reasoning.
Low-overlap views reduce shared visual evidence across frames, making it difficult to establish correspondences.
As shown in Fig.~\ref{fig:overlap_grid_qual}, attention is weakly aligned with the target object under low-overlap settings, often shifting to nearby context or unrelated regions.
As overlap ratio increases, attention becomes more consistent with the target mask, indicating that shared visual content helps the model establish cross-view correspondence.

\begin{figure*}[h!]
\centering
\includegraphics[width=0.9\textwidth]{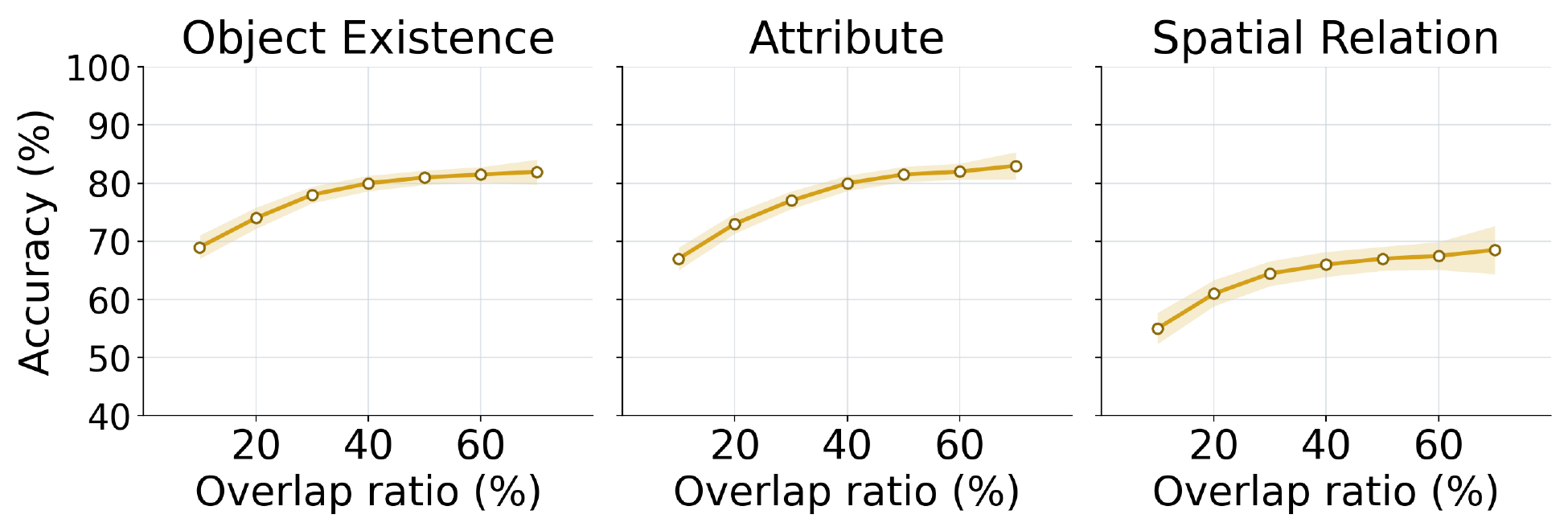}
    \caption{VQA accuracy by overlap ratio.}

    \label{fig:overlap_acc_quant}
\end{figure*}

Fig.~\ref{fig:overlap_acc_quant} further shows that VQA accuracy improves as the overlap ratio increases across Object Existence, Attribute, and Spatial Relation tasks.
This indicates that the low-overlap condition in \ourbenchmark{} makes it harder to associate target-related observations across frames, especially for tasks that require reasoning about source-target relations.

\noindent\textbf{Multi-hop aggregation under low overlap.}
Beyond pairwise alignment, some examples require multi-hop aggregation across three or more views.
This occurs when the target instance is invisible in the query frame and target-visible frames have little or no overlap with the query frame.
In such cases, the question cannot be solved with only two views, and the model must connect partial observations through intermediate views.

\begin{figure*}[h!]
\centering
\includegraphics[width=0.99\textwidth]{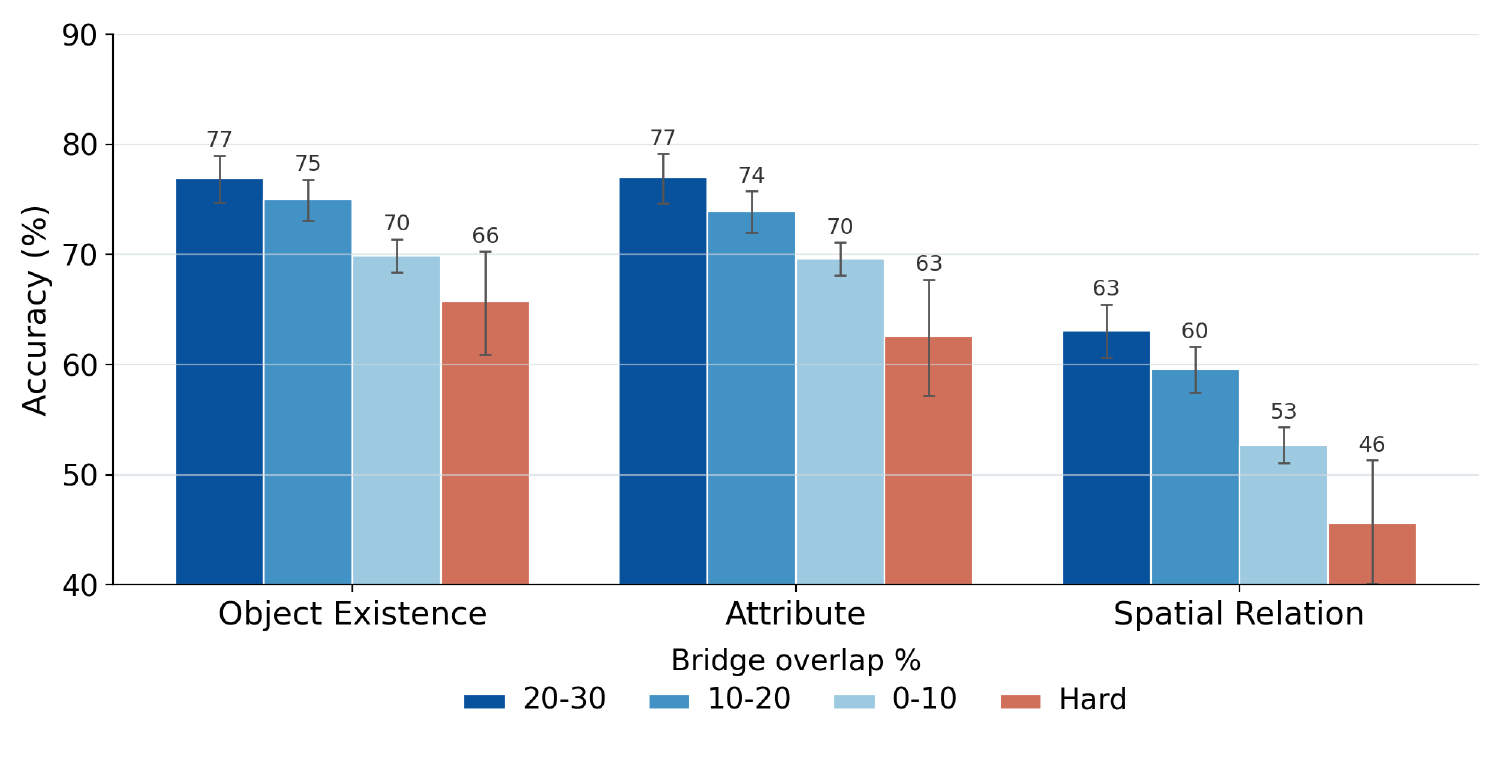}
    \caption{Effect of multi-hop aggregation under low-overlap views.}

    \label{fig:multihop_quant}
\end{figure*}

We quantify this effect using the bridge overlap between the query frame and target-visible frames.
For each example, we measure the maximum overlap between the query frame and any frame where the target instance is visible.
VQA samples with nonzero bridge overlap can still be partially resolved through direct query-reference alignment.
In contrast, hard cases have no overlap between the query frame and any target-visible frame, requiring reasoning through intermediate views.
As shown in Fig.~\ref{fig:multihop_quant}, accuracy decreases as bridge overlap becomes smaller, and the hard split shows the lowest performance.
These results indicate that low-overlap reasoning becomes especially difficult when the model cannot directly connect the query view to a target-visible view and must instead rely on multi-hop integration through intermediate views.

\begin{figure*}[h!]
\vspace{1.0em}
\centering
\includegraphics[width=0.99\textwidth]{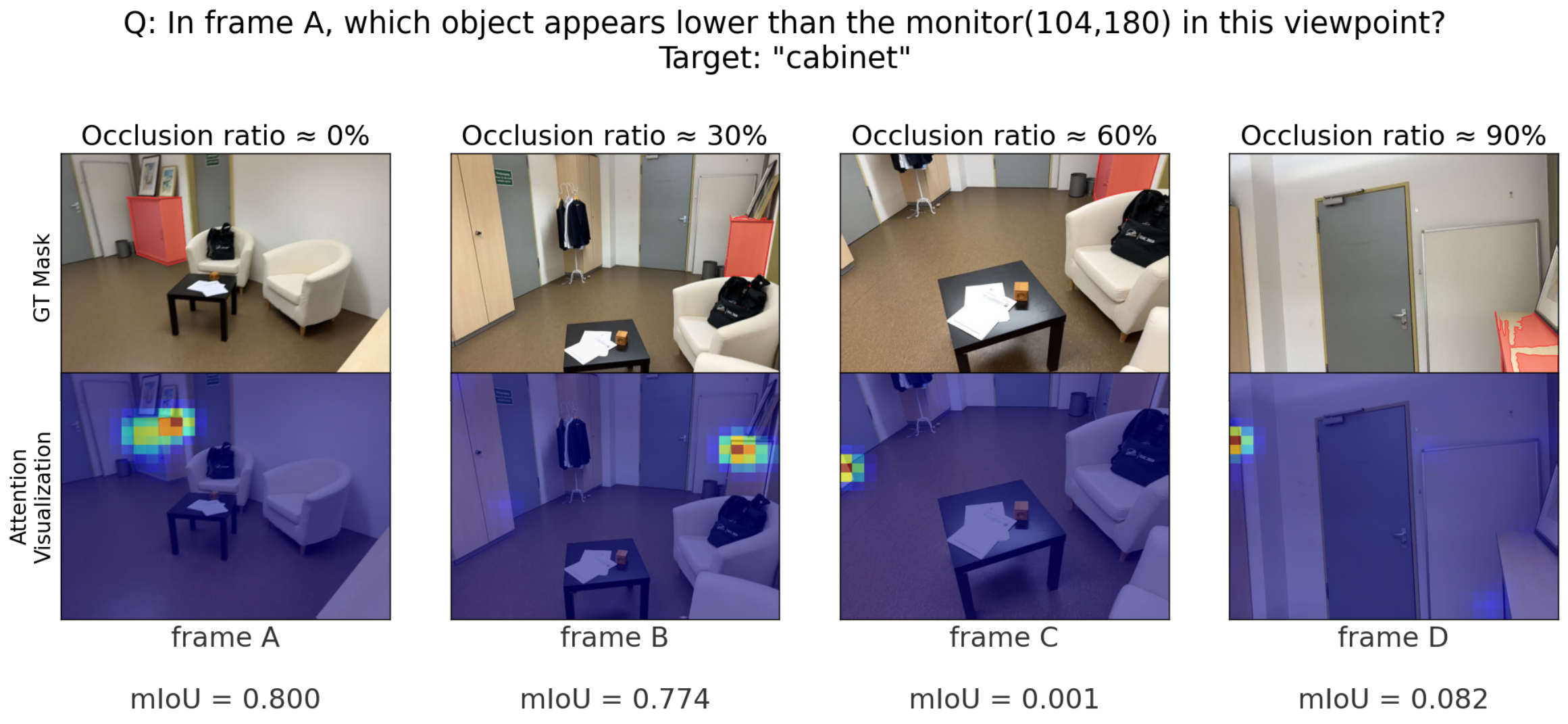}
    \caption{Qualitative analysis of object grounding under varying occlusion ratios.}

    \label{fig:occ_attn_qual}
\end{figure*}

\noindent\textbf{Occlusion.}
Finally, we analyze how occlusion affects object grounding and spatial reasoning.
Occlusion makes only a partial region of the target object observable, which can remove the visual cues needed to identify the object and relate it to other instances.
As shown in Fig.~\ref{fig:occ_attn_qual}, attention is well aligned with the target under low occlusion, but shifts toward surrounding context or visible distractors as occlusion becomes severe.

\begin{figure*}[h!]
\centering
\includegraphics[width=0.99\textwidth]{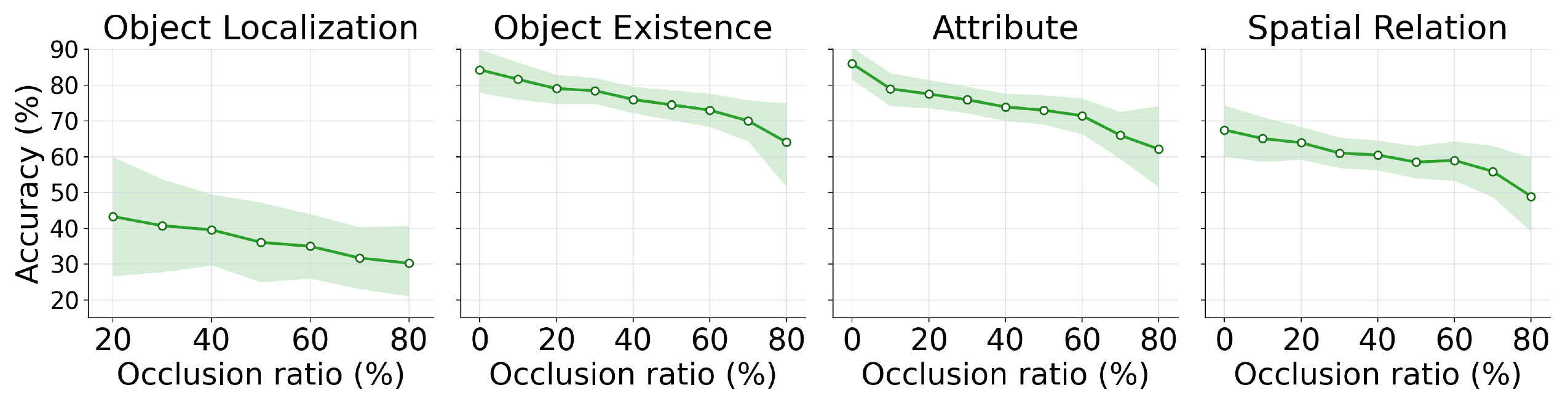}
    \caption{VQA accuracy by occlusion ratio.}

    \label{fig:occ_acc_quant}
\end{figure*}

\begin{figure*}[h!]
\centering
\includegraphics[width=0.99\textwidth]{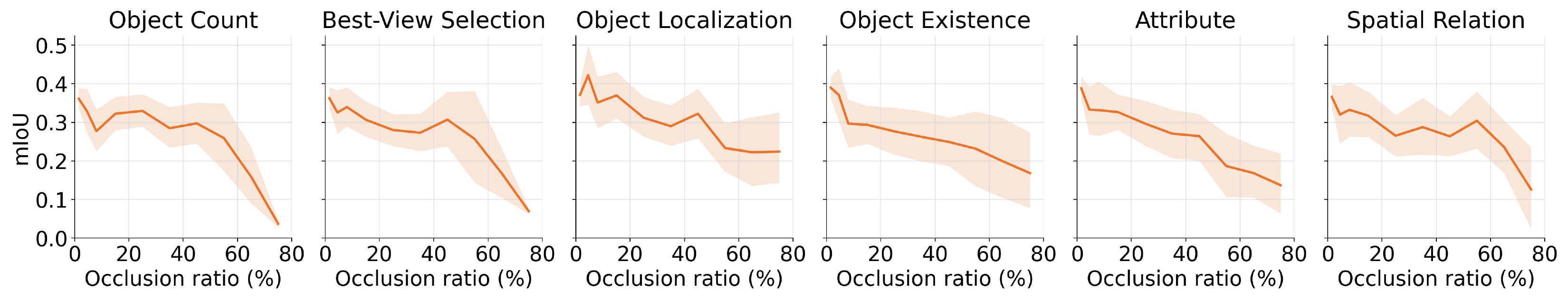}
    \caption{Grounding mIoU by occlusion ratio.}

    \label{fig:occ_miou_quant}
\end{figure*}

The quantitative results show the same pattern.
Fig.~\ref{fig:occ_acc_quant} shows that VQA accuracy decreases with higher occlusion and Fig.~\ref{fig:occ_miou_quant} further shows that grounding mIoU also decreases across task categories.
Together, these results suggest that occlusion degrades spatial reasoning by weakening target grounding and reducing the reliability of visual cues for cross-view inference.

\noindent\textbf{Summary.}
Overall, our analysis shows that failures in multi-view spatial reasoning arise from both per-frame perception limits and cross-view integration challenges.
Small or partially observable objects weaken object grounding, while duplicate same-category instances make it difficult to preserve correspondence to the same physical object across views.
The realistic conditions in \ourbenchmark{} further amplify these difficulties.
Partial visibility removes direct single-frame shortcuts, low-overlap views reduce the shared visual content needed for cross-view alignment, and severe occlusion weakens the visual cues required for grounding and relation inference.
Together, these findings show that \ourbenchmark{} captures realistic and measurable failure modes of current VLMs and enables systematic analysis of robust multi-view spatial reasoning.
\section{Additional Experiments}
To further validate the efficiency of our design beyond indoor environments, we conduct an additional ablation study on SpatialMosaic-Outdoor, which extends our benchmark to large-scale outdoor scenes. Following the same protocol as in Sec.~\ref{sec:ablation_studies}, all ablated models are fine-tuned using identical training data, optimization, and decoding settings to ensure a fair comparison.
Table~\ref{tab:ablation_mosaic_outdoor} shows that the model with a geometric encoder generally outperforms its geometry-removed variant on SpatialMosaic-Outdoor, achieving higher average performance and gains on most tasks. The gains are particularly notable in tasks involving Existence and Cam-Obj Dist.
This pattern is consistent with the indoor ablations in Tables~\ref{tab:ablation_merged}, further confirming that the proposed geometry encoder contributes to robust multi-view spatial reasoning across both indoor and outdoor layouts.
\\
\begin{table}[h]
\caption{\textbf{Ablation study on SpatialMosaic-Outdoor.}}
\centering
\setlength{\tabcolsep}{8pt}
\label{tab:ablation_mosaic_outdoor}
\scriptsize
\resizebox{\columnwidth}{!}{
\begin{tabular}{l|c|ccccc|cccc}
\toprule
 & & \rotatebox{75}{Obj. Count} & \rotatebox{75}{Best View.} & \rotatebox{75}{Obj. Exist.} & \rotatebox{75}{Obj. Rel.} & \rotatebox{75}{Obj. Loc.} & \rotatebox{75}{Obj. Count} & \rotatebox{75}{Cam-Obj Dist.} & \rotatebox{75}{Obj-Obj Dist.} & \rotatebox{75}{Obj. Loc.} \\
\textbf{Methods} & Avg. &
\multicolumn{5}{c|}{\cellcolor{yellow!10}Multiple-Choice Answer} &
\multicolumn{4}{c}{\cellcolor{orange!10}Numerical Answer} \\
\midrule
\multicolumn{11}{l}{\cellcolor{navyblue!5}\textit{SpatialMosaic-Outdoor.}} \\
LLaVA-NeXT-Video & 61.6 & 76.4 & 68.3 & 71.7 & \textbf{71.2} & \textbf{60.2} & 75.4 & 30.1 & 31.1 & \textbf{30.4} \\
LLaVA-NeXT-Video + VGGT                  & \textbf{63.3} & \textbf{78.3} & \textbf{69.3} & \textbf{73.6} & 69.2 & \textbf{60.2} & \textbf{79.4} & \textbf{40.8} & \textbf{36.7} & 29.7 \\
\bottomrule
\end{tabular}
}
\end{table}

\section{Statistics of SpatialMosaic-Bench}
\label{sec:supple_statis}

\begin{figure*}[ht!]
\centering
\begin{minipage}{\textwidth}
    \centering
    \includegraphics[width=\linewidth,height=0.2\textheight,keepaspectratio]{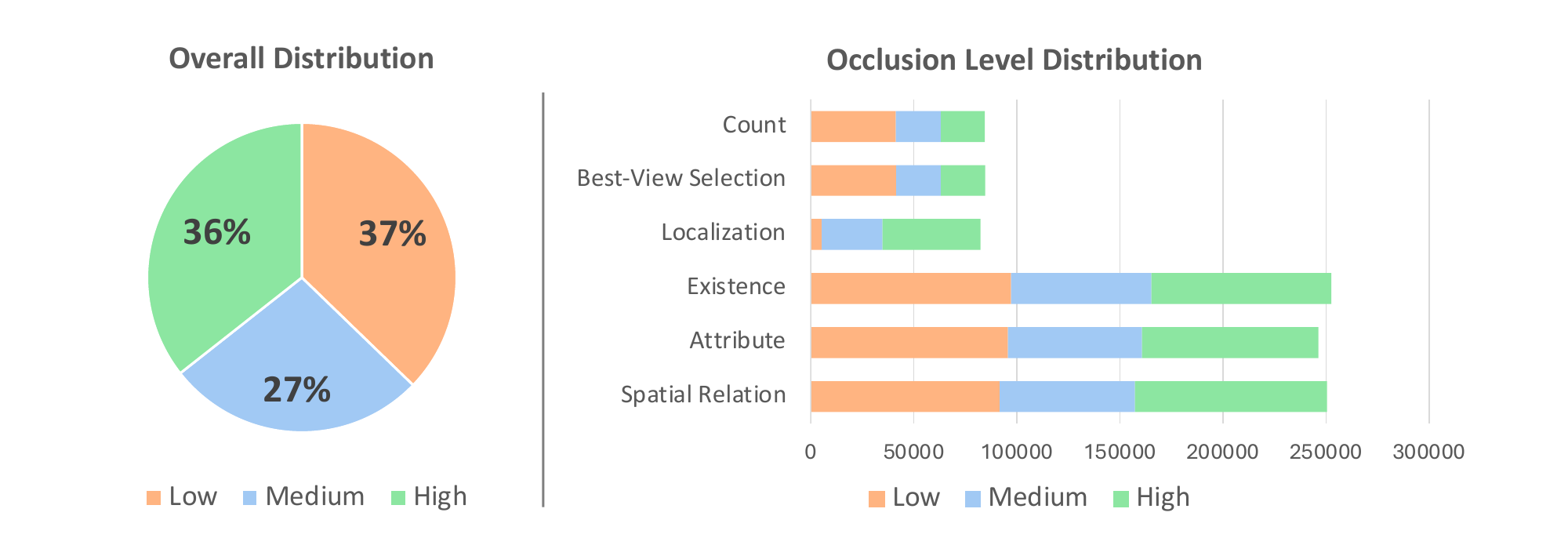}
    \caption*{(a) Occlusion Level}
\end{minipage}
\hfill
\begin{minipage}{\textwidth}
    \centering
    \includegraphics[width=\linewidth,height=0.2\textheight,keepaspectratio]{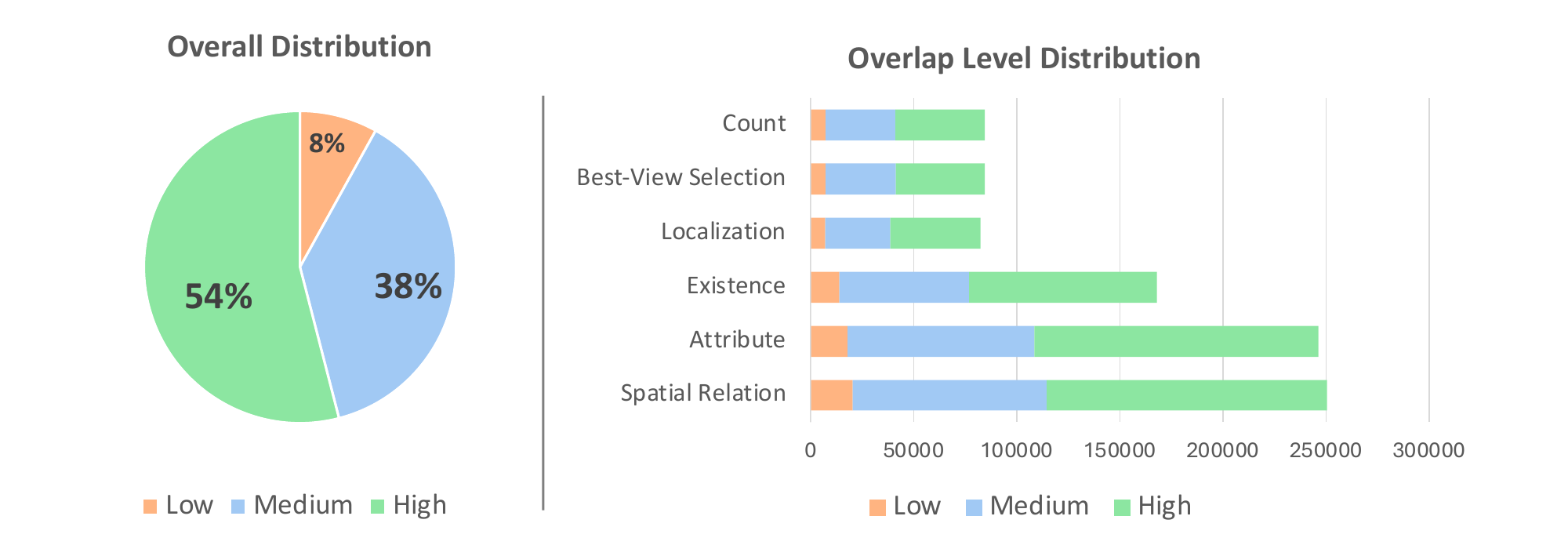}
    \caption*{(b) Overlap Level}
\end{minipage}
\hfill
\begin{minipage}{\textwidth}
    \centering
    \includegraphics[width=\linewidth,height=0.2\textheight,keepaspectratio]{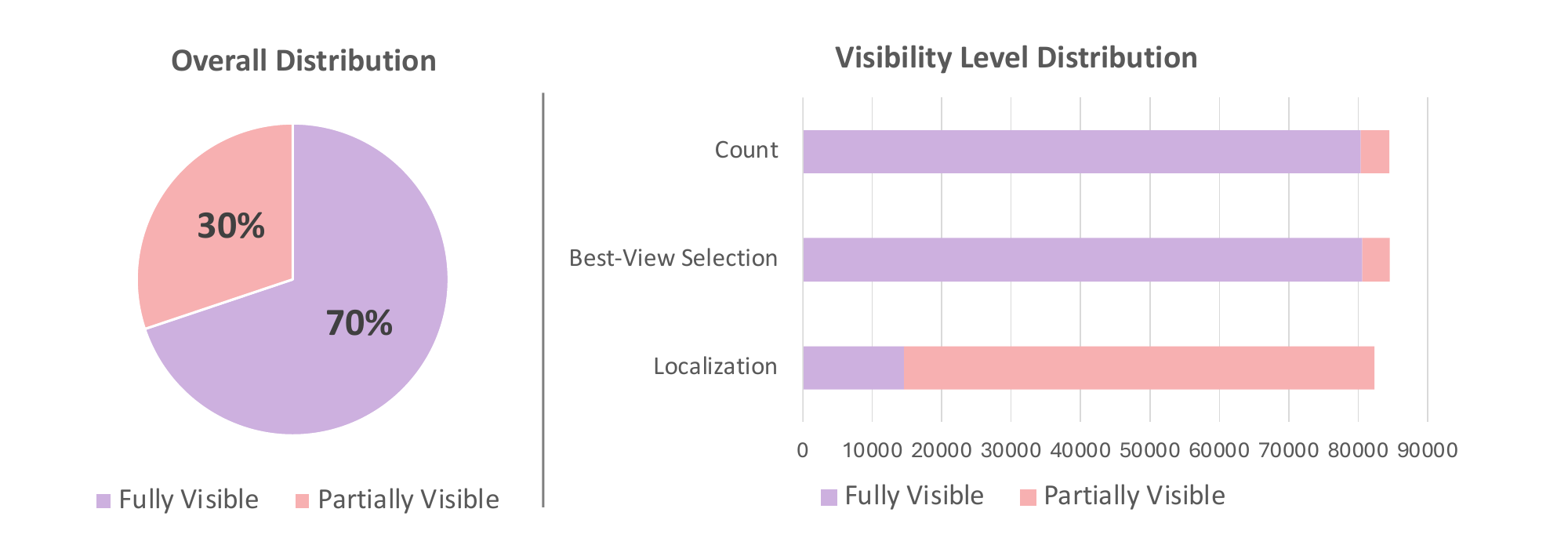}
    \caption*{(c) Visibility Level}
\end{minipage}
\caption{Difficulty Level Distribution}
\label{fig:qa_distribution}
\end{figure*}

We provide detailed statistics of \ourbenchmark{} across different difficulty levels in Fig.~\ref{fig:qa_distribution}. Our benchmark, spanning both indoor and outdoor scenes, contains a total of 1M QA pairs distributed across six main task categories: Count, Best-View Selection, Existence, Attribute, and Localization. To ensure a comprehensive evaluation of spatial reasoning capabilities, we emphasize challenging scenarios by generating more samples from high and medium difficulty levels. Note that we enforce the target object to be invisible in the query frame for attribute, existence, and spatial relation tasks; thus, all QA samples from these tasks fall under the Partially Visible category in the visibility-level distribution.

\section{Analysis under Different Conditions}
\label{app:analysis_conditions}
\begin{figure*}[t]
    \centering
    \includegraphics[width=\textwidth]{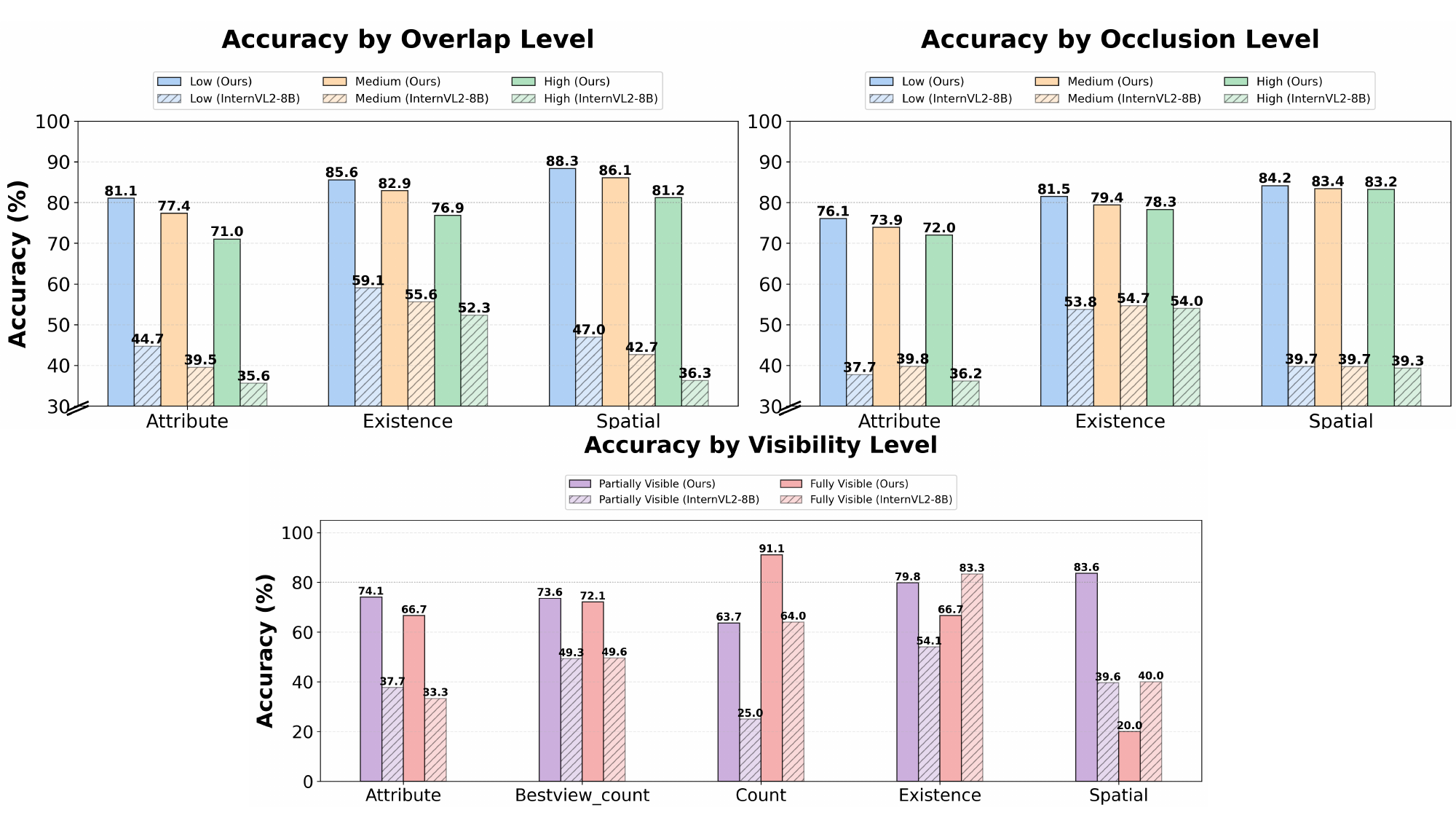}
    \caption{Comparison between VGGT-enhanced model and InternVL2-8B on \ourbenchmark{}.}
    \label{fig:accuracy_comparison}

\end{figure*}

We evaluate the proposed baseline model on \ourbenchmark{} under varying difficulty levels: Occlusion, Overlap, and Visibility. Performance consistently declines as the difficulty increases, confirming that our categorization accurately captures the challenges of multi-view spatial reasoning. We further compare this baseline with InternVL2-8B~\cite{chen2024far} across all tasks (Fig.~\ref{fig:accuracy_comparison}).
\noindent
\textbf{Visibility Level.}
For the Object count task, performance drops significantly on the Partially Visible condition, reflecting the inherent difficulty of counting objects that are not fully visible in every frame. For Best-View Selection combined with counting (Bestview count), both Partially and Fully Visible cases show similar moderate performance, as the model must both count instances and identify the optimal frame. Our model substantially outperforms InternVL2-8B on Partially Visible cases across all tasks.
\noindent
\textbf{Overlap \& Occlusion Level.}
For Attribute, Existence, and Spatial Relation tasks, performance consistently declines as task difficulty increases (from Low to High levels), with InternVL2-8B exhibiting a larger performance drop. In contrast, Count and Best-View Selection tasks are not significantly affected by low overlap and occlusion conditions. Instead, we empirically observe that their performance is primarily influenced by two factors: the total number of visible instances of the target object category across all frames and the frame-wise distribution of visible instance counts.

\section{Technical Details for SpatialMosaic}

\subsection{Spatial Annotation Framework}
\label{sec:spatial_annotation_framework}
\noindent
The spatial annotation framework establishes the geometric signals used throughout the VQA generation pipeline in \ourbenchmark{}. In this section, we provide a technical overview of the annotation process introduced in Sec.~\ref{sec:dataset}, focusing on how instance-level visibility, occlusion, and multi-view overlap are computed and stored for later use. We directly reference the definitions in Sec.~\ref{sec:dataset} and clarify how these quantities are applied in the downstream QA-generation stages.

\noindent
\textbf{Instance visibility.}
For each object instance, visibility is determined using the object-level occlusion ratio and FoV occlusion ratio defined in Sec.~\ref{subsec:data_preparation}. Using the ScanNet++ point clouds and calibrated camera parameters, each instance's 3D points are projected into every view, and visibility is computed by rendering both the scene-level depth map and per-instance depth map. FoV truncation is computed via the extended intrinsic formulation. These two metrics together form the per-frame visibility profile used in all subsequent stages.

\noindent
\textbf{Per-frame visibility masks.}
Using the visible point sets defined in Eq.~\Cref{eq:4}, we construct binary masks for each instance in each frame. These masks define which portion of the 3D geometry is observable and directly ground the multi-view filtering and relation computation.

\begin{figure*}[t]
\centering
\begin{minipage}{\textwidth}
    \centering
    \includegraphics[width=\linewidth,height=0.20\textheight,keepaspectratio]{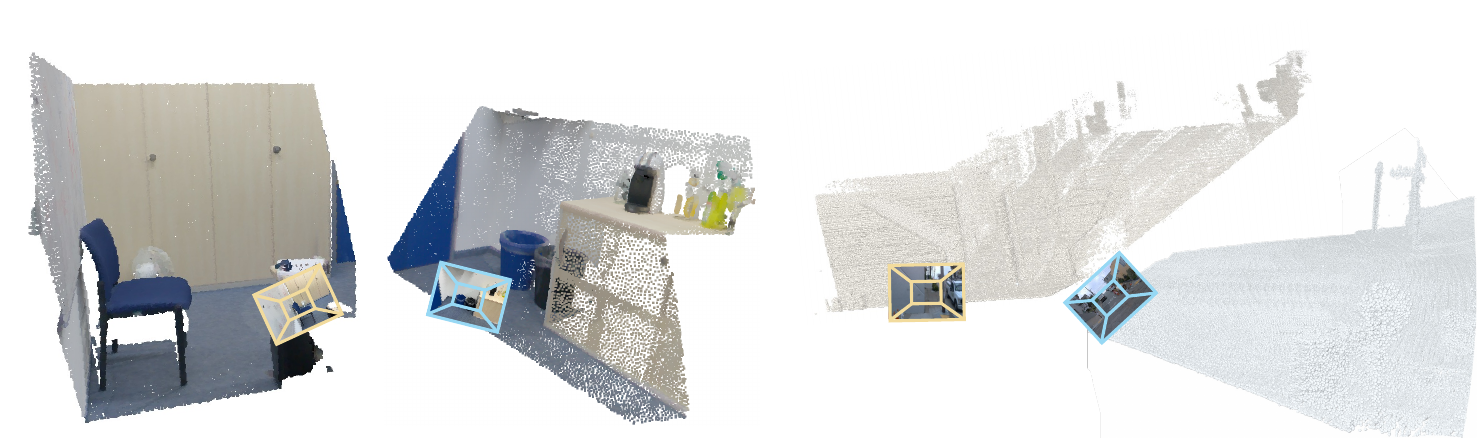}
    \caption*{(a) overlap ratio: 0\% (left: indoor, right: outdoor)}
\end{minipage}
\hfill
\begin{minipage}{\textwidth}
    \centering
    \includegraphics[width=\linewidth,height=0.20\textheight,keepaspectratio]{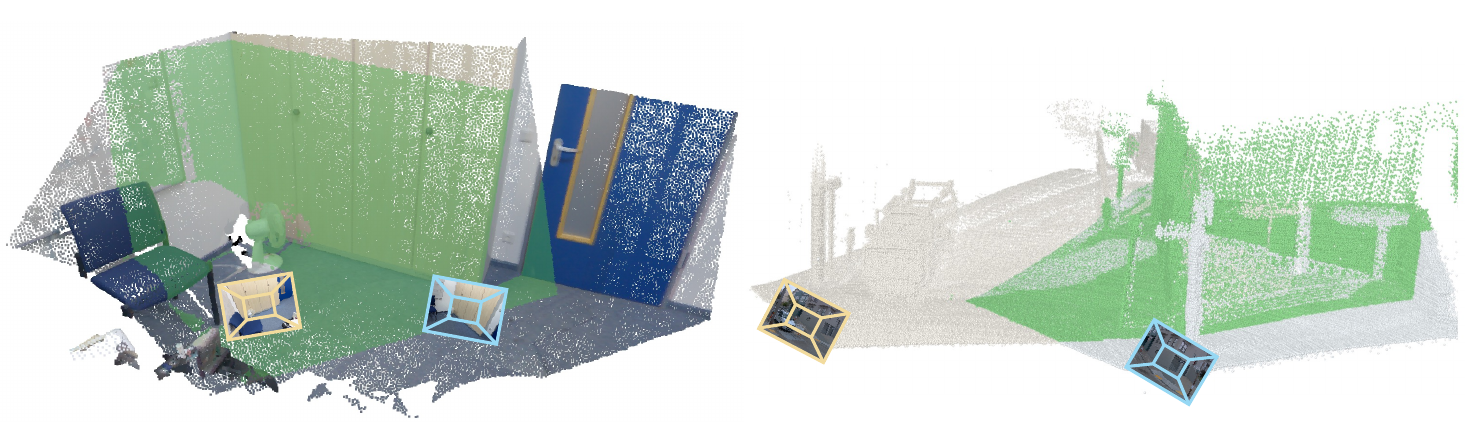}
    \caption*{(b) overlap ratio: 30\% (left: indoor, right: outdoor)}
\end{minipage}
\hfill
\begin{minipage}{\textwidth}
    \centering
    \includegraphics[width=\linewidth,height=0.20\textheight,keepaspectratio]{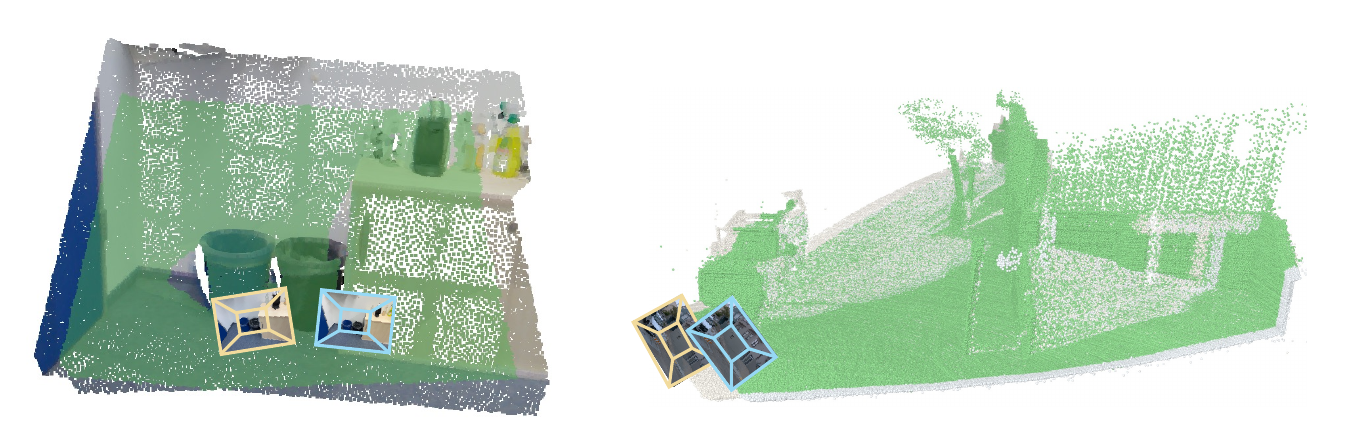}
    \caption*{(c) overlap ratio: 60\% (left: indoor, right: outdoor)}
\end{minipage}
\caption{Qualitative examples of varying overlap ratios}
\label{fig:overlap_ratio_qual}
\end{figure*}

\begin{figure*}[t]
    \centering
    \includegraphics[width=\textwidth,height=0.20\textheight,keepaspectratio]{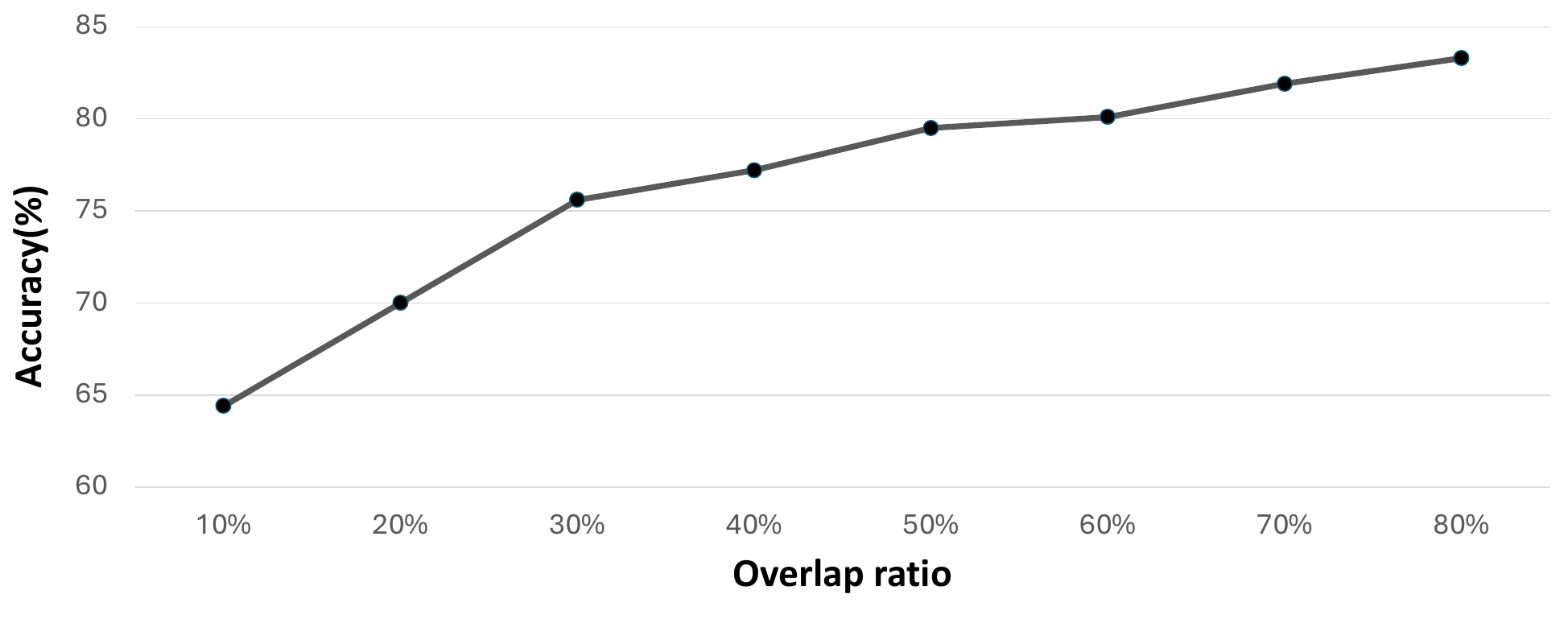}
    
    \caption{Quantitative analysis on different overlap ratios}
    \label{fig:overlap_ratio_quan}
\end{figure*}

\noindent
\textbf{Overlap computation.}
Frame overlap is computed using the intersection-over-union of visible 3D point sets as introduced in Eq.~\Cref{eq:6}. This ensures that multi-view sampling is guided by geometric overlap rather than superficial image similarity. 
Following Sec.~\ref{subsec:QA_Generation_and_Relations}, we construct VQA samples using only image pairs whose overlap ratio is below a predefined threshold $\tau$. To determine an appropriate value of $\tau$, we conduct both quantitative and qualitative analyses on a subset of the data. Figure~\ref{fig:overlap_ratio_quan} provides a quantitative analysis of model performance across different overlap ratios. Accuracy varies substantially within the 10--30\% overlap range, indicating meaningful differences in task difficulty. Based on this observation, we set $\tau = 0.3$ to retain sufficiently challenging low-overlap pairs while avoiding overly easy cases. Figure~\ref{fig:overlap_ratio_qual} presents qualitative examples of image pairs with varying overlap ratios.

\noindent
\textbf{Bounding box transformation.}
As described in Sec.~\ref{subsec:QA_Generation_and_Relations}, spatial relations are computed by comparing the positions of objects in the viewpoint of the selected query frame. Since the relation is determined by evaluating how the oriented bounding boxes of two instances are ordered along the camera-frame axes, each bounding-box vertex is first transformed into the camera coordinate system of the query view. The transformation is:
$v^{(c)} = R_{wc}(v - t_{wc})$
where $v$ is a bounding-box vertex in world coordinates, $R_{wc}$ is the world-to-camera rotation matrix, and $t_{wc}$ is the camera-center translation. The resulting vertices $v^{(c)}$ define the camera-frame bounding boxes used for axis-aligned separation when computing the directional relation in necessary tasks.

\subsection{Data Generation Pipeline}
\noindent
The data generation pipeline constructs QA samples by applying task-dependent constraints on the annotated geometric information in Sec.~\ref{sec:spatial_annotation_framework}. Once visibility, occlusion, and overlap statistics are available, the pipeline proceeds through the following steps:

\noindent
\textbf{Frame combination construction.}
For each scene, candidate multi-view combinations are formed by enumerating frame sets that satisfy the required view count and the overlap constraint. Only combinations whose internal view-overlap stays below the specified threshold are retained, ensuring sparse and complementary viewpoints.

\noindent
\textbf{Valid instance set collection.}
Within each retained combination, all object instances that appear in at least one of the included frames are collected. Instances that violate the partial-visibility requirement, such as appearing in every frame or remaining nearly fully occluded across all frames, are removed. The remaining instances carry their semantic labels, visibility flags, and camera-frame bounding boxes.

\noindent
\textbf{Query-frame and object selection.}
A query frame is randomly selected from the valid combinations. Depending on the task, the pipeline verifies whether the instances in that frame satisfy the required visibility conditions (e.g., visible source and invisible target pairs for multi-category tasks, or category uniqueness for localization). If the condition is not met, the pipeline attempts another configuration. The task-specific visibility conditions are elaborated in Sec.~\ref{sec:supple_task}.

\noindent
\textbf{Task-specific geometric computation.}
Geometric quantities are computed only after a valid configuration is found. These computations are task-specific, including directional separations between instance bounding boxes, visible-pixel statistics, or merged instance counts across views. All computations operate directly on the pre-annotated camera-frame geometry.

\noindent
\textbf{Answer and distractor generation.}
The computed values are inserted into the task templates. Distractors are generated using the rules defined for each task, such as orthogonal-axis relations or offset-based count alternatives. All options are de-duplicated, validated, and randomized.

\noindent
\textbf{QA assembly.}
The final QA entry records the selected frame combination, the question constructed from the template with appropriate instances, the multiple-choice options, and all geometric metadata required for evaluation. After such a process is completed, a single QA is generated, and the pipeline then moves on to the next sample.

\section{SpatialMosaic Task Descriptions}
\label{sec:supple_task}
\label{app:task}

\newcommand{\annot}[1]{\textcolor{gray!80}{\% #1}}

\begin{algorithm}[h!]
{\footnotesize
\SetAlgoLined
\textbf{Input:} \ Scene list $\mathcal{S}$, scene-level metadata $\mathcal{M}$, frame-level metadata $\mathcal{F}$\\
\textbf{Output:} Generated QA set $\mathcal{Q}$\\
\textbf{Initialization:}\\
\Indp
Build per-scene overlap tables from $\mathcal{F}$\\
Initialize QA buffers and global counters\\
\Indm
\textbf{Main loop over scenes:}\\
    \For{$s \in \mathcal{S}$}{
        $F_s \leftarrow \mathcal{F}[s]$\\
        $M_s \leftarrow \mathcal{M}[s]$ \\
        $OV_s \leftarrow \textit{read\_overlap\_table}\mathcal{F}[s]$\\
        \BlankLine
        \annot{Step 1. Occlusion extraction:}\\
        $(O_s, L_s) \leftarrow \textit{extract\_occlusion}(F_s)$\\
        \BlankLine
        \annot{Step 2. Frame-combination construction:}\\
        $C_s \leftarrow \textit{sample\_valid\_combos}(O_s, L_s, s)$\\
        $C_s \leftarrow \textit{overlap\_filtering}(C_s, OV_s)$\\
        \BlankLine
        \annot{Step 3. Scene-level QA generation:}\\
        \For{$t \in T$}{
            $G_t \leftarrow \textit{get\_qa\_generator}(t)$\\
            $Q_s \leftarrow G_t(s, M_s, F_s, C_s)$\\
        }
        \BlankLine
        \annot{Step 4. Accumulate results:}\\
        $\mathcal{Q} \leftarrow \mathcal{Q} \cup Q_s$\\
    }
}
    \caption{BaseQAGenerator}
\end{algorithm}
The BaseQAGenerator serves as a foundation for the QA generation process. For each frame in the scene, instance-level occlusion ratios $O_s$, together with their category labels $L_s$, are computed as explained in Sec~\ref{subsec:data_preparation}. Valid frame combinations are first sampled by selecting frame sets that satisfy the per-instance occlusion constraints. These candidate combinations are then filtered using the precomputed overlap table calculated through Eq.~\Cref{eq:6}, such that $C_s$ represents the filtered frame combinations that satisfy both occlusion and low-overlap conditions. For each task type $t \in T$, a task-specific QA generator $G_t$ produces the corresponding set of QA pairs, $Q_s$. Finally, all QAs generated for the scene are merged into the global QA set.

\begin{algorithm}[h!]
{\footnotesize
\SetAlgoLined
\caption{Object Count (Single-Category)}
\textbf{Input:} \ scene id $s$, per-scene metadata $M_s$,\\
per-scene frame metadata $F_s$, Frame combinations $C_s$ \\
\textbf{Output:} \ $Q_s$ \\ 

\BlankLine
\annot{Select frame combination from combination sets $C_s$}\\
\For{$C \in C_s$}{ 
    \annot{Step 1. Per-frame visible-instance extraction:}\\
    \For{$f \in C$}{
        $I_c(f) = \textit{per\_frame\_instance}(c, f)$
    }
\BlankLine
\annot{Step 2. Multi-view aggregation:}\\
$V_c \leftarrow \bigcup_{f \in C} I_c(f)$ \\
$\text{GT} \leftarrow |V_c|$ \\
\BlankLine
\annot{Step 3. Multiple-choice option generation:}\\
$\mathcal{D} \leftarrow \textit{count\_distractor}(\text{GT})$\\
$\mathfrak{O} \leftarrow \{\text{GT}\} \cup \text{Sample}_3(\mathcal{D})$
\BlankLine
\annot{Step 4. Output assembly:}\\
$Q_s = filling\_qa(\mathfrak{O}, \mathcal{T}, \text{GT})$
}
}
\end{algorithm}
\noindent
\textbf{Object count.}
The Object Count task determines how many instances of an object category $c$ are visible throughout the frame combination $C$. The task first identifies the per-frame visible-instances $I_c(f)$ for all frame combinations, defined as the function $I_c(f) = \{\, i \mid \text{cat}(i)=c,\; \text{occ}(i,f) \le \tau \,\}$. Multi-view aggregation merges all visible instance sets across the selected frame set, forming the union $V_c$. The ground-truth count is then obtained as $\text{GT}$, representing the total number of unique instances observed. For answer choice generations, a distractor pool $\mathcal{D} = \{\max(1,\; \text{GT}+\delta) \mid \delta \in \{-3,-2,-1,1,2,3\}\}$ generates incorrect options of a small offset from $\text{GT}$, where three of them are sampled as distractors in $\mathfrak{O}$. Finally, the \textit{filling\_qa} function utilizes all task-specific variables, including the question template $\mathcal{T}$, to generate the final QA instance as described in Fig.~\ref{fig:data_generation}.

\begin{algorithm}[h!]
{\footnotesize
\SetAlgoLined
\caption{Object Bestview (Single-Category)}
\textbf{Input:} \ scene id $s$, per-scene metadata $M_s$,\\
per-scene frame metadata $F_s$, Frame combinations $C_s$ \\
\textbf{Output:} \ $Q_s$ \\ 
\For{$C \in C_s$}{ 
\annot{Step 1. Per-frame visible-instance extraction:}\\
    \For{$f \in C$}{
        $I_c(f) = \textit{per\_frame\_instance}(i, c, f)$
    }
    \BlankLine
    \annot{Step 2. Multi-view aggregation:}\\
    $V_c \leftarrow \bigcup_{f \in F} I_c(f)$ \\
    $\text{GT} \leftarrow |V_c|$ \\
    \BlankLine
    \annot{Step 3. Best-view selection:}\\
    \For{$f \in C$}{
        $n_c(f) \leftarrow |I_c(f)|$   \\
        $A_c(f) \leftarrow \sum_{i \in I_c(f)} \textit{vispix}(i,f)$
    }
    $f_b \leftarrow \arg\max_{f \in F}\big( n_c(f),\ A_c(f) \big)$\\
    \BlankLine
    \annot{Step 4. Multiple-choice option generation:}\\
    {\footnotesize
    $\mathfrak{O} = 
    \{\, 
    (\mathrm{GT}, f_b),\ 
    (\mathrm{GT}', f_b),\ 
    (\mathrm{GT}, f_b'),\ 
    (\mathrm{GT}', f_b')
    \,\}\,
    $
    }\\
    $f_b' \leftarrow \text{random\_choice}(C \setminus \{f_b\})$
    \BlankLine
    \annot{Step 5. Output assembly:}\\
    $Q_s = filling\_qa(\mathfrak{O}, \mathcal{T}, \text{GT})$
    }
}
\end{algorithm}
\noindent
\textbf{Best-view selection.}
The Best-view selection task determines how many instances of an object category $c$ are visible throughout the sampled frames $f$, and calculates the frame $f_b$ that gives the most informative view. The per-frame visible-instance extraction and multi-view aggregation are identical to the object count task. To select the best-view frame, we first measure how many instances of category $c$ are visible in a frame through $n_c(f)$, then calculate the total visible-pixel area $A_c(f)$ for those instances computed by $vispix(i,f)$ over all visible instances $i$ in $I_c(f)$. The best frame $f_b$ is determined by the highest visible count; if multiple frames have the same count, ties are broken by visible-pixel comparison. Options $\mathfrak{O}$ contains one correct answer pair with the correct $\text{GT}$ and the correct best frame $f_{b}$, while three other distractor pairs either have an incorrect count $GT'$ from $\mathcal{D}$, an incorrect best frame $f_b'$ from $C\setminus\{f^{*}\}$, or both.

\begin{algorithm}[h!]
{\footnotesize
\SetAlgoLined
\caption{Object\hspace{0.16em}Localization\hspace{0.16em}(Single-Category)}
\textbf{Input:} \ scene id $s$, per-scene metadata $M_s$,\\
per-scene frame metadata $F_s$, Frame combinations $C_s$ \\
\textbf{Output:} \ $Q_s$ \\  
\For{$C \in C_s$}{ 
\annot{Step 1. Per-frame visible-instance extraction:}\\
    \For{$f \in C$}{
        $I_c(f) = \textit{per\_frame\_instance}(i, c, f)$
    }
    
    \annot{Step 2. Multi-view aggregation:}\\
    $V_c \leftarrow \bigcup_{f\in F} I_c(f)$ \hfill \\

    \annot{Step 3. Localization supervision:}\\
    \If{$i_t$ is visible in $f_q$}{
        GT $\leftarrow$ $\text{``Yes; } (x_t,y_t)\text{''}$\\
    }
    \Else{
        GT $\leftarrow$ ``No''
    }
    
    \annot{Step 4. Multiple-choice option generation:}\\
    \If{GT == ``Yes; $(x_t,y_t)$''}{
        $\mathfrak{O} = \{\mathrm{GT}, \{\text{Yes;}(x_n,y_n)\}_{n=1}^{2}, \text{No}\}$\\
    }
    \Else({GT =``No''}){
        $\mathfrak{O} = \{\mathrm{GT}, \{\text{Yes;}(x_n,y_n)\}_{n=1}^{3}$\\
    }
    
    \annot{Step 5. Output assembly:}\\
    $Q_s = filling\_qa(\mathfrak{O}, \mathcal{T}, \text{GT}, f_q)$
    }
}
\end{algorithm}
\noindent
\textbf{Object localization.}
The Object Localization task determines whether a target instance $i_t$ exists in the query frame $f_q \in C$, and returns its 2D bounding box center coordinates $(x_t,y_t)$ if it is visible. We construct the $\text{GT}$ by determining whether the instance is visible in the query frame using the function $\textit{is\_visible}(i,f)=\mathds{1}[\mathrm{occ}(i,f) \le \tau]$. If the instance is visible, the $\text{GT}$ returns a "Yes" with the instance's 2D bounding box center coordinates $(x_t,y_t)$, and the distractor options in $\mathfrak{O}$ consist of two positive options with incorrect coordinates ($\{(x_n, y_n)\}_{n=1}^{2}$) and a negative option. If the instance is not visible, the $\text{GT}$ returns a "No", and the distractor options in $\mathfrak{O}$ consist of three positive options with incorrect coordinates ($ \{(x_n, y_n)\}_{n=1}^{3}$). Incorrect coordinates are randomly sampled within the bounding box area of the target instance to avoid trivializing the task.

\begin{algorithm}[h!]
{\footnotesize
\SetAlgoLined
\caption{Multi-Category Tasks}
\textbf{Input:} \ scene id $s$, per-scene metadata $M_s$,\\
per-scene frame metadata $F_s$, Frame combinations $C_s$ \\
\textbf{Output:} \ $Q_s$ \\ 
\For{$C \in C_s$}{ 
    \annot{Step 1. Per-frame visible-instance extraction:}\\
    \For{$f \in F_s$}{
        $I_c(f) = \textit{per\_frame\_instance}(i, c, f)$
    }
    
    \BlankLine
    \annot{Step 2. Instance-pair construction:}\\
    \ForEach{combo $(i,C)$}{
        $f_q \in C$ \\
        $(i_s, i_t) \leftarrow \textit{select\_src\_tgt\_objects}(f_q)$ \\
    }
    
    \BlankLine

    \annot{Step 3. Spatial-relation evaluation:}\\
    {\footnotesize
    $R \leftarrow \textit{compute\_relation}(i_s, i_t, a)$ \\
    }
    \BlankLine
    \annot{Step 4. Question formation:}\\
    {\footnotesize
    $D \leftarrow \textit{relation\_distractor}(R)$\\
    
    $\mathfrak{O} = \{R, D_1, D_2, D_3\}$
    
    }
    
    \BlankLine
    \annot{Step 5. Output assembly:}\\
    $Q_s = filling\_qa(\mathfrak{O}, \mathcal{T}, \mathcal{R}, f_q, i_s, i_t)$

    }
}
\end{algorithm}
\noindent
\textbf{Multi-category tasks.}
Multi-category tasks share the same framework and sampling logic, where their core function is to determine the spatial relation between two object instances. After the initial per-frame visible-instance verification, the pipeline samples source and target instance pairs $i_s$ and $i_t$ from a randomly selected query frame $f_{q}$ with the following constraints: (1) $i_s$ and $i_t$ cannot be the same object category, (2) the source instance $i_s$ must be visible in the query frame $f_{q}$ aligned with the predefined visibility threshold, and (3) the target instance $i_t$ must not be visible in the query frame $f_{q}$ with the same conditions. If all conditions are validated, we obtain an instance pair $(i_s,i_t)$ for spatial inference. Directional relations $R$ are computed by transforming a 3D oriented bounding box into the coordinate system of $f_{q}$ to analyze their minimum and maximum coordinates along axis $a$ (where $a \in \{x, y, z\}$) for strict geometric assessment. 

Although all multi-category tasks share the same process up until computing the spatial relation between the instance pair $(i_s, i_t)$, the difference arises from constructing distractors $D$ and options $\mathfrak{O}$:
\begin{itemize}
    \item \textbf{Occlusion-Aware Existence}
    The occlusion-aware existence task generates a question asking if the target instance exhibits a specific directional relationship from the source instance (e.g., Does the cup appear to the right of the wallet?). The corresponding options $\mathfrak{O}$ are binary, containing a "Yes" and a "No", which analyzes the model's capability of determining whether a given relation is true or not. If the relation specified in the question is equal to $R$ from \textit{compute\_relation()}, the correct answer is "Yes"; otherwise, it is "No". 
    \item \textbf{Occlusion-Aware Attribute}
    The occlusion-aware attribute task provides the spatial relation and the source instance, and asks which object satisfies the specified spatial relation with the source instance in the query frame $f_{q}$ (e.g., What object appears to the left of the phone?). When constructing the options $\mathfrak{O}$, we construct an answer pool and a distractor pool: the answer pool contains the correct instance $i_t$, and the distractor pool contains other instances  $i_x$ in the scene that exhibit the opposite relation mentioned in the question by calculating $\textit{compute\_relation}(i_s, i_x, a)$. If the computed relation $R$ is opposite to the queried relation, $i_x$ is added to the distractor pool. In $\mathfrak{O}$, we sample one option from the answer pool (containing the target instance) and the remaining three from the distractor pool, ensuring that only one object satisfies the query condition. Note that constructing valid multiple-choice options for the attribute task requires at least four distinct object categories to be visible within the given frames. Since WOD contains only four annotated object categories in total, we do not generate attribute-based VQA samples for the outdoor subset.
    \item \textbf{Occlusion-Aware Spatial Relation}
    The occlusion-aware spatial relation task provides the source instance, target instance, and query frame $f_q$, and asks which spatial relation holds between the two objects (e.g., Where does the pillow appear relative to the door?). Upon evaluation on a specific axis $a$, we first calculate the true spatial relation $R$ between $i_s$ and $i_t$ using \textit{compute\_relation()}. The resulting relation $R$ serves as the correct answer and is stored in the answer pool. The distractor pool consists of spatial relations that differ from the true relation between $i_s$ and $i_t$. Therefore, the first distractor $D_1$ is the opposite relation of $R$ along the same axis $a$. The other two distractors $D_2$ and $D_3$ are derived by computing the relations between $i_s$ and $i_t$ along the two remaining orthogonal axes $b$ and $c$. Since $R_b$ and $R_c$ represent the true relations along axes $b$ and $c$, their opposite relations are added to the distractor pool. For example, if the target instance is located "to the left", "below", and "farther" from the source instance, and the evaluation is performed along the x-axis (left/right), the answer pool contains "left", while the distractor pool contains "right", "above", and "closer". However, this construction may introduce a potential axis-based bias, since when two relations from the same axis are included, one of them must be correct. To mitigate this issue, we additionally construct a bias-reduced VQA variant, where distractors are not sampled from the same axis as the correct relation. As illustrated in Fig.~\ref{fig:relation_example} and~\ref{fig:relation_example_outdoor}, this variant includes only relations from orthogonal axes, resulting in three-option multiple-choice questions that prevent axis-based shortcuts. 
\end{itemize}

\section{Human Verification and Quality Check}
\label{app:human_verification}

We conduct human verification to assess the reliability of SpatialMosaic-Bench, which is constructed through an automatic data generation pipeline. We sample 100 questions per task category for a total of 3,000 inspected QAs, consisting of 1,700 indoor and 1,300 outdoor examples. 
Each example is inspected by annotators using a structured checklist. Annotators verify whether: (1) the question is understandable and unambiguous, (2) the answer can be inferred from the provided multi-view images, (3) the ground-truth answer is correct, (4) the target object is correctly grounded, (5) the related object used for spatial comparison is correctly grounded, and (6) the assigned visibility condition, including partial visibility, occlusion, or low-overlap, is valid.
Beyond these standard criteria, annotators further examine whether the automatically computed geometric annotations are perceptually reasonable. In particular, they check whether the estimated occlusion ratio matches the visible evidence in the frames and whether ambiguity arises from extremely small, heavily occluded, or visually indistinguishable instances. Examples with unreliable geometric annotations or ambiguous visual evidence are removed. Disagreements are resolved by annotators.
This verification is designed to assess both language-level quality and geometry-level reliability. Our protocol explicitly checks object grounding, visibility-condition labels, and the perceptual validity of automatically computed geometric annotations, which are central to SpatialMosaic-Bench. After filtering problematic cases, 2,728 of the 3,000 questions are retained, yielding an overall acceptance rate of $90.9\%$.

\section{Bias Analysis}
\label{app:bias_analysis}
We provide a detailed analysis of the bias-mitigation strategies used in \ourbenchmark{}. (1) To prevent answer-option imbalance in Multiple-Choice Question(MCQ) tasks, the position of the correct answer is randomly assigned while enforcing a uniform distribution over all answer choices. This prevents the model from relying on trivial answer-selection heuristics, such as favoring a specific option index, and ensures that performance reflects the model's reasoning capability rather than positional biases. (2) Unlike prior VQA datasets that predominantly generate QA pairs from frequently occurring object categories, we construct a uniform number of QA pairs across all available categories. By leveraging the full set of object categories rather than focusing on common instances, we increase semantic diversity and mitigate skewed label distributions and object co-occurrence biases. This design encourages models to leverage geometric information for prediction rather than relying on category priors or language cues. (3) Query frames are randomly sampled from candidate frames, while answers are determined by explicit geometric criteria. For example, in best-view selection tasks, the correct answer is derived from geometric signals such as visible object instances and pixel coverage to reduce frame-index and viewpoint biases. (4) Finally, we examine whether performance is influenced by the aforementioned biases or dataset priors rather than reasoning capability. We observe that model performance consistently decreases as task difficulty increases, requiring more complex cross-view geometric reasoning (see \App{analysis_conditions} for details). These results indicate that accuracy is driven by the model’s multi-view reasoning capability rather than by bias-induced shortcuts or dataset priors.

\section{Architecture Details of the Multi-view VLM Baseline}
\label{app:architecture_details}

The proposed baseline constructs a joint representation of a 3D scene by extracting visual tokens and geometric tokens from multi-view images. These tokens are fused via cross-attention to form a geometry-aware representation, which is then projected into the language model input space for answer generation. Given multi-view images $\{\mathbf{I}_v\}_{v=1}^{V}$, we employ two encoders: a visual encoder $E_{vis}$ and a geometric encoder $E_{geo}$.

\noindent
\textbf{Visual encoder.}
We use a pretrained CLIP ViT as the visual encoder. For each view $\mathbf{I}_v$, the visual encoder produces a sequence of patch-level visual tokens:
\begin{equation}
F_{vis}^{(v)} = E_{vis}(\mathbf{I}_v) \in \mathbb{R}^{T_{vis}^{(v)} \times d},
\end{equation}
where $T_{vis}^{(v)}$ is the number of visual tokens for view $v$ and $d$ is the feature dimension. We aggregate visual tokens across all views by concatenating them along the token dimension:
\begin{equation}
F_{vis} = \left[ F_{vis}^{(1)} ; F_{vis}^{(2)} ; \cdots ; F_{vis}^{(V)} \right] \in \mathbb{R}^{T_{vis} \times d}.
\end{equation}

\noindent
\textbf{Geometric encoder.}
For geometric encoding, we adopt VGGT~\cite{wang2025vggt} as $E_{geo}$. VGGT jointly processes multi-view images and produces geometry-aware representations that capture scene-level geometric structure. Given the multi-view images, the encoder yields spatial features and camera tokens:
\begin{equation}
(F_{spa}, z) = E_{geo}(\{\mathbf{I}_v\}_{v=1}^{V}),
\end{equation}
where $F_{spa} \in \mathbb{R}^{T_{spa} \times d}$ denotes spatial features and $z \in \mathbb{R}^{V \times d}$ denotes camera tokens. We concatenate the spatial features and camera tokens to obtain the geometric token set:
\begin{equation}
F_{geo} = \left[ F_{spa} ; z \right] \in \mathbb{R}^{(T_{spa} + V) \times d}.
\end{equation}

\noindent
\textbf{Cross-attention fusion.}
We fuse the visual tokens $F_{vis}$ with the geometric tokens $F_{geo}$ through cross-attention to obtain geometry-aware visual tokens:
\begin{equation}
F_{fuse} =
\sigma\left(
\frac{(F_{vis}W_q)(F_{geo}W_k)^T}{\sqrt{d_k}}
\right)(F_{geo}W_v),
\end{equation}
where $W_q$, $W_k$, and $W_v$ are learnable projection matrices, $d_k$ is the key dimension, and $\sigma$ denotes the softmax operation. Following LLaVA-NeXT-Video~\cite{llavanextvideo}, the fused tokens $F_{fuse}$ are passed through a two-layer projector to obtain $F'_{fuse}$. The projected tokens are then concatenated with question tokens $F_{question}$ and fed into the language model backbone for answer generation.

\section{Experimental setting}
\label{app:experimental_setting}
As described in Sec.~\ref{sec:dataset}, we construct the SpatialMosaic dataset using ScanNet++~\cite{scannetpp} and WOD~\cite{Sun_2020_CVPR}, spanning both indoor and outdoor scenes. We select 757 scenes for training and 179 for testing.
To enable rich and diverse VQA generation, we utilize all annotated object categories present in each scene. Moreover, to ensure that the generated QA pairs cover a wide variety of target–source object combinations, each target object category is explicitly paired with multiple source-object categories. This pairing strategy enables the pipeline to generate a large and diverse set of QA pairs across a wide variety of object pairs.
For a fair comparison, all ablated models are fine-tuned under identical training settings using the training splits of SpatialMosaic and VSI-Bench, where the latter follows the data generation procedure described by VSI-Bench. For training, we use 8 NVIDIA H200 GPUs and dual AMD EPYC 7763 CPUs with a batch size of 4 for one epoch, which takes approximately 16 hours per model. To conduct an evaluation on our \ourbenchmark{}, we compare with various open-source MLLM baselines, with inference time ranging from 1 to 5 hours depending on model size.
Since full-scale comparison experiments are extremely time-intensive, we conduct the fine-tuning and quantitative evaluation in the main paper using a reduced subset of the dataset consisting of 200K training samples and 100K test samples. We used accelerator library with DeepSpeed ZeRO stage 2 optimization for distributed training. The learning rate is set to $2 \times 10^{-5}$, the weight decay to $0.0$, and we adopt a cosine learning rate scheduler. Training is performed for 5 epochs. Both the visual and geometry encoders are frozen, and multi-view features are integrated through a 3D-fusion module composed of a cross-attention layer followed by a projection layer.

\begin{figure*}[!htb]
    \centering
    \includegraphics[width=\textwidth]{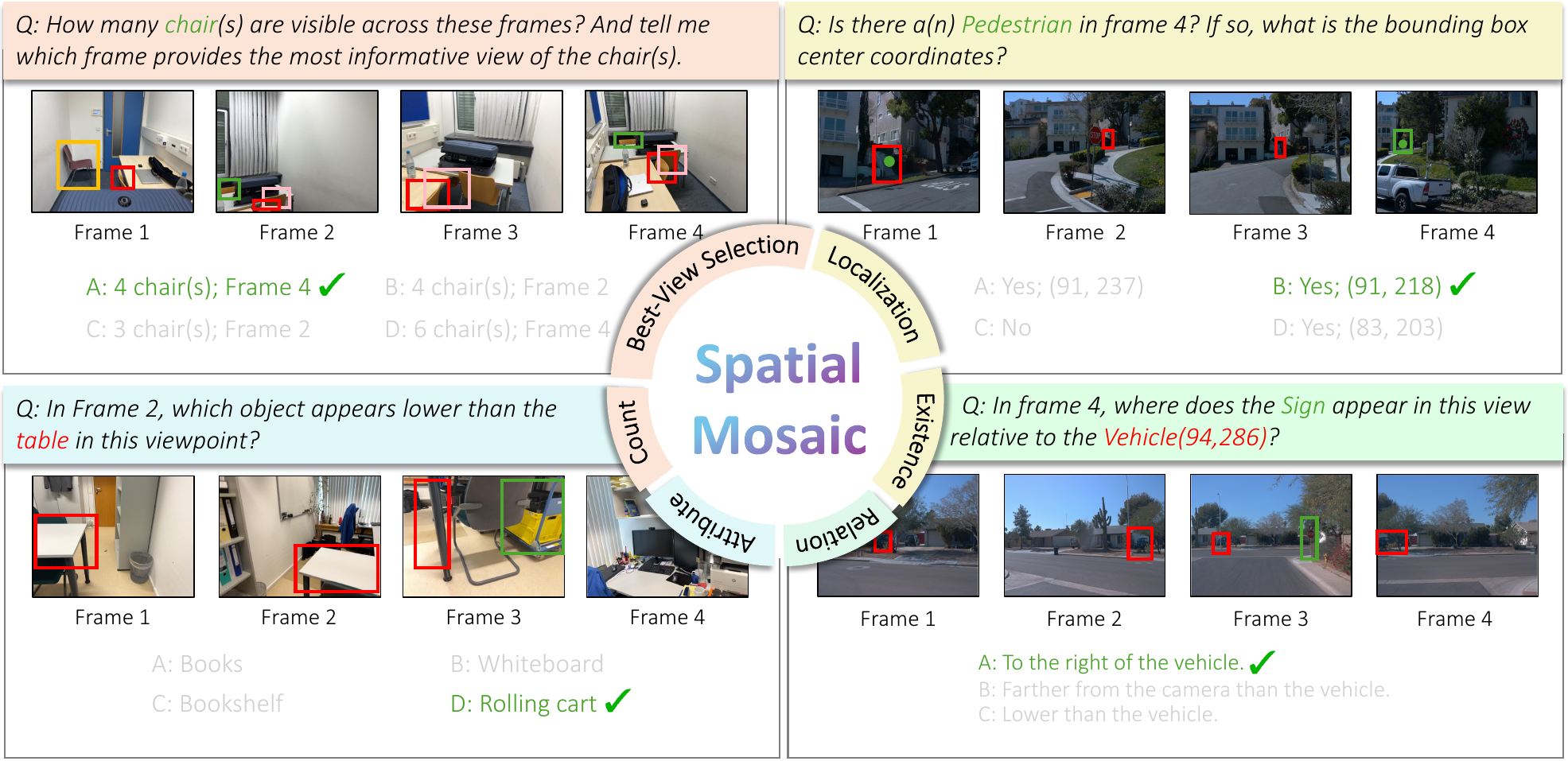}
    \caption{We present \textbf{SpatialMosaic}, a benchmark designed to evaluate 3D spatial reasoning from fragmented visual cues across multiple viewpoints, spanning both indoor and outdoor scenes. Our benchmark focuses on three challenging real-world scenarios: partial visibility, occlusion, and low-overlap views.}
    \label{fig:main_figure}
\end{figure*}

\section{SpatialMosaic Data Samples}
\label{sec:supple_samples}
\begin{figure*}[t]
    \centering
    \includegraphics[width=\textwidth,height=0.25\textheight,keepaspectratio]{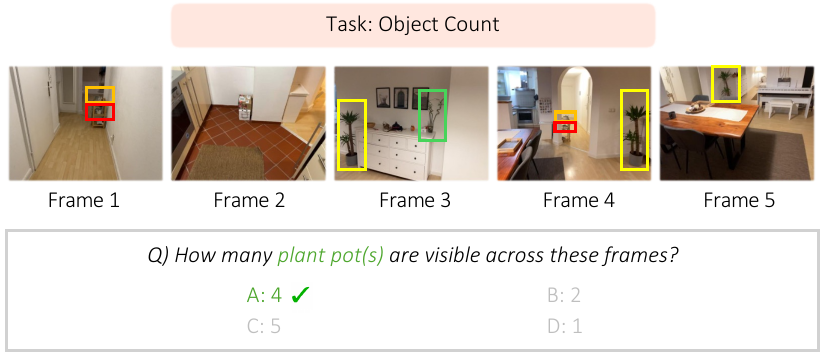}
    \clearpage
    
    \caption{Object Count example}
    \label{fig:count_example}
\end{figure*}

\begin{figure*}[t]
    \centering
    \includegraphics[width=\textwidth,height=0.26\textheight,keepaspectratio]{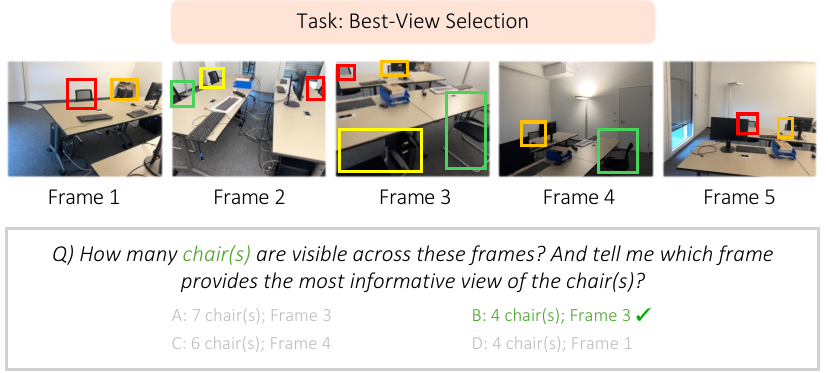}
    \clearpage
    
    \caption{Best-View Selection example}
    \label{fig:bestview_example}
\end{figure*}

\begin{figure*}[t]
    \centering
    \includegraphics[width=\textwidth,height=0.25\textheight,keepaspectratio]{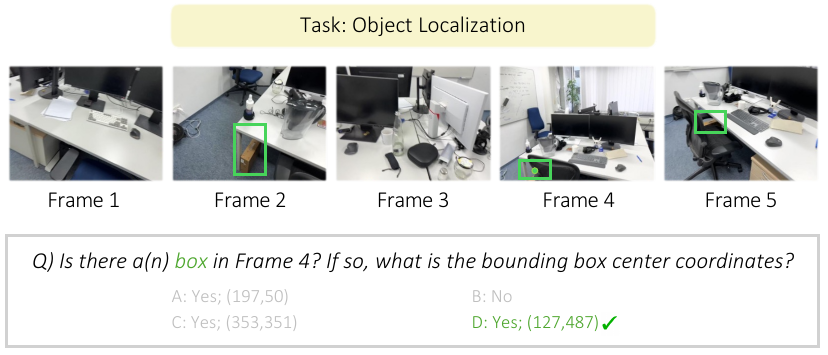}
    \clearpage
    
    \caption{Object Localization example}
    \label{fig:localization_example}
\end{figure*}

\begin{figure*}[t]
\centering
\begin{minipage}{\textwidth}
    \centering
    \includegraphics[width=\linewidth,height=0.25\textheight,keepaspectratio]{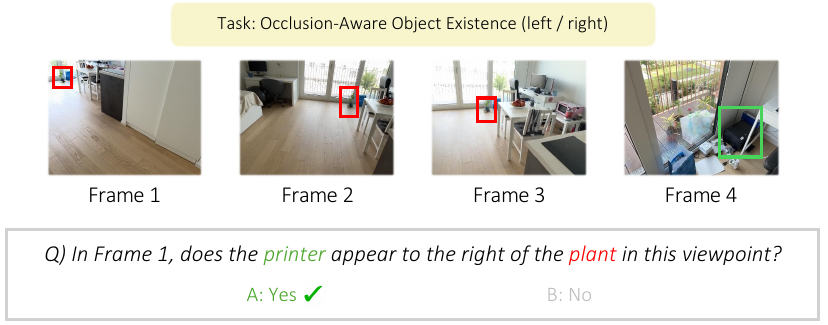}
    \caption*{(a) left / right}
\end{minipage}
\hfill
\begin{minipage}{\textwidth}
    \centering
    \includegraphics[width=\linewidth,height=0.25\textheight,keepaspectratio]{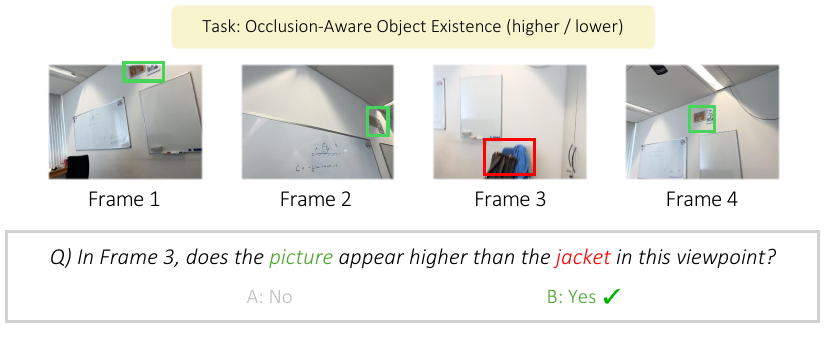}
    \caption*{(b) higher / lower}
\end{minipage}
\hfill
\begin{minipage}{\textwidth}
    \centering
    \includegraphics[width=\linewidth,height=0.25\textheight,keepaspectratio]{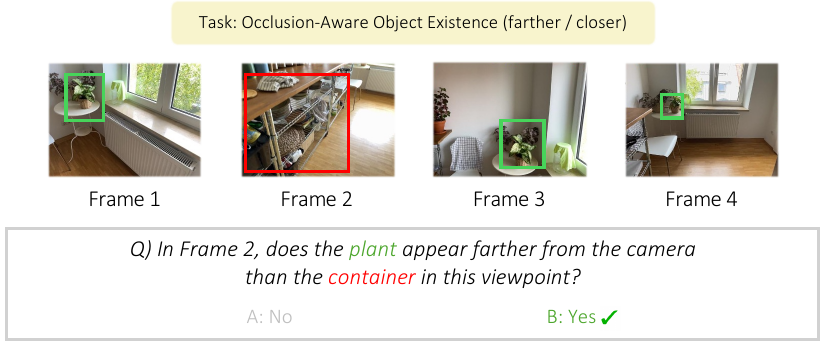}
    \caption*{(c) farther / closer}
\end{minipage}
\caption{Occlusion-Aware Object Existence examples}
\label{fig:existence_example}
\end{figure*}

\begin{figure*}[t]
\centering
\begin{minipage}{\textwidth}
    \centering
    \includegraphics[width=\linewidth,height=0.25\textheight,keepaspectratio]{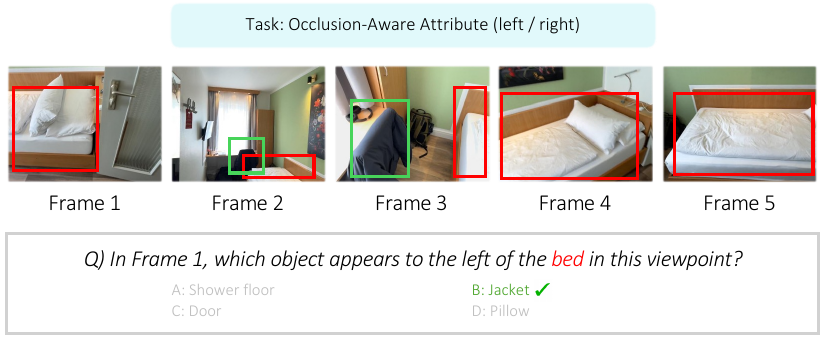}
    \caption*{(a) left / right}
\end{minipage}
\hfill
\begin{minipage}{\textwidth}
    \centering
    \includegraphics[width=\linewidth,height=0.25\textheight,keepaspectratio]{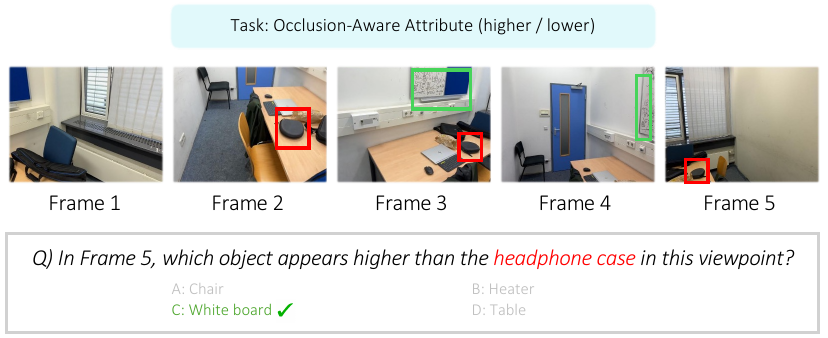}
    \caption*{(b) higher / lower}
\end{minipage}
\hfill
\begin{minipage}{\textwidth}
    \centering
    \includegraphics[width=\linewidth,height=0.26\textheight,keepaspectratio]{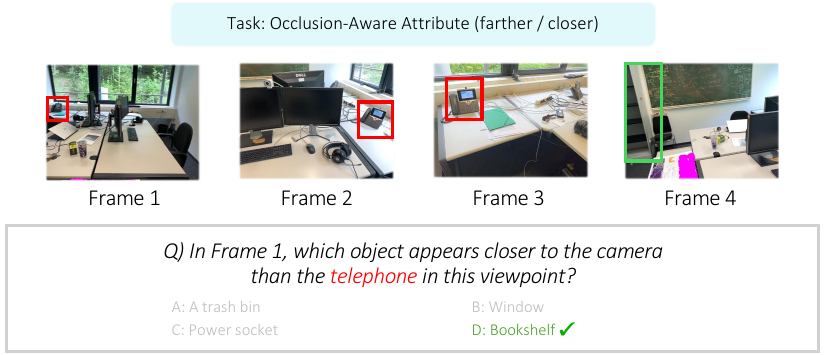}
    \caption*{(c) farther / closer}
\end{minipage}
\caption{Occlusion-Aware Attribute examples}
\label{fig:attribute_example}
\end{figure*}

\begin{figure*}[t]
\centering
\begin{minipage}{\textwidth}
    \centering
    \includegraphics[width=\linewidth,height=0.25\textheight,keepaspectratio]{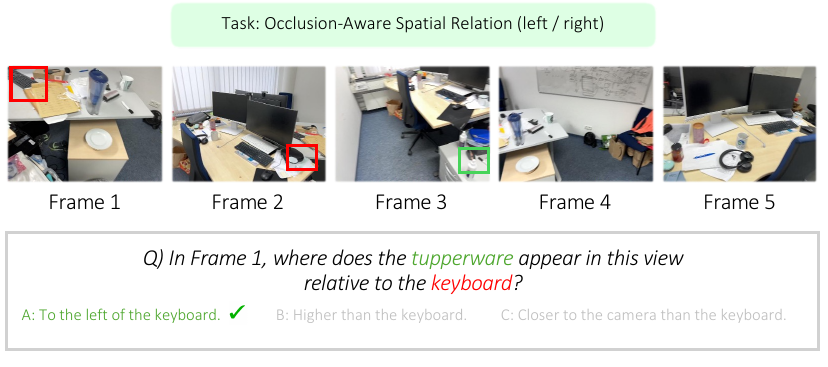}
    \caption*{(a) left / right}
\end{minipage}
\hfill
\begin{minipage}{\textwidth}
    \centering
    \includegraphics[width=\linewidth,height=0.25\textheight,keepaspectratio]{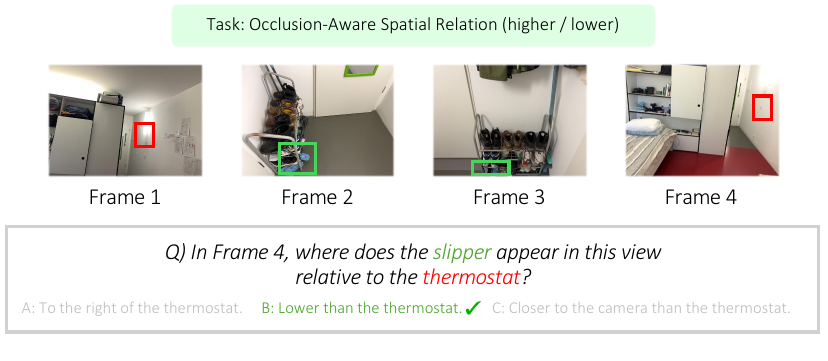}
    \caption*{(b) higher / lower}
\end{minipage}
\hfill
\begin{minipage}{\textwidth}
    \centering
    \includegraphics[width=\linewidth,height=0.25\textheight,keepaspectratio]{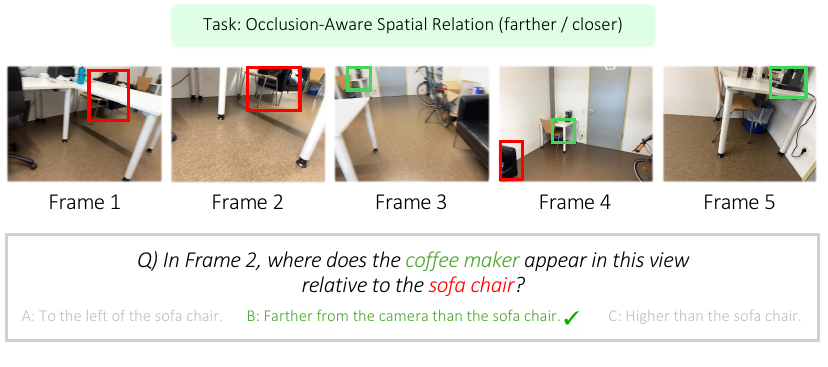}
    \caption*{(c) farther / closer}
\end{minipage}
\caption{Occlusion-Aware Spatial Relation examples}
\label{fig:relation_example}
\end{figure*}

\begin{figure*}[t]
    \centering
    \includegraphics[width=\textwidth,height=0.25\textheight,keepaspectratio]{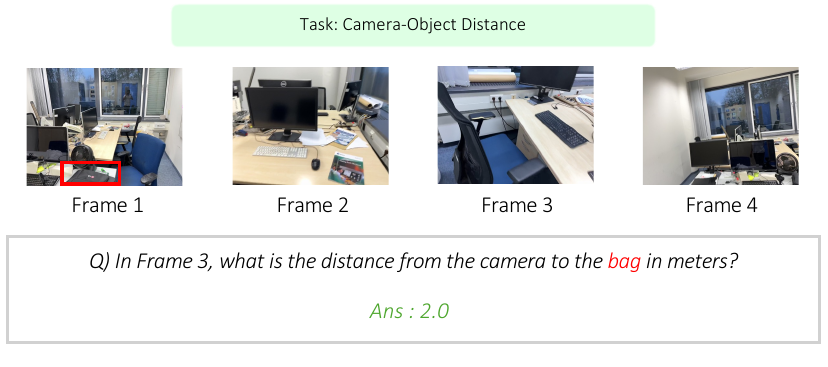}
    \clearpage

    \caption{Camera-Object Distance example}
    \label{fig:cam_obj_dist_example}
\end{figure*}

\begin{figure*}[t]
    \centering
    \includegraphics[width=\textwidth,height=0.25\textheight,keepaspectratio]{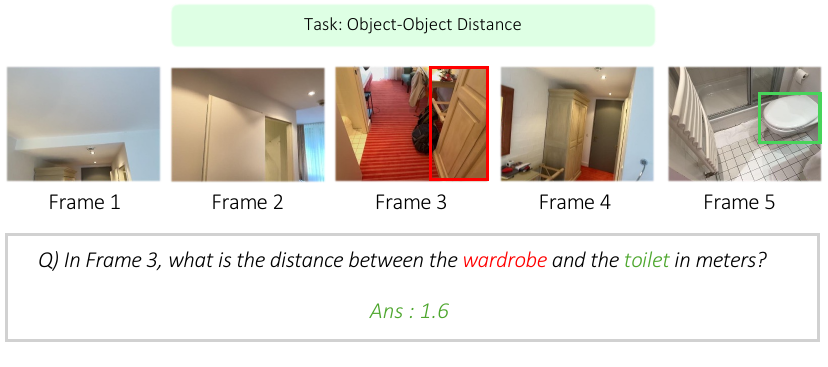}
    \clearpage

    \caption{Object-Object Distance example}
    \label{fig:obj_obj_dist_example}
\end{figure*}

\begin{figure*}[t]
    \centering
    \includegraphics[width=\textwidth,height=0.25\textheight,keepaspectratio]{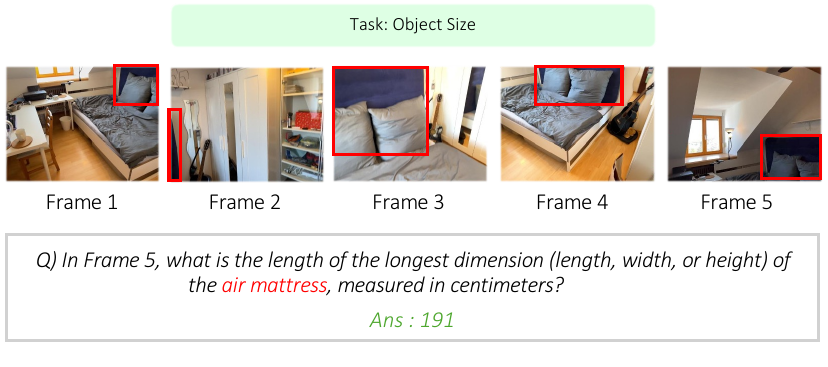}
    \clearpage

    \caption{Object Size example}
    \label{fig:obj_size_example}
\end{figure*}

\begin{figure*}[t]
    \centering
    \includegraphics[width=\textwidth,height=0.25\textheight,keepaspectratio]{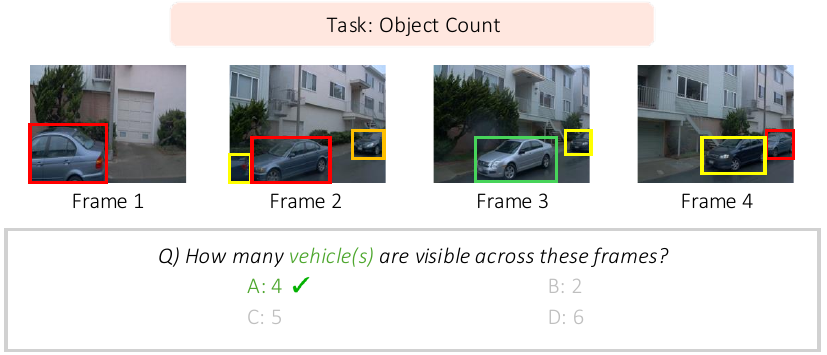}
    \clearpage
    
    \caption{Object Count example (outdoor)}
    \label{fig:count_example_outdoor}
\end{figure*}

\begin{figure*}[t]
    \centering
    \includegraphics[width=\textwidth,height=0.26\textheight,keepaspectratio]{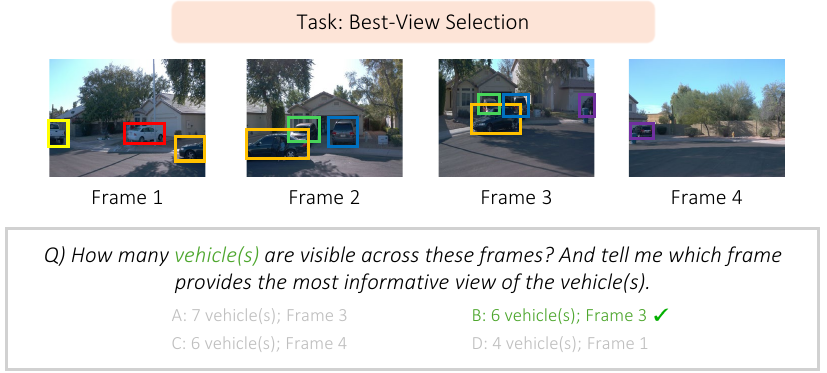}
    \clearpage
    
    \caption{Best-View Selection example (outdoor)}
    \label{fig:bestview_example_outdoor}
\end{figure*}

\begin{figure*}[t]
    \centering
    \includegraphics[width=\textwidth,height=0.25\textheight,keepaspectratio]{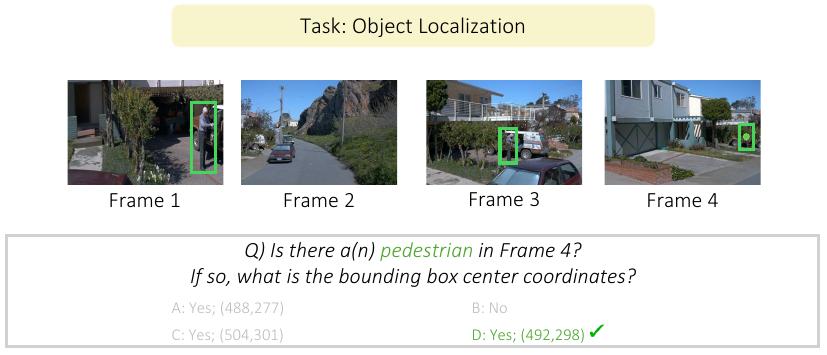}
    \clearpage
    
    \caption{Object Localization example (outdoor)}
    \label{fig:localization_example_outdoor}
\end{figure*}

\begin{figure*}[t]
\centering
\begin{minipage}{\textwidth}
    \centering
    \includegraphics[width=\linewidth,height=0.24\textheight,keepaspectratio]{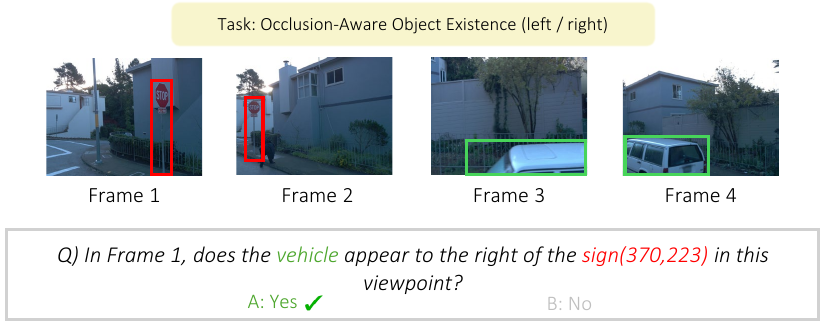}
    \caption*{(a) left / right}
\end{minipage}
\hfill
\begin{minipage}{\textwidth}
    \centering
    \includegraphics[width=\linewidth,height=0.25\textheight,keepaspectratio]{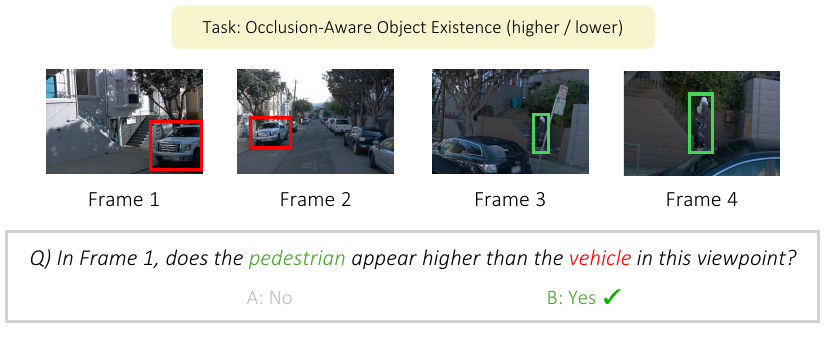}
    \caption*{(b) higher / lower}
\end{minipage}
\hfill
\begin{minipage}{\textwidth}
    \centering
    \includegraphics[width=\linewidth,height=0.25\textheight,keepaspectratio]{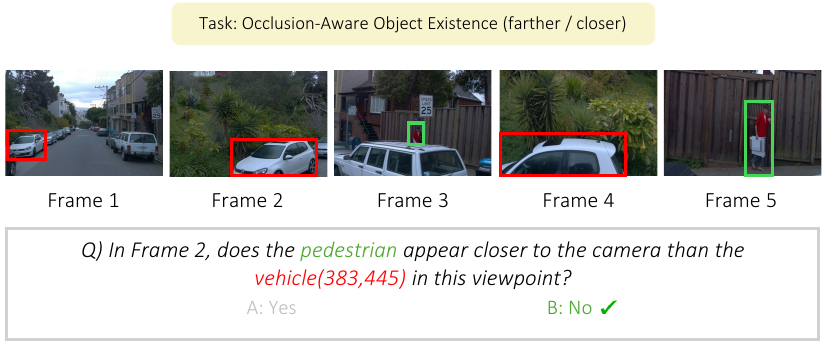}
    \caption*{(c) farther / closer}
\end{minipage}
\caption{Occlusion-Aware Object Existence examples (outdoor)}
\label{fig:existence_example_outdoor}
\end{figure*}

\begin{figure*}[t]
\centering
\begin{minipage}{\textwidth}
    \centering
    \includegraphics[width=\linewidth,height=0.25\textheight,keepaspectratio]{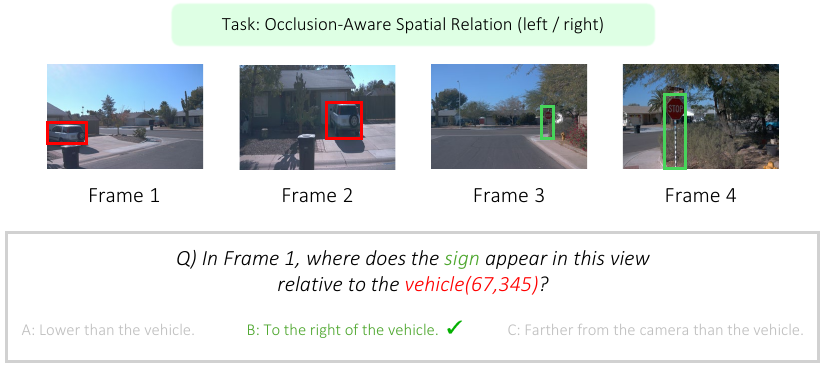}
    \caption*{(a) left / right}
\end{minipage}
\hfill
\begin{minipage}{\textwidth}
    \centering
    \includegraphics[width=\linewidth,height=0.25\textheight,keepaspectratio]{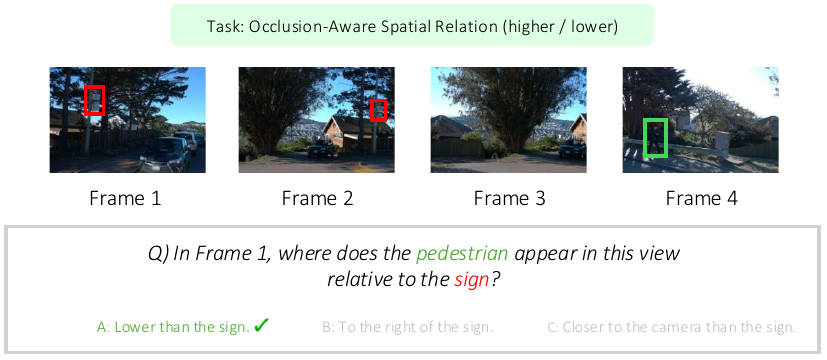}
    \caption*{(b) higher / lower}
\end{minipage}
\hfill
\begin{minipage}{\textwidth}
    \centering
    \includegraphics[width=\linewidth,height=0.25\textheight,keepaspectratio]{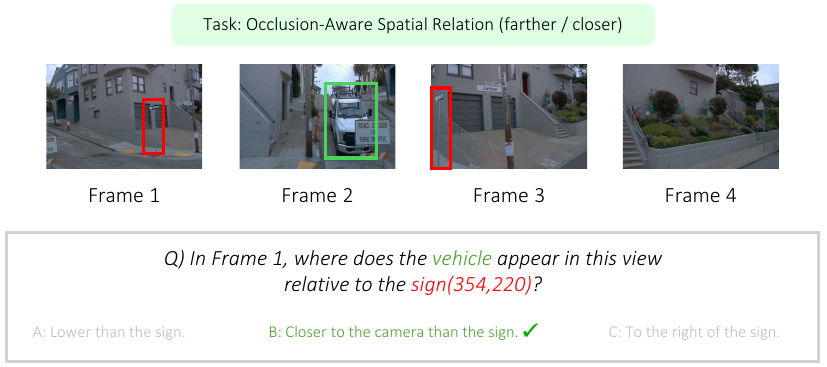}
    \caption*{(c) farther / closer}
\end{minipage}
\caption{Occlusion-Aware Spatial Relation examples (outdoor)}
\label{fig:relation_example_outdoor}
\end{figure*}

\begin{figure*}[t]
    \centering
    \includegraphics[width=\textwidth,height=0.25\textheight,keepaspectratio]{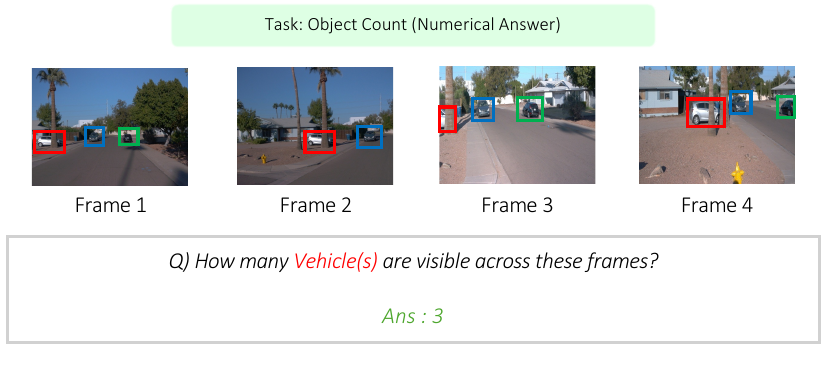}
    \clearpage

    \caption{Object Count example (outdoor, numerical)}
    \label{fig:obj_count_num_example_outdoor}
\end{figure*}

\begin{figure*}[t]
    \centering
    \includegraphics[width=\textwidth,height=0.25\textheight,keepaspectratio]{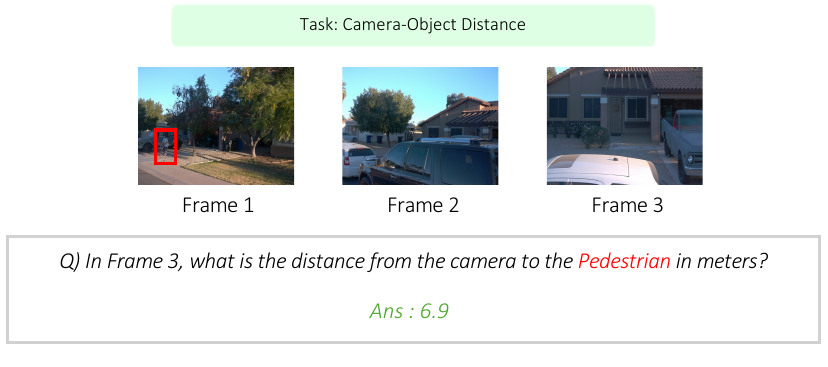}
    \clearpage

    \caption{Camera-Object Distance example (outdoor)}
    \label{fig:cam_obj_dist_example_outdoor}
\end{figure*}

\begin{figure*}[t]
    \centering
    \includegraphics[width=\textwidth,height=0.25\textheight,keepaspectratio]{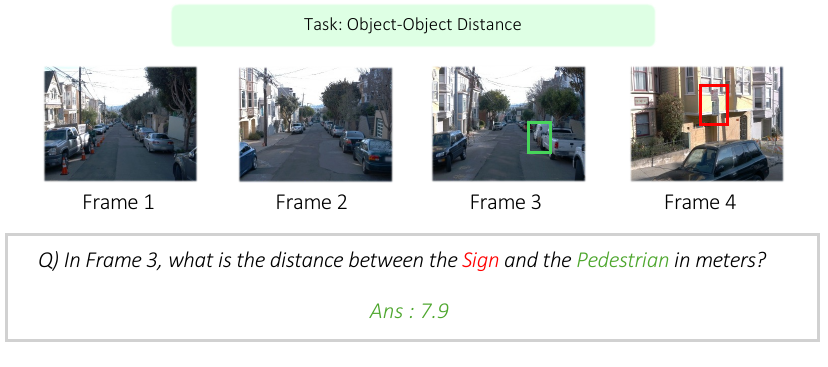}
    \clearpage

    \caption{Object-Object Distance example (outdoor)}
    \label{fig:obj_obj_dist_example_outdoor}
\end{figure*}

\ourdataset{} dataset provides 17 sub-tasks. Fig.~\ref{fig:main_figure} shows representative task examples from \ourdataset{}, illustrating how our questions are constructed under partial visibility, occlusion, and low-overlap views. We provide additional QA examples from both indoor and outdoor scenes, as shown in Fig.~\ref{fig:count_example} -~\ref{fig:obj_obj_dist_example_outdoor}.
\newcommand{\colorFrameID}[1]{\textcolor{gray}{#1}}
\newcommand{\colorCategory}[1]{\textcolor{red}{#1}}
\newcommand{\colorObject}[1]{\textcolor{red}{#1}}
\newcommand{\colorObjectOne}[1]{\textcolor{red}{#1}}
\newcommand{\colorObjectTwo}[1]{\textcolor{cyan}{#1}}
\newcommand{\colorSource}[1]{\textcolor{cyan}{#1}}
\newcommand{\colorTarget}[1]{\textcolor{red}{#1}}
\newcommand{\colorRelation}[1]{\textcolor{teal}{#1}}

\begin{table*}[t]
\caption{Question templates used in the \textbf{SpatialMosaic} benchmark. LR: Left/Right, AB: Higher/Lower, FB: Closer/Farther.}
\centering
\small
\renewcommand{\arraystretch}{1.12}
\setlength{\tabcolsep}{4pt}

\resizebox{\textwidth}{!}{%
\begin{tabular}{>{\raggedright\arraybackslash}p{2.8cm} >{\raggedright\arraybackslash}p{9.0cm} >{\centering\arraybackslash}p{2.2cm}}
\toprule
\textbf{Task} & \textbf{Question Template} & \textbf{Answer Type} \\
\midrule

Object Count 
& How many \colorCategory{\{category\}}(s) are visible across these frames?
& \{Number\} \\
\midrule

Best View Selection 
& How many \colorCategory{\{category\}}(s) are visible across these frames? And tell me which frame provides the most informative view of the \colorCategory{\{category\}}(s)?
& \{Number\} + \{Frame ID\} \\
\midrule

Existence (LR)
& In \colorFrameID{\{frame\_id\}}, does the \colorObjectOne{\{object1\}} appear \colorRelation{\{to the left of / to the right of\}} the \colorObjectTwo{\{object2\}} in this viewpoint?
& \{Yes/No\} \\
\midrule

Existence (AB)
& In \colorFrameID{\{frame\_id\}}, does the \colorObjectOne{\{object1\}} appear \colorRelation{\{higher than / lower than\}} the \colorObjectTwo{\{object2\}} in this viewpoint?
& \{Yes/No\} \\
\midrule

Existence (FB)
& In \colorFrameID{\{frame\_id\}}, does the \colorObjectOne{\{object1\}} appear \colorRelation{\{closer to / farther from\}} the camera than the \colorObjectTwo{\{object2\}} in this viewpoint?
& \{Yes/No\} \\
\midrule

Attribute (LR)
& In \colorFrameID{\{frame\_id\}}, which object appears \colorRelation{\{to the left of / to the right of\}} the \colorObject{\{object\}} in this viewpoint?
& \{Object name\} \\
\midrule

Attribute (AB)
& In \colorFrameID{\{frame\_id\}}, which object appears \colorRelation{\{higher than / lower than\}} the \colorObject{\{object\}} in this viewpoint?
& \{Object name\} \\
\midrule

Attribute (FB)
& In \colorFrameID{\{frame\_id\}}, which object appears \colorRelation{\{closer to / farther from\}} the camera than the \colorObject{\{object\}} in this viewpoint?
& \{Object name\} \\
\midrule

Spatial Relation (LR / AB / FB)
& In \colorFrameID{\{frame\_id\}}, where does the \colorTarget{\{target\}} appear in this view relative to the \colorSource{\{source\}}?
& \{Spatial Rel.\} \\
\midrule

Localization
& Is there a(n) \colorTarget{\{target\}} in \colorFrameID{\{frame\_id\}}? If so, what is the bounding box center coordinates?
& \{Coordinates\} \\
\bottomrule
\end{tabular}%
}
\label{tab:supp_qa_templates}
\end{table*}
\section{SpatialMosaic VQA Templates}
\label{sec:supple_temple}
\label{app:templates}
We leverage an automated VQA generation pipeline to construct extensive question-answer pairs. The corresponding templates for each task are listed in Table~\ref{tab:supp_qa_templates}.

\section{Failure Cases}
\label{sec:supple_failure}
\begin{figure*}[t]
\centering
\begin{minipage}{\textwidth}
    \centering
    \includegraphics[width=\linewidth,height=0.25\textheight,keepaspectratio]{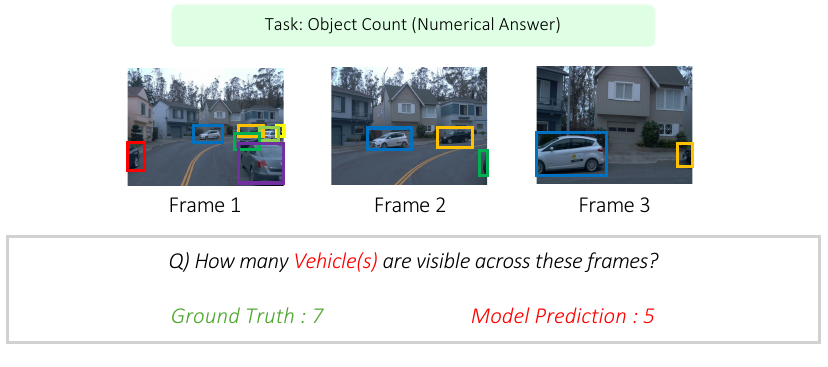}
    \caption*{(a)}
\end{minipage}
\hfill
\begin{minipage}{\textwidth}
    \centering
    \includegraphics[width=\linewidth,height=0.25\textheight,keepaspectratio]{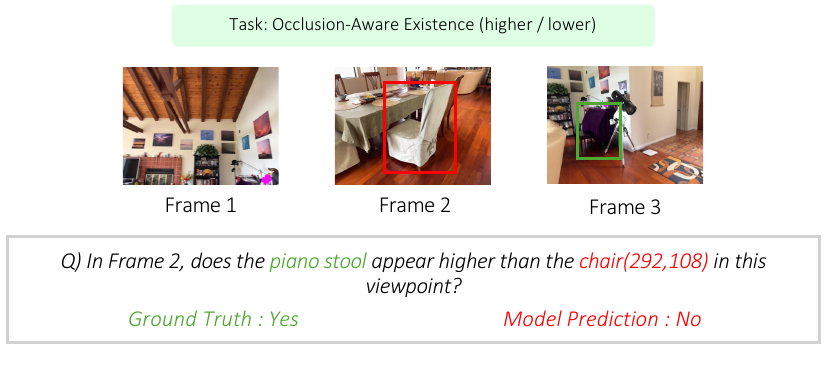}
    \caption*{(b)}
\end{minipage}
\hfill
\begin{minipage}{\textwidth}
    \centering
    \includegraphics[width=\linewidth,height=0.25\textheight,keepaspectratio]{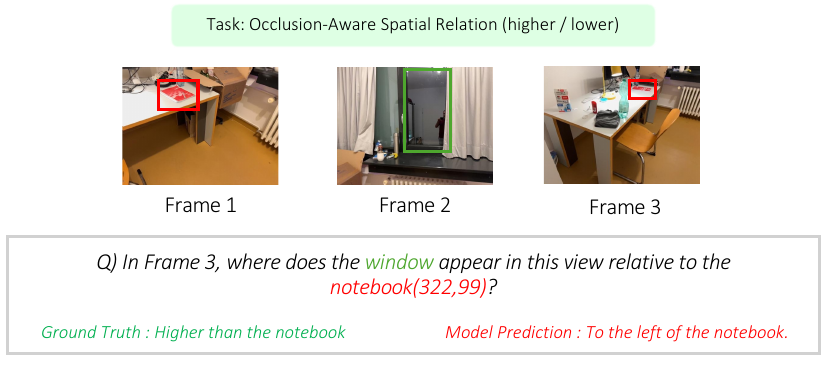}
    \caption*{(c)}
\end{minipage}
\caption{Examples of failure cases.}
\label{fig:failure_cases}
\end{figure*}

While the model fine-tuned on \ourdataset{} performs strongly overall, failures occur in low-overlap cases where cross-view correspondence cues are too sparse for reliable spatial reasoning, as depicted in (Fig.~\ref{fig:failure_cases}).

\clearpage




\clearpage

\end{document}